\RequirePackage{fix-cm}
\documentclass[twocolumn]{svjour3}          % twocolumn
\smartqed  % flush right qed marks, e.g. at end of proof
\usepackage{graphicx}
\usepackage{verbatim}
\usepackage{multirow}
\usepackage{caption}
\usepackage{subcaption}
\usepackage[table]{xcolor}
\usepackage{url}
\usepackage{hyperref}
\usepackage{cite}

\begin{document}

\title{UESegNet: Context Aware Unconstrained ROI Segmentation Networks for Ear Biometric}
%\title{Insert your title here%\thanks{Grants or other notes


\authorrunning{Pattern Analysis and Applications}

	\author{Aman Kamboj       \and
		Rajneesh Rani \and Aditya Nigam \and Ranjeet Ranjan Jha  %etc.
	}

	\institute{Aman Kamboj \and Rajneesh Rani \at
		National Institute of Technology Jalandhar \\
		Punjab, India - 144011\\
		\email{amank.cs.16@nitj.ac.in, ranir@nitj.ac.in}           %  \\
		\and
		  Aditya Nigam \and Ranjeet Ranjan Jha \at
		Indian Institute of Technology Mandi\\
		Himachal, India - 175005\\
		\email{aditya@iitmandi.ac.in, d16044@students.iitmandi.ac.in}
	}

\date{Received: date / Accepted: date}

\maketitle

\begin{abstract}
Biometric-based personal authentication systems have seen a strong demand mainly due to the increasing concern in various privacy and security applications. Although the use of each biometric trait is problem dependent, the human ear has been found to have enough discriminating characteristics to allow its use as a strong biometric measure. To locate an ear in a 2D side face image is a challenging task, numerous existing approaches have achieved significant performance, but the majority of studies are based on the constrained environment. However, ear biometrics possess a great level of difficulties in the unconstrained environment, where pose, scale, occlusion, illuminations, background clutter etc. varies to a great extent. To address the problem of ear localization in the wild, we have proposed two high-performance region of interest (ROI) segmentation models UESegNet-1 and UESegNet-2, which are fundamentally based on deep convolutional neural networks and primarily uses contextual information to localize ear in the unconstrained environment. Additionally, we have applied state-of-the-art deep learning models viz; FRCNN (Faster Region Proposal Network) and SSD (Single Shot MultiBox Detecor) for ear localization task. To test the model's generalization, they are evaluated on six different benchmark datasets viz; IITD, IITK, USTB-DB3, UND-E, UND-J2 and UBEAR, all of which contain challenging images. The performance of the models is compared on the basis of object detection performance measure parameters such as IOU (Intersection Over Union), Accuracy, Precision, Recall, and F1-Score. It has been observed that the proposed models UESegNet-1 and UESegNet-2 outperformed the FRCNN and SSD at higher values of IOUs i.e. an accuracy of 100\% is achieved at IOU 0.5 on majority of the databases. This performance signifies the importance and effectiveness of the models and indicates that the models are invariant to environmental conditions. 
\end{abstract}

% \fi

\section{Introduction}
\begin{figure}[h]
\centering
  \includegraphics[width=8cm , height=5cm]{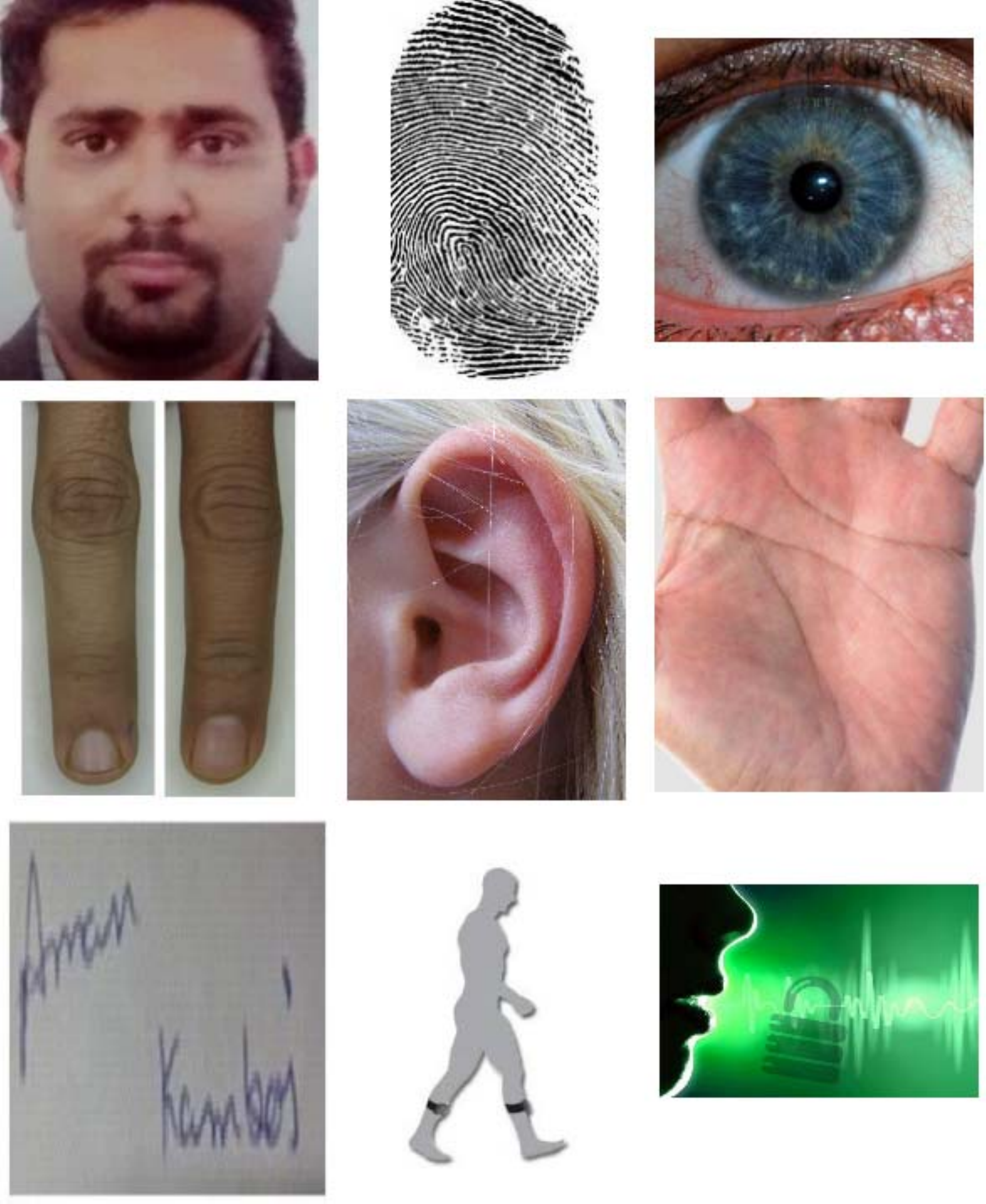}
  \caption{Well Known Biometrics Traits}
  \label{fig:trait}
\end{figure}

In the modern world, personal authentication based on physiological characteristics plays an important role in the society. With increasing concern over security, an automated and reliable human identification system is required for various applications such as law enforcement, health-care, banking, forensic and information systems etc. There are three common ways for person authentication: possession, knowledge, and biometrics. In the possession-based method, the user has to keep some tokens, identity cards or keys whereas in knowledge-based method, the user has to remember certain pin, password etc. The possession and knowledge-based methods are significant for personal authentication but they have limitations, for example in the possession-based method, 
there may be chance that item under possession get stolen or lost and in the knowledge-based method, one may forget the secret information required for authentication. As a result, one's identity can be forged and security can be compromised.
However biometric-based authentication system is based on physiological or behavioral traits of human in which there is no chance to forget or lose them. The Fig.\ref{fig:trait} shows some well-known biometrics traits used for person authentication. Researchers have reported various approaches based on physiological characteristics such as face \cite{face1,Face2}; fingerprint \cite{finger1,finger2}; iris \cite{Iris1,Iris2}; palmprint \cite{palm2,palm1}; knuckle print \cite{Knuckle1,Knuckle2,Knuckle3}; ear \cite{Ear1,Ear2}; and behavioral characteristics such as voice \cite{Voice}; gait \cite{Gait} and signature \cite{Signature} for authentication. However, there is still scope of improving the overall performance of the aforementioned authentication methods.  
	
Recognition of a person using ear has gained much attention due to its uniqueness and several advantages over the other biometrics. In 1989, A.Iannarelli \cite{Iannare1} conducted two experiments to prove the uniqueness of the ear. In his first experiment, he gathered ear images of random person and found that each of them  were different. In his second experiment, he examined identical twins and found that even though the other physiological features are same but the ears are not identical. The studies supported the uniqueness of the ear and motivated researchers to use ear for person authentication. Moreover, the ear is a non-intrusive biometric which can be captured easily at a distance, whereas fingerprint, iris, palm-print etc. are intrusive biometrics that cannot be captured at a distance and need more user cooperation. Ear images can be acquired using digital cameras, however, a dedicated hardware is required for acquisition of images for fingerprint, iris, palm-print etc. Unlike the face, it has a stable structure and is not affected by age, expression etc. In addition, ear images are smaller in size as compared to face and work well under low resolution. %It is also useful for partial face matching.
    
\begin{figure}[h]
		\centering	
		\includegraphics[width=8cm , height=5cm]{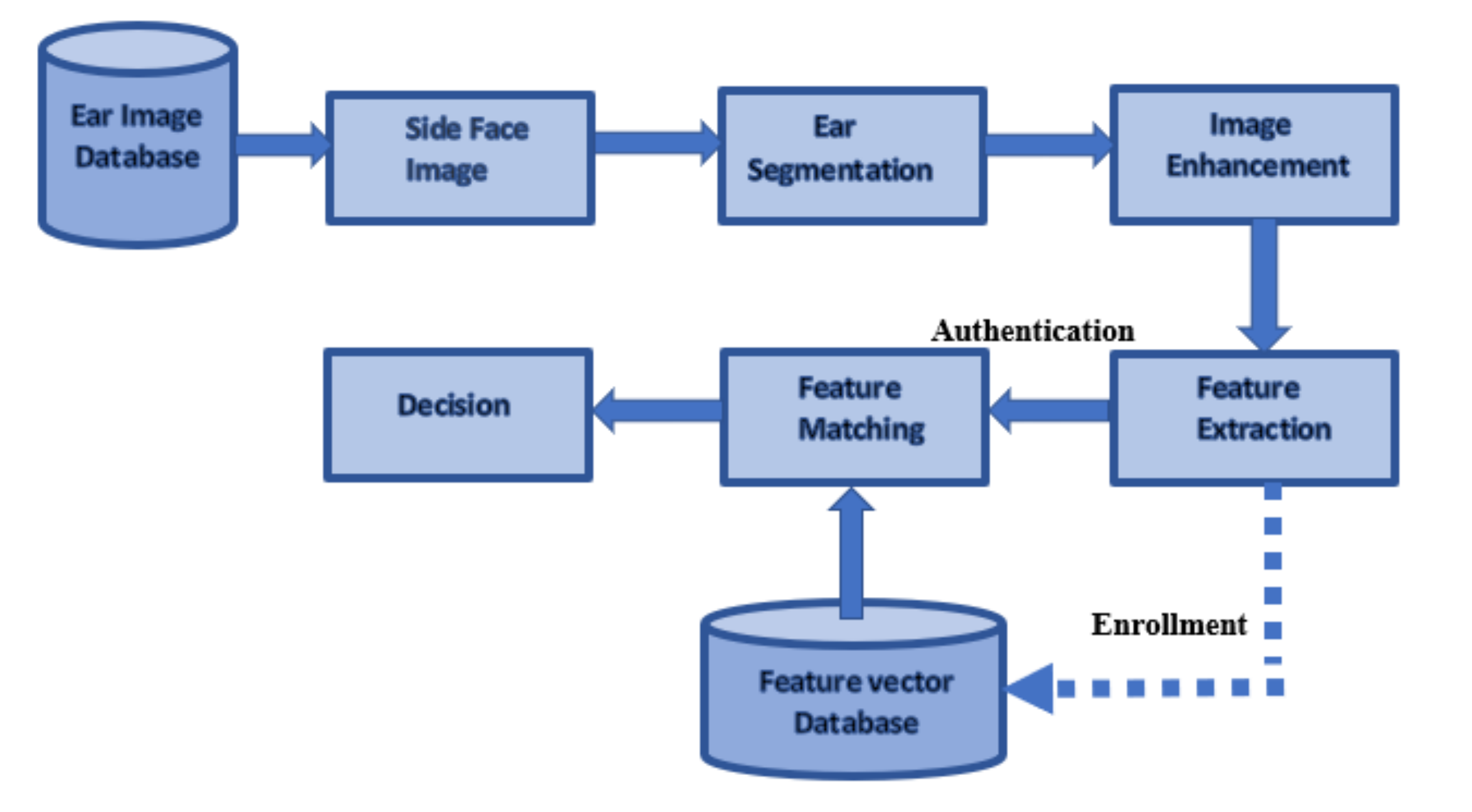}
		\caption{Overall Process of Biometric Authentication System}
        \label{process}
\end{figure}

An ear based biometric authentication system for human recognition is a multi-stage process as shown in Fig. \ref{process}. In the initial stage, a database of side face images is prepared using some acquisition devices. Further, from the image the desired part of the trait, known as the region of interest (ear) is segmented. In the next stage, image ROI goes through enhancement steps like alignment and correction. Afterwards, unique features are extracted and stored in the database (this is known as the enrollment process). At the authentication time, test image goes through similar stages and extracted features are matched against stored features in a database to authenticate the claim. 

\begin{figure}[h]
		\centering	
		\includegraphics[width =8cm, height=6cm]{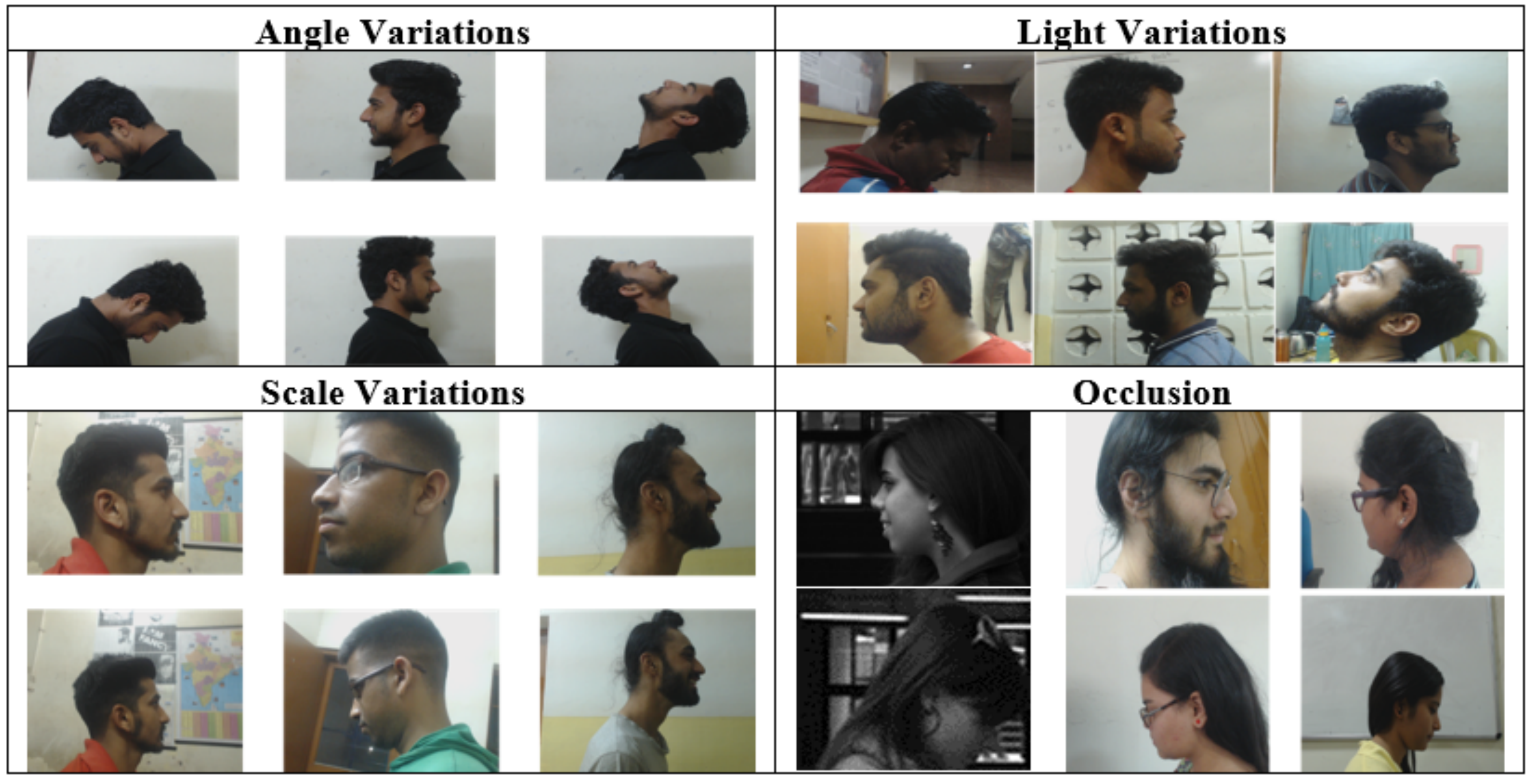}
		\caption{Images of unconstrained environment}
        \label{ear1}
	\end{figure}

 The very first step in any in biometric-based authentication system is to extract the desired Region of Interest (ROI). As it plays a pivot role in overall performance. In the past, many researchers have worked on ear detection in the constrained environment, where the images are being captured under some controlled setting. In this paper, our focus is on ear detection from side face images captured in the unconstrained environment (wild). In unconstrained environment, the images can vary in terms of occlusion by (hair, earrings), pose, light, blur, scale, variations (refer Fig.\ref{ear1}). The detection of the ear in the side face images captured in wild possesses a great level of challenge. So, there is a need to develop an appropriate automated system to perform the ear localization from the side face image in the real imaging conditions.

\subsection{Related Work}

This section discusses some of the well known and recent ear localization approaches from the side face image of a person, which are based on machine learning and deep learning techniques.

\subsubsection{Machine learning approaches for ear localization} 

In \cite{14}, the authors presented an ear detection for online biometrics applications. They have used a gaussian classifier to segment the ear from the skin and non-skin areas and then applied Laplacian of Gaussian to find the edges in the skin area. In addition, authors have used Log-Gabor filter and SIFT  for features extraction. The experiment was conducted on IIT Delhi database, which consist of 100 subjects with 7 samples each. The results shows that SIFT features (GAR (genuine acceptance rate) =95\%, FAR (False acceptance rate)=0.1\%) are better than Log-Gabor (GAR=85\%, FAR=0.1\%). In \cite{20}, the authors proposed an ear localization technique from side face images. It is based on connected components of a graph obtained from the edge map of face images. The proposed technique is shape, rotation and scale invariant. The experiment was performed on IIT Kanpur database of face images under varying background and poor illumination and UND-E and UND-J2 collections. The method achieved 99.25\% accuracy on IIT Kanpur database and 99.25\% on the UND-J2 collection and 96.34\% on UND-E collection.	In \cite{36}, the authors presented an automatic ear detection based on three geometric features viz; elongation, compactness and rounded boundary. Elongation is the ratio between the boundary height and width of the ear, and this ratio should be greater than 0.5. Compactness is the ratio of area and perimeter of the object (human ear's perimeter is less than its area). The third feature is the boundary of ear which is most rounded in the human body. This experiment has performed on UND-J2 dataset of 200 side face images and achieved an accuracy of 98\%. In \cite{40}, the authors have presented ear localization using context information and feature level fusion. The proposed approach has four stages: Initially, edges and shapes are extracted from the depth of an image and texture feature. In the next stage, extracted components are fused together in the image domain, afterwards, these components are merged with each other to ear candidates and score for each candidate is calculated. Then in the final stage, the rectangular box of the best ear is returned as an ear region. The proposed method can detect both left and right ear and is invariant to rotation. The proposed technique localizes the ear and also estimate the orientation of the ear. The experiment was conducted on UND-J2 collection having color images with depth for 404 different subjects with total of 1776 images. The proposed method achieved an accuracy of 99\% on profile face images.

	 A binary particle swarm optimization based on entropy for ear localization under an uncontrolled environment conditions (such as varying pose, background occlusion, and illumination) is discussed in \cite{33}. The technique calculates values for entropy map and the highest value is used to localize the ear in the side face image. To remove the background region, they applied dual-tree complex wavelet transform. The experiment was conducted on four different benchmark face datasets: CMU PIE, Pointing Head Pose, Color FERET, and UMIST, and achieved localization accuracy of 82.50\%, 83.90\%, 90.70\% and 77.92\% respectively. In \cite{34}, authors have presented a method for ear localization using entropy cum hough transformation. They have used skin segmentation for preprocessing of the input image. To extract the features, they have used entropic ear localizer and ellipsoid ear localizer, and a combination of both for localization of ear. In addition, they have used  ear classifier based on ellipsoid for the verification of the presence of ear in facial images. The experiment was performed on five face databases (FERET, Pointing Head Pose, UMIST, CMU-PIE, and FEI) and achieved localization accuracy of 100\% on FEI and UMIST, 70.94\% on PHP, 73.95\% on FERET and 70.10\% on CMU-PIE databases. In \cite{38}, the authors proposed a deformable template-based approach for ear localization. The deformable template is used for matching, is able to adapt different shapes and tolerate a certain range of transformation. They have used template matching with dynamic programming approach to localize ear. The experiment is tested on 212 face profile images. All the images were captured under the uncontrolled environment. The method achieved 96.2\% localization accuracy and 0.14\% false positive rate.
	
    \subsubsection{Deep learning approaches for ear localization} 
    Recently, the deep learning models have improved state-of-the-art in image processing. Various Artificial intelligence tasks such as classification and detection have obtained improved performance with the advent of deep learning. The object detection models of deep learning like F-RCNN (Faster region based convolution neural network \cite{FRCNN}), SSD (Single Short Multi Box Detector \cite{SSD}), R-FCN (Region Based Fully Convolution Network \cite{RFCN}), YOLO (You Only Look Once \cite{Yolo}), SSH (Single Stage Headless Face Detector \cite{SSH}), SegNet (Segmentation Network \cite{SegNet}) have achieved state-of-the-art in object detection accuracy. Some of the recent approaches based on deep learning for ear detection are discussed below:

In \cite{1}, the authors proposed a faster region-based CNN model to localize ear in multiple scale face images  captured under the uncontrolled environment (images with large occlusion, scale and pose variations). The RCNN (Region based convolutional neural network) recognizes the ear using morphological properties but sometimes it fails to detect ear from similar objects. This model is trained on multiple scale of images to identify three regions viz; head, pan-ear, and ear. Then, a region based filtering approach is applied to identify the exact location of ear. The experiment was tested on UND-J2, UBEAR databases. The model has achieved ear localization accuracy of 100\% on UND-J2 database and 98.66\% on UBEAR database. In \cite{10}, authors have used an geometric morphometrics for automatic ear localization and CNN for automatic feature extraction. The CNN network is trained on manually landmarked examples, and the network is able to identify morphometric landmarks on ear's images, which almost matches with human landmarking. The ear images and manual landmarking is obtained from CANDELA initiative (consist of 7500 images). This model has been tested on 684 images and achieved an accuracy of 91.86\%. In \cite{26}, presented pixel-wise ear localization using convolutional encoder-decoder. This model is based on SegNet architecture for distinguishing pixel between ear and non-ear. The experiment was conducted on Annotated Web Ears (AWE) dataset of 1,000 annotated images from 100 distinct subjects. In addition, they have also compared the performance with the HAAR method. This model has achieved 99.21\% ear localization accuracy while HAAR based method obtained an accuracy of 98.76\%.
    
    From the study of literature it has been found that much of reported work is performed on either constrained environment or in quasi unconstrained environment (wild). This may be due to the lack of ear databases in the wild. Although researcher have not considered Intersection Over Union (IOU) parameter to measure the accuracy of their model. However, In \cite{26}, the authors proposed a method for localization of both the ears in face image captured in the wild, but this method cannot be used for ear recognition purpose as it detects both the ears in the front face. In \cite{1}, the authors  have proposed multiple scale faster region-based CNN for ear localization on the unconstrained side face image database but did not considered IOU parameter to measure the accuracy of their model.

\subsection{Intersection Over Union Parameter}    
    
\begin{figure}[h]
		\centering	
		\includegraphics[width =8cm, height=5cm]{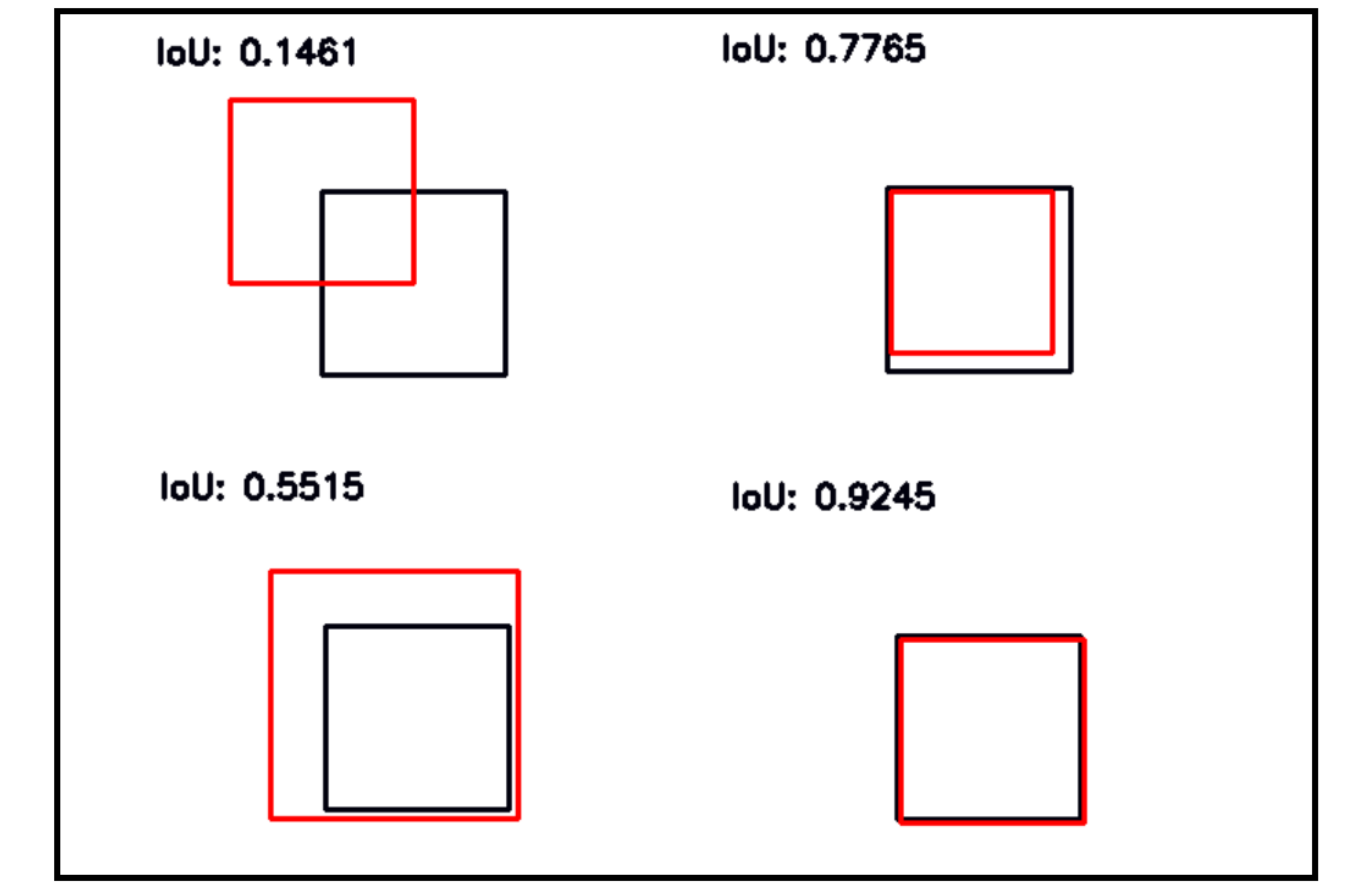}
		\caption{IOU for various bounding box, The bounding box in black is for predicted box and in red for ground truth bounding box}
        \label{IOU}
	\end{figure}

 In the literature, it has been found that researchers have proposed various methods for localization of ear in the side face image of the person and achieved satisfactory results, but ignored the parameter Intersection Over Union (IOU) to measure the accuracy. This is a very important parameter to measure the performance of any object localization task as it indicates, how much area of the predicted bounding box is overlapped with ground truth box. The value of IOU ranges from 0 to 1; where 0 indicates that the boxes do not overlap at all, 0.5 to 0.6  indicates poor overlapping, 0.75  good overlapping and 0.9 for excellent overlapping as shown in Fig. \ref{IOU}. The higher value of IOU indicates better accuracy. An IOU $>$ 0.9 indicates tightly overlapping of predicted and ground truth boxes. However an IOU=0.8 also indicates a very closed overlapping, so in this paper we have measured the performance of models till an IOU=0.8 by considiring it best for biometric authentication system. 

\subsection{Major Contributions}
\begin{enumerate}

\item To address the problem of ear localization two models UESegNet-1 and UESegNet-2 are proposed which utilizes the contextual information to localize ear in the 2D side face images captured in the wild. 

\item To access the performance of proposed models, we have modified existing state-of-the-art deep learning models FRCNN and SSD for ear localization task and compared their performance with our proposed models. 

\item To evaluate the performance of ear ROI segmentation models six different benchmark datasets (constrained and unconstrained) are used.
%\item To evaluate the performance of ear ROI segmentation models six different benchmark datasets (constrained and unconstrained environment) are used.

\item To measure the performance of models, An IOU parameter is used, which has been ignored by most of the state-of-the-art methods.  
 
\end{enumerate}

\subsection{Models Justification}
Ear localization is a very important and crucial step for ear based biometric authentication system and this need to be accurate at higher values of IOUs (Intersection over Union). In the literature, most of the work is performed on the constrained environment. But, ear localization in 2D side face images for the unconstrained environment is a very challenging problem. We have applied existing deep learning models FRCNN and SSD and evaluated their performance on both constrained and unconstrained datasets. These models performed good for constrained datasets, but their results are not satisfactory for unconstrained datasets at higher values of IOUs. On the observation, it has been found that these models do not consider contextual information for localization task. However, the contextual information plays a crucial role in the case of ear localization from side face images. Hence we have proposed two models, UESegNet-1 and UESegNet-2, which are fundasmentally based on deep learning and utilizes the contextual information to localize the ear. The result of these models are found promising for unconstrained datasets at higher values of IOUs. 

The rest of the paper is organized as follows: section 2 discusses the detailed architecture of proposed models for ear ROI segmentation. The section 3 provides the details of benchmark ear datasets. Testing protocol and various model evaluation parameters are described in section 4. The section 5 discusses the results of models and performance comparison with existing state-of-the-art methods, and the next section concludes the overall work of this paper.

\section{Deep Learning Based Ear ROI Segmentation Models}

Deep learning has gained much attention in the various object detection task and has achieved significant performance. In this paper, we have discussed four methods inspired by state-of-the-art methods for object detection, to localize the ear in 2D side face images captured in wild. The section is divided into two parts: ear segmentation by existing and proposed models. In the first part we have modified two models FRCNN and SSD for ear localization task and in the second part we have proposed two models viz; UESegNet1 and UESegNet2 which utilize the context information to localize the ear. The models uses existing CNN network (ResNet-50, VGG-16, Alex-Net etc.) as a base to extract discriminate features, which consist of a series of layers including convolutional, batch normalization, max pooling etc. It is known that for the training of any deep learning model from scratch, one need millions of input data otherwise a case of over-fitting arises. To overwhelm this problem, we have used pretrained-weight of VGG-16 (trained on ImageNet dataset) for training our models. The detailed architecture and training details for these models are discussed in detail as below:

\subsection{Ear ROI Segmentation by Existing Models}

In literature FRCNN and SSD have achieved excellent results in the object detection task, so we have deployed these models for ear localization. The detailed discussion about these models is as below:

\subsubsection{FRCNN: Faster Region Proposal Network}

\begin{figure}[h]
		\centering	
		\includegraphics[width=8cm,height=12cm]{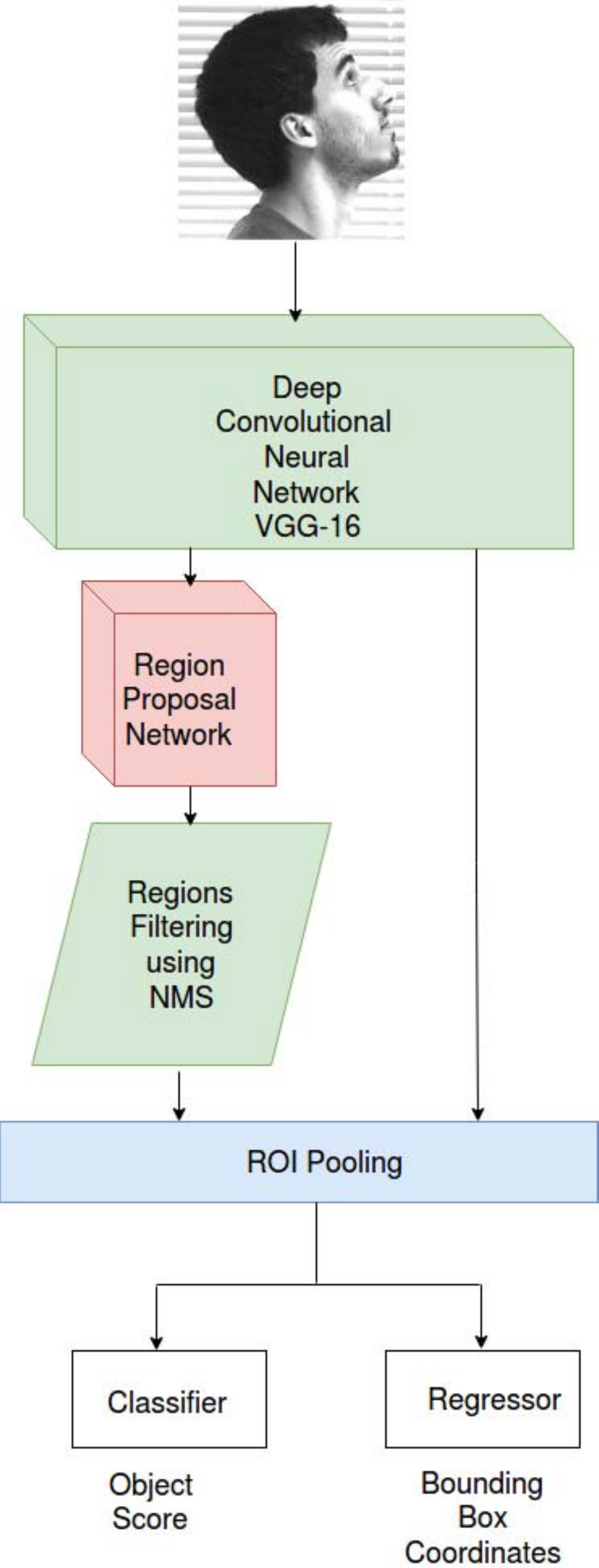}
		\caption{Architecture of FRCNN \cite{FRCNN} }
        \label{Arch_FRCNN}
	\end{figure}

The  Faster RCNN is proposed by \cite{FRCNN}, which consist of several components (shown in Fig. \ref{Arch_FRCNN})  viz; Shared layers, RPN (region proposal network), ROI (Region of Interest) pooling layer, classification and regression heads. Initially the shared layers of VGG-16 are used to get the aggregate features known as feature map. Afterwards, the feature map of the shared layer is given to the RPN, where a sliding window of size $3\times3$ convolved over this, and for each center pixel it produces K anchor boxes of different scales [128$\times$128, 256$\times$256, 512$\times$512] and ratios [1:1, 2:1, 2:2].
The RPN layer predicts  2*K (objectness score) and 4*K (box coordinates) relative to K anchor boxes, which are later fed to NMS (Non-Maximum Suppression) module to eliminate the redundant boxes. The regions produced by RPN layer are variable in size, so they are given to ROI pooling layer which converts these regions into fixed size (14$\times$ 14) and applies max-pooling on these boxes. The filtered regions are then given to classification and regression heads for the prediction of class score and bounding box coordinates.

\textbf{Training Strategy}: During training, Adam Optimizer (learning rate = 0.00001) and stochastic gradient descent (learning rate = 0.001) are used for RPN layer and overall layers respectively.
The model is trained for 100 epochs and for each epoch RPN is trained for another 200 epochs. The model uses binary cross entropy loss for classification and mean squared error for regression. During training, the model tries to minimize these losses.

\subsubsection{SSD: Single Shot MultiBox Detector}
\begin{figure}[h]
		\centering	
		\includegraphics[width=9cm,height=11cm]{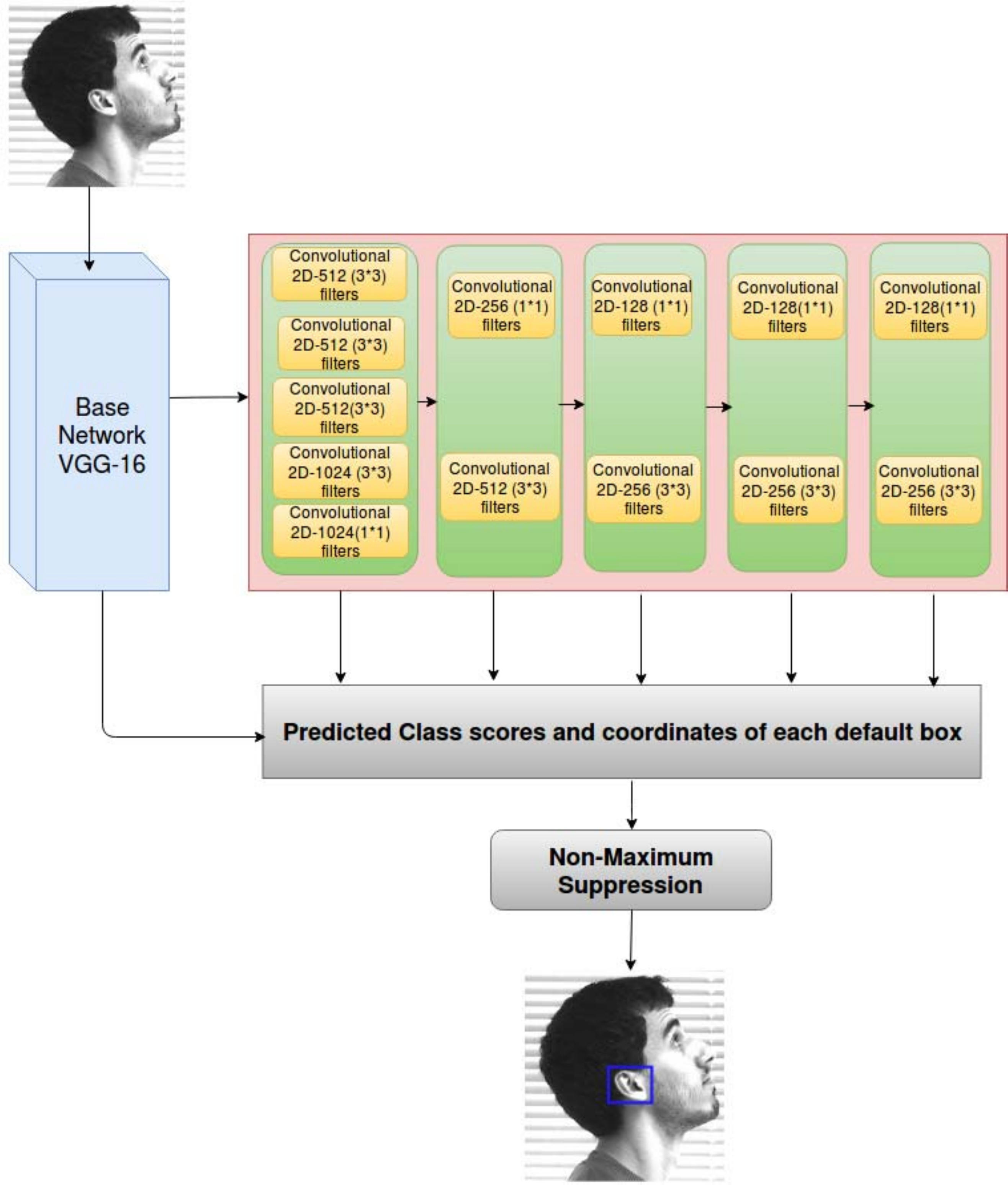}
		\caption{Architecture of SSD \cite{SSD} }
        \label{ssd1}
	\end{figure}
The overall architecture of SSD is shown in Fig. \ref{ssd1}. This model is proposed by \cite{SSD}, which consist of two components viz; Base Network (CNN model) and Additional Series of Convolutional Layers. The base network is taken from state-of-the-art CNN models such a VGG-16, VGG-19, ResNet-50, Alex-Net and, Inception etc. In this paper, we have used VGG-16 as a base network to extract meaningful feature. After base network, there are 5 set of convolution layers which progressively reduces the size of the feature map and hence help to predict bounding boxes at multiple scales. As it is shown in Fig. \ref{ssd1}, the first set of layers contains five convolution layers in which first 4 layers have filters of size $3\times 3$ and last layer with filter size of $ 1 \times 1$. The last layer is used for aggregating the features of all the channels in the feature map. The output feature map of the first set is given to the prediction module, and to the second set simultaneously. For set two, we have two convolution layers with filters size $1\times 1$ and $3\times 3$ which help further to aggregate the features. The output of this set is given to both third set and prediction module respectively. Similarly, for other sets, we have different convolution layers and which are connected to the prediction module. Finally, different offset to the default boxes (as in Faster RCNN \cite{FRCNN}) of different ratios and scales and their associated confidences are provided by each set of convolution layers. The predicted default boxes of feature maps are fed to NMS (Non-Maximum-Suppression) module. 
This module compares defaults boxes to the ground truth and provide the boxes having Intersection Over Union (IOU) $>$ 0.5. 

\textbf{Training Strategy}: During training, stochastic gradient descent is used with momentum = 0.9, Initial learning  rate = 0.001,  Final  learning  rate = 0.0001, and weight decay = 0.00001. The model is trained for 100 epochs and uses two types of losses viz; Classification loss and Regression loss. The classification loss is calculated using cross entropy loss and regression loss is calculated using smooth L1 loss. 

\subsection{Ear ROI Segmentation by Proposed Models}

To address the problem of ear localization, we have proposed two models UESegNet-1 and UESegNet-2. The detailed architecture and implementation details is discussed as below:    

\subsubsection{UESegNet-1}

\begin{figure}[h]
		\centering	
		\includegraphics[width=8cm,height=6cm]{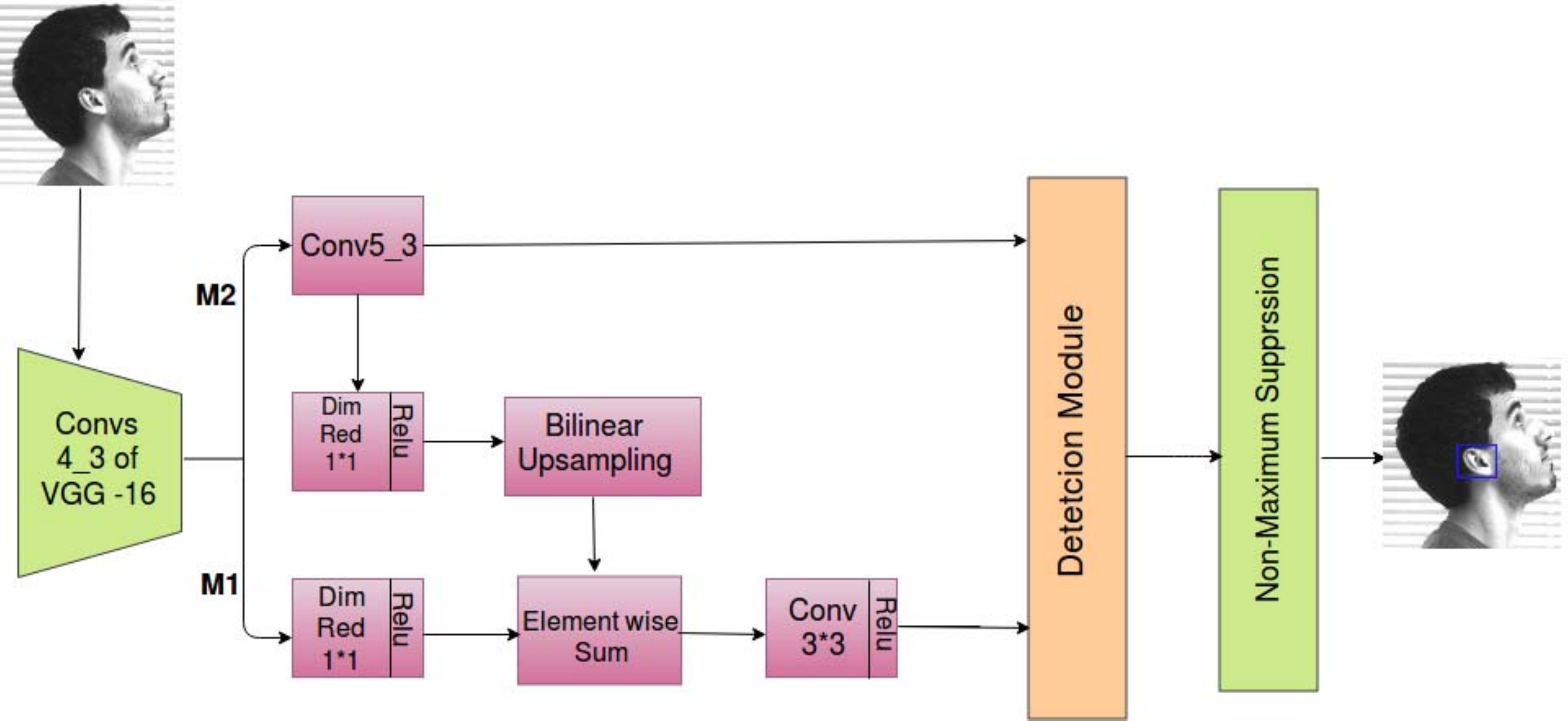}
		\caption{Architecture of UESegNet-1 }
        \label{SSH}
	\end{figure}

\begin{figure}[h]
		\centering	
		\includegraphics[scale=0.60]{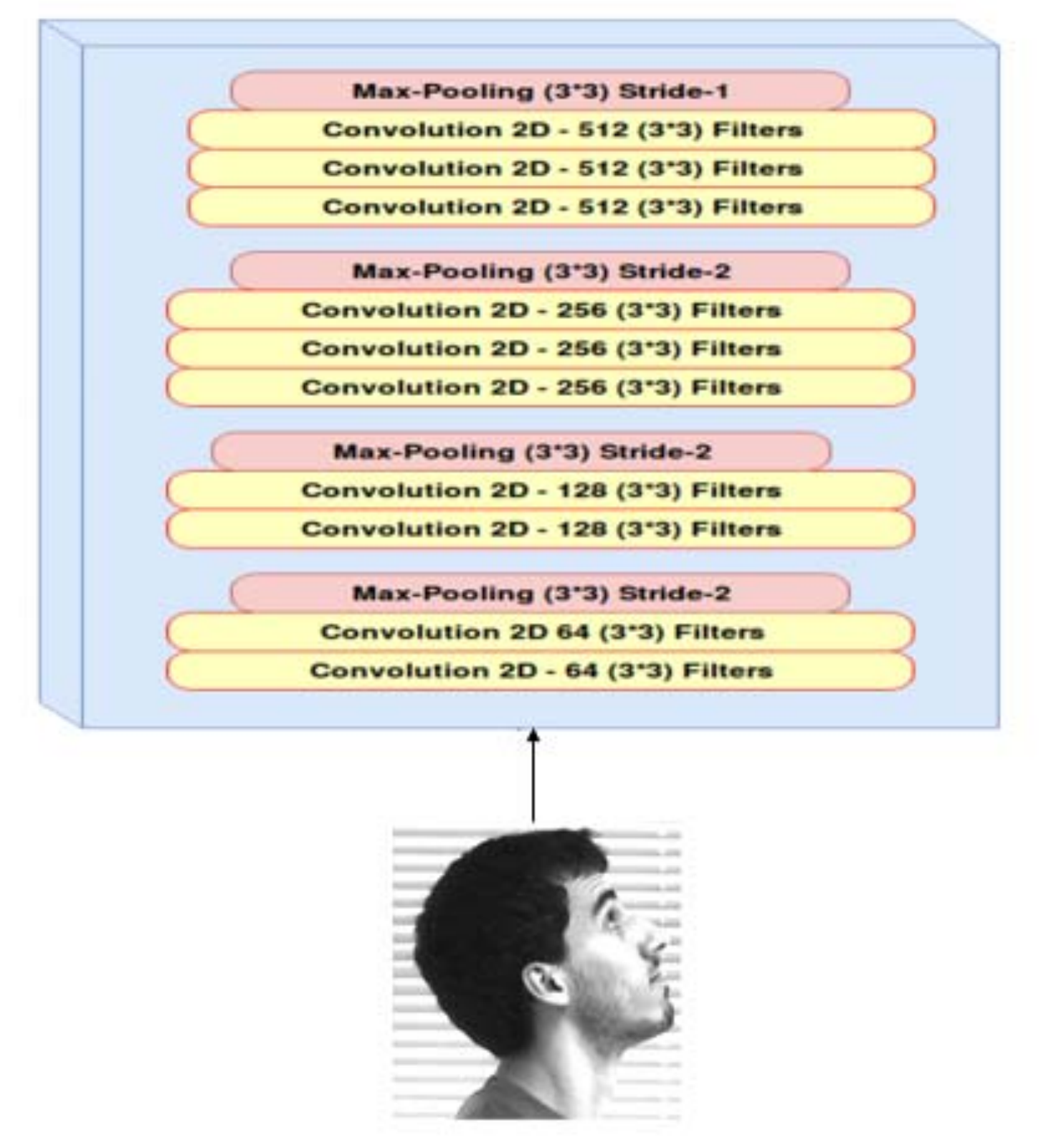}
		\caption{Base Network}
        \label{base}
	\end{figure}

The architecture of UESegNet-1 is shown in Fig. \ref{SSH}, which takes side face images as the input and produces segmented ears. However, unlike FRCNN, this is a single stage architecture which performs localization and classification. In this proposed architecture, localization is performed at two levels to incorporate the scale-invariance.  Initially, we have taken VGG-16 as a base network (refer to Fig. \ref{base}) which is common for both levels. However, we have abridged the VGG model by eliminating all the fully connected layers and left with only convolution layers. Since later layers of the VGG provides aggregate features which are helpful in localization properly, hence we prefer to take feature maps from those layers. The VGG-16 network contains several convolution and pooling layers. As it can be seen in Fig. \ref{base}, that there are 10 convolution layers and 4 max-pooling layers, which is pruned version of VGG. Each convolution layer in this network contains filters size of $3\times 3$, which convolves on image and provides output feature map. In initial convolution layers, these filters learn the local features such as edges, lines etc., but in later convolution layers filters started to learn aggregated features such as shape, box etc. In addition, the network has max pooling layer to reduce feature map and to make these features invariant to rotation and translation. The feature maps obtained after $10^{th}$ and $13^{th}$ convolution layers has been given to the different levels M1 and M2.

At the first level M1, the feature maps of the convolution layers $4_3$ and $5_3$ (of VGG)  with dimension $40 \times 40 \times 512$ and $20 \times 20 \times 512$ have taken respectively. At this level, we have used the idea of feature map fusion for merging these two feature maps. However the dimension of both feature maps are different hence bi-linear up-sampling are applied on second feature map to come up with the same size as first, and then these feature maps are combined using element-wise sum. In addition, we reduce the number of the channel from 512 to 128 (using $1 \times 1$ convolutions) to reduce memory consumption without compromising with overall performance. As the network combines two types of aggregate features hence we come up with a sharp feature map. Now, this sharp feature map is convolved with $3 \times 3$ filters which further help in moving towards more aggregate features. 

\begin{figure}[h]
		\centering	
		\includegraphics[width=8cm,height=4cm]{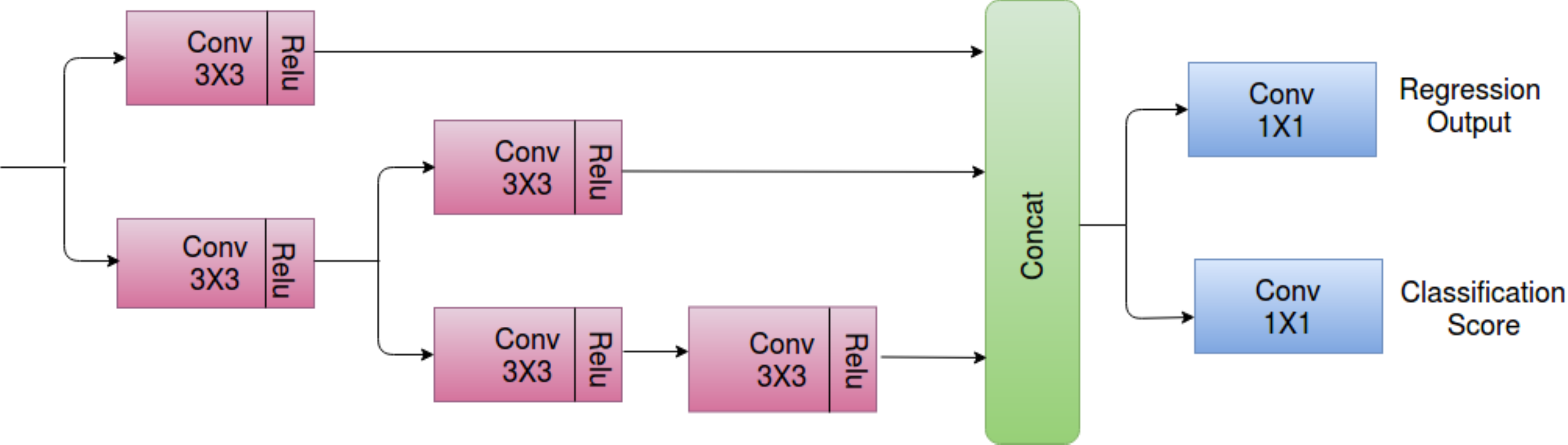}
		\caption{UESegNet-1 Detection Module ( utilize context information)}
        \label{SSHDet}
	\end{figure}

Up to this point, the architecture has focused only on aggregate features. However, the context information also plays a crucial role as surrounding region of the ear has significant texture information, which helps to classify and localize the ear against nearby parts. As the context information is important hence few layers are added regarding context as shown in Fig. \ref{SSHDet}, which consist of three context layers with $3 \times 3$, $5 \times 5$ (two $3 \times 3$ equivalent to $5 \times 5$) and $7 \times 7$ ( three $ 3 \times 3$ equivalent to $7 \times 7$). However, a large filter has more parameters as compared to few small sequential filters, so we prefer to take small filters for reducing the overall complexity. The output feature maps of aforementioned layers are further concatenated and provided to the classification head and regression head, which gives the classification score and regression output respectively.\\

At M2 level, the output feature of VGG-Conv$5_3$ layer is taken as this feature map contains more aggregate information. The context layers used at M1 level are also applied at M2 level as shown in Fig. \ref{SSH}. The output feature maps of context layers have further concatenated and given to the classification head and regression head, which do the final prediction and returns bounding boxes along with classification score. Finally, non-maximum suppression (NMS) algorithm (as discussed below) has been applied over all the predicted boxes (from M1 and M2) by taking threshold 0.7 to eliminate redundant boxes.\\

\textbf{Non Maximum Suppression Algorithm:}
\begin{enumerate}
\item Sort all boxes of a class using confidence scores.
\item Calculate IOU (Jaccard Index) of first box with every other box.
\item If IOU overlap $> 0.7$, remove the other box.
\item Otherwise keep the other box.
\item Repeat the above steps for each box in sorted order.
\end{enumerate}
\textbf{Training Strategy}:
It would amiss with the model if we go for training from scratch. As we have only 7100 images of ear hence if we train the network from scratch then the case of over-fitting will arise. To avert this problem, we have used weights of VGG-16 (Pre-trained on Image-net Dataset). However this weight matrix is defined for RGB images, so we convert all the images into RGB. In addition, we have taken different hyper parameters such as stochastic gradient descent, epoch $=$ 100, momentum $=$ 0.9, Initial learning rate $=$ 0.003, Final learning rate $=$ 0.004, weight decay $=$ 0.0004 etc.

\textbf{Loss function of UESegNet-1:} The UESegNet-1 has two types of loss functions: Classification loss and regression loss; which are calculated as per equation (1).     
   \begin{equation}
  \sum_{k}\frac{1}{N_{k}^c}\sum_{i\epsilon A_k}l_c(p_{i},g_{i})+\lambda \sum_{k}\frac{1}{N_{k}^r}\sum_{i\epsilon A_k}I(g_i=1)l_r(b_{i},t_{i})
	\end{equation}
  Here $l_c$ is ear classification loss\\
  $A_k$ is set of anchors defined in detection module\\
  $p_i$ is predicted category of label\\
  $g_i$ is ground truth label\\  
  Here $l_r$ is ear regression smooth L1 loss\\
  $N_{k}^c$ is number of anchors in detection module\\
  $b_i$ is predicted coordinates of $i^th$ anchor box\\
  $t_i$ is ground truth coordinates  of $i^th$ anchor box\\  
  $\lambda$ is a constant weight
  
 As each detection module is defined on different scales  ( M1 is defined for the smaller object as compared to M2 ) hence the size of each anchor box would be selected accordingly. M1 will be assigned with smaller anchor boxes as compared M2. The condition for assigning any anchor box to the ground-truth is based on Intersection over Union (IOU). Hence, anchor boxes with IOU greater than 0.5 are called positive anchor boxes and participate in overall loss function.

\subsubsection{UESegNet-2}

\begin{figure}[h]
		
		\centering	
		\includegraphics[scale=0.40]{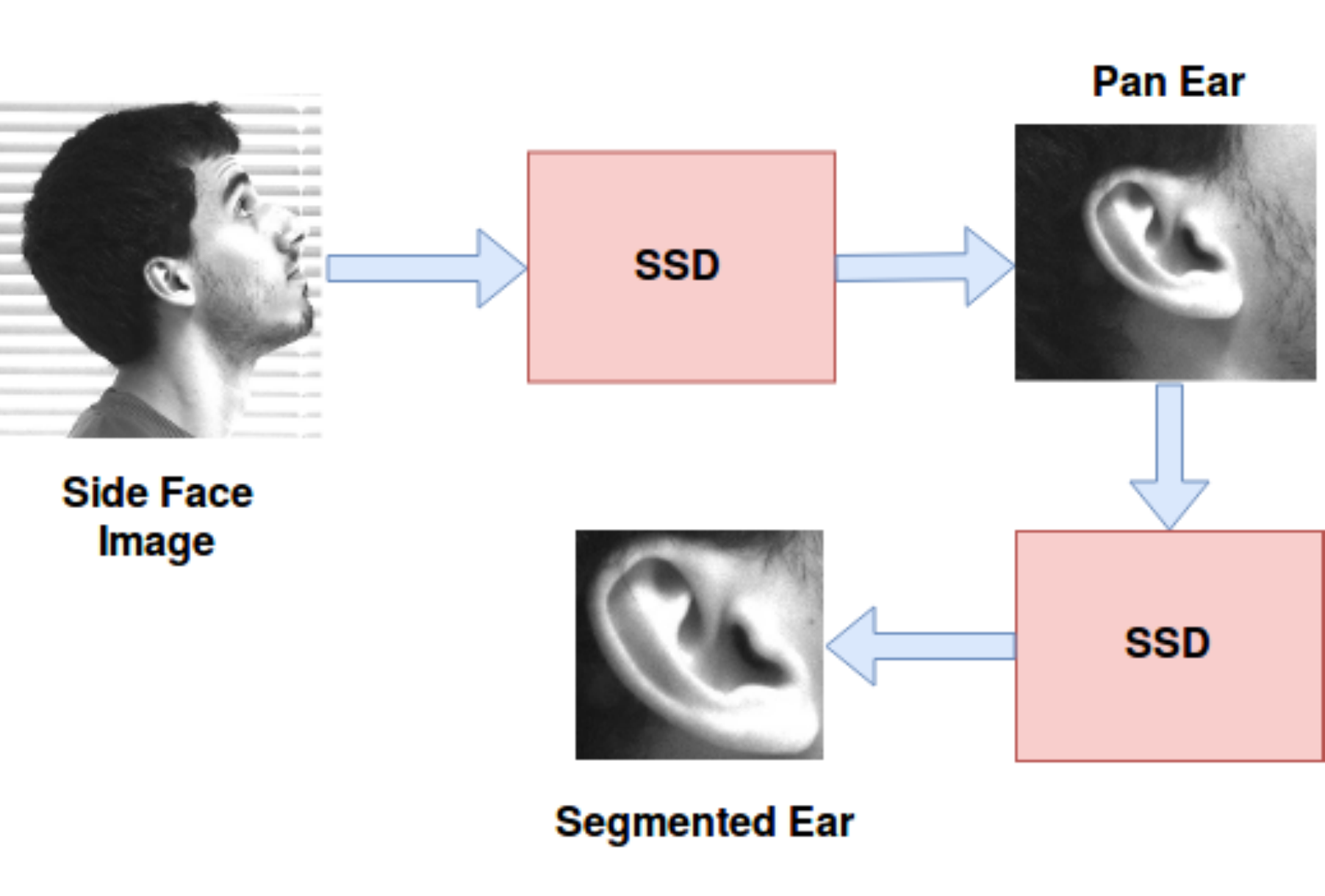}
		\caption{Architecture of UESegNet-2}
        \label{UnESegNet4}
	\end{figure}

The architecture of UESegNet-2 is a two-stage SSD \cite{SSD} as shown in Fig. \ref{UnESegNet4}. The context information is very important for any segmentation network, hence we have combined two same networks sequentially for the same. Initially, we have trained the first network on original images. Further, all the training images are tested on this network for the prediction of bounding boxes. Afterward, we have generated data for the second network by increasing the size of all predicted bounding boxes by 50 pixels in each direction to include the context information. These new predicted boxes have been used as the input for the second network, and ground truths are changed accordingly. Afterward, the second network is trained for new images. At test time, both the models are combined and giving better performance than a single model.      

\textbf{Training Strategy}: During training the network identify default boxes corresponds to their ground truth boxes. The model match each $j^th$ ground truth box to the corresponding $i ^th$ default box and consider boxes with Intersection Over Union $(IOU>0.5)$ and calculate a matrix $x_{ij}$ as follows:

\begin{enumerate}
\item If $IOU > 0.5, x_{ij} = 1$
\item Else, $x_{ij} = 0$

\end{enumerate}

\begin{figure}[h]
		\centering	
		\includegraphics[scale=0.40]{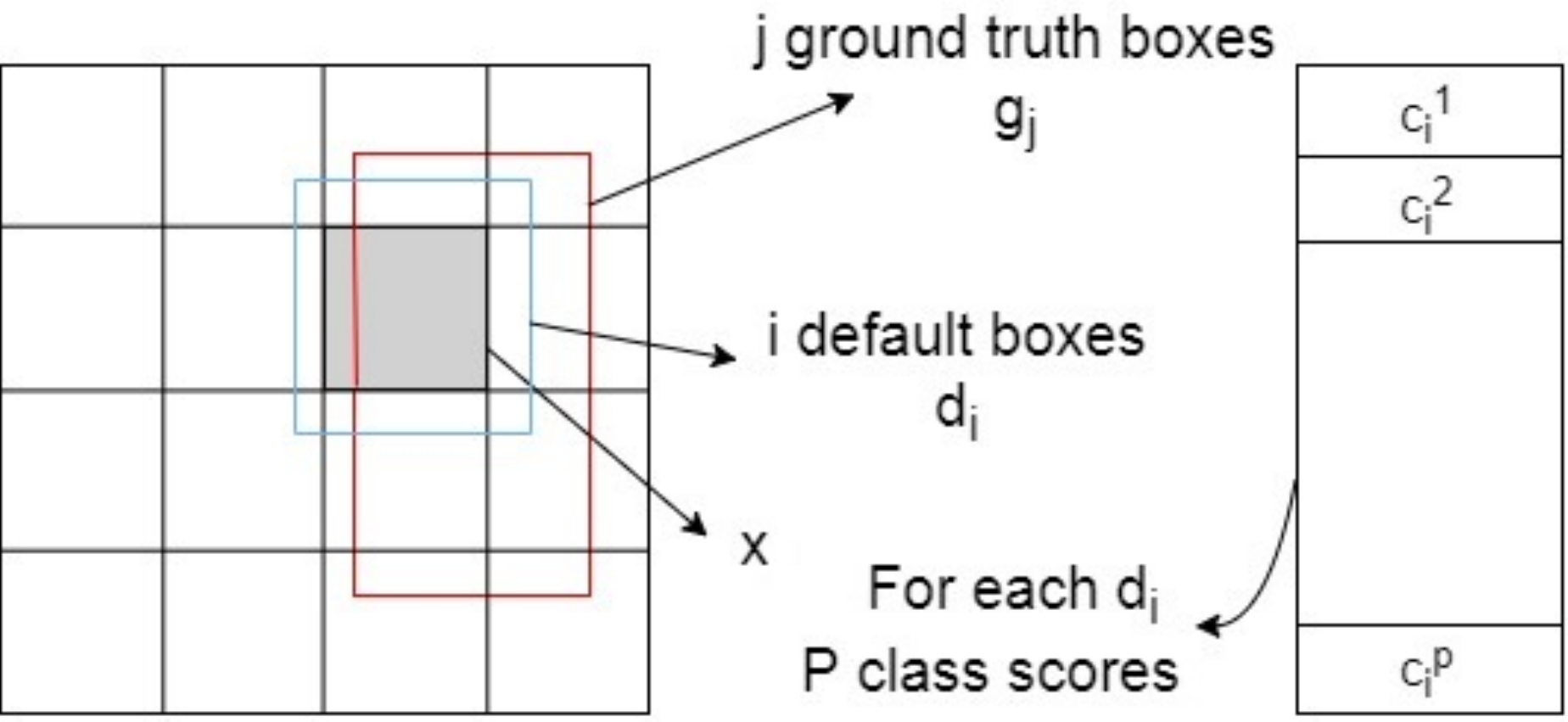}
		\caption{Box prediction }
        \label{box}
	\end{figure}

For each ground truth different default boxes are selected  based on varying location, aspect ratio, and scale. The model predicts bounding box $l_{i}$ having 4 parameters $c_x$, $c_y$, $w$ and $h$ for every default  box $d_{i}$ and also predicts p class scores  as shown in Fig. \ref{box}.

Here,
    $c_x = $Centre $x$ coordinate of predicted box  \\ 
    $c_y = $Centre $y$ coordinate  of predicted box \\
    $w = $width  of predicted box \\
    $h = $height of predicted box \\
    $c_{i}^{p}$= Confidence score of each class
    
 The output of each cell c would be k $\times$ (c + 4). Here k is number of filters for each cell c, and for each feature map of size m $\times$ n. it provides output feature map of (c + 4) $\times$ m $\times$ n $\times$ k.
In addition, we have taken different hyper parameters such as SGD (stochastic gradient descent), epoch $=$ 100, Initial learning rate $=$ 0.003, Final learning rate $=$ 0.004, weight decay $=$ 0.0004, momentum $=$ 0.8 etc.

\textbf{Loss function of UESegNet-2:} The UESegNet-2 have two  losses:
   1) Regression Loss
   2) Confidence Loss and is calculated using equation (2)
    \begin{equation}
     L(x,c,l,g) = \frac{1}{N}[L_{conf}(x,c) + \alpha L_{reg}(x,l,g)]
     \end{equation}
  
    Here, $N =$ number of boxes having IOU (Jaccard Index $> 0.5$ )\\
         $x =$ pixel under consideration\\
         $c =$ class scores\\
         $l =$ predicted boxes\\
         $g =$ Ground truth boxes\\
         $\alpha$ is a constant weight.
         
 \textbf{Regression loss:} 
The regression loss is a smooth L1 Loss ( as per equation 3) and calculated between ground truth box $g_{j}$ and predicted box $l_{i}$.

%x kij smooth L1 (l i m − ĝ j m )
%\sum_{m\in cx,cy,w,h}

\begin{equation}
L_{reg} =\sum_{i\in Pos}^{N}\sum_{ m\in c_x,c_y,w,h}x_{ij}^{k} smooth_{L1}(l{i}^{m}-\hat{g}_{j}^{m})
\end{equation}

\textbf{Confidence loss:} For each box $i$, we have $p$ confidence scores $c_{i}^{p}$, where, \\
    $c_{i}^{1} =$ Confidence of class 1 \\
    $c_{i}^{2} =$ Confidence of class 2 \\
    $c_{i}^{p} =$ Confidence of class p \\
 
\begin{equation}
L_{conf}(x,c) = -\sum_{i\in Pos}^{N}x_{ij}^{p}log(\hat{c}_{i}^{p}) - \sum_{i\in Neg}log(\hat{c}_{i}^{0})
\end{equation}
 
 Here, $$c_{i}^{p} : \hat{c}_{i}^{p} = \frac{e^{({c}_{i}^{p})}}{\sum_{p}^{}e^{({c}_{i}^{p})}}$$

The model tries to maximize confidence of matched predictions (positive boxes) and minimize the confidence of negative boxes.

\section{Benchmark Datasets used for Ear Detection}

%\subsection{\textbf{Database}}

Researchers have provided various benchmarked datasets for ear based biometric authentication system. In this work, we have used six different datasets as  discussed below:\\

\begin{figure}[h]		
\centering	
		\includegraphics[scale=0.30]{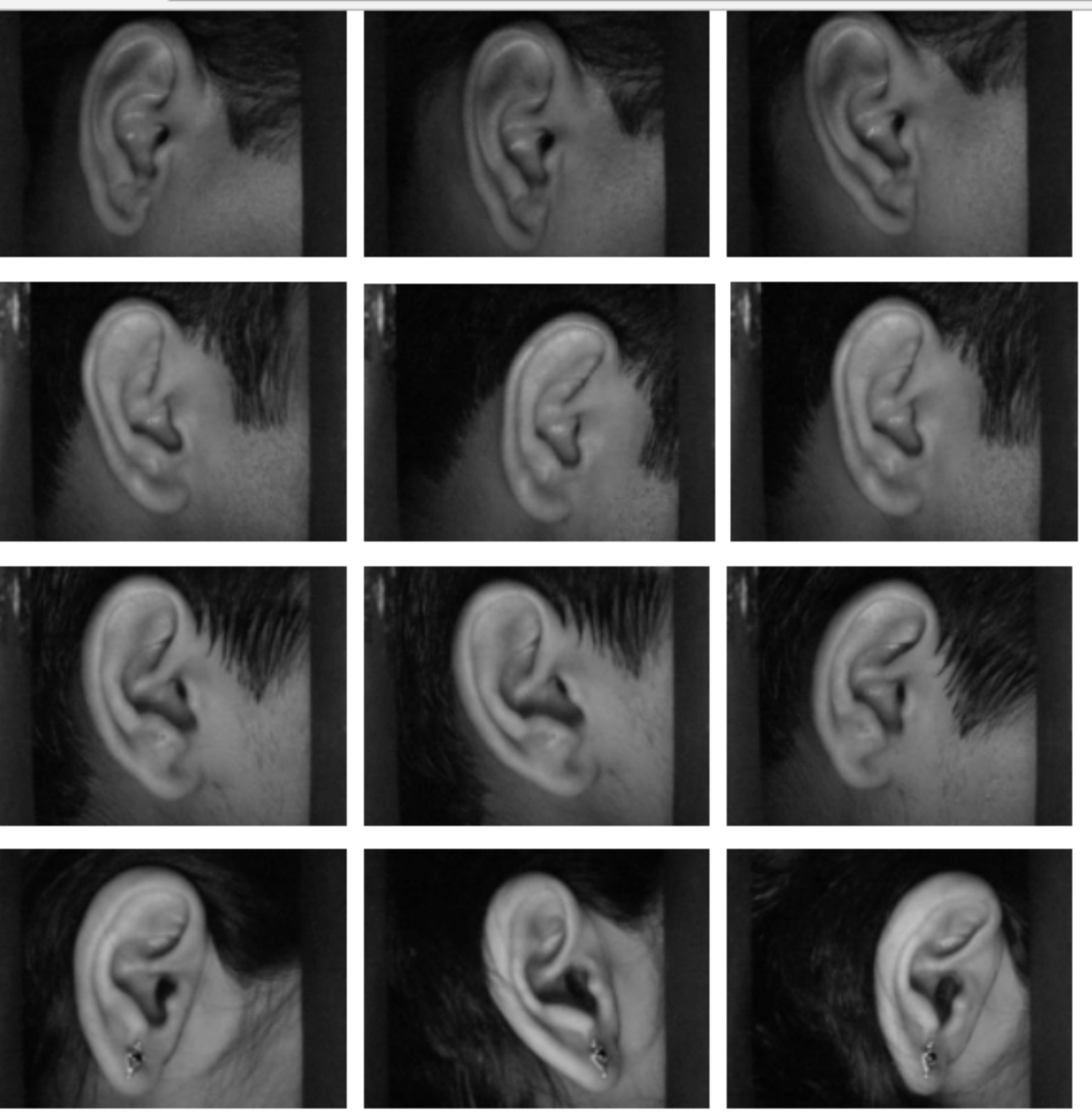}
		\caption{Sample images of IIT Delhi dataset}
        \label{Delhi}
	\end{figure}

\textbf{IITD:} The Indian Institute of Delhi dataset was contributed by  \cite{IITdelhi}, contains ear images of the students and staff at IIT Delhi. The dataset has been acquired during Oct 2006 - Jun 2007, which consist of 121 distinct subjects, and there are three images per subject in gray-scale format. These images were captured in the indoor environment and all the subjects are in the age of 14 to 58 year with slight angle variations. Fig. \ref{Delhi} shows sample images.

\begin{figure}[h]
		\centering	
		\includegraphics[scale=0.35]{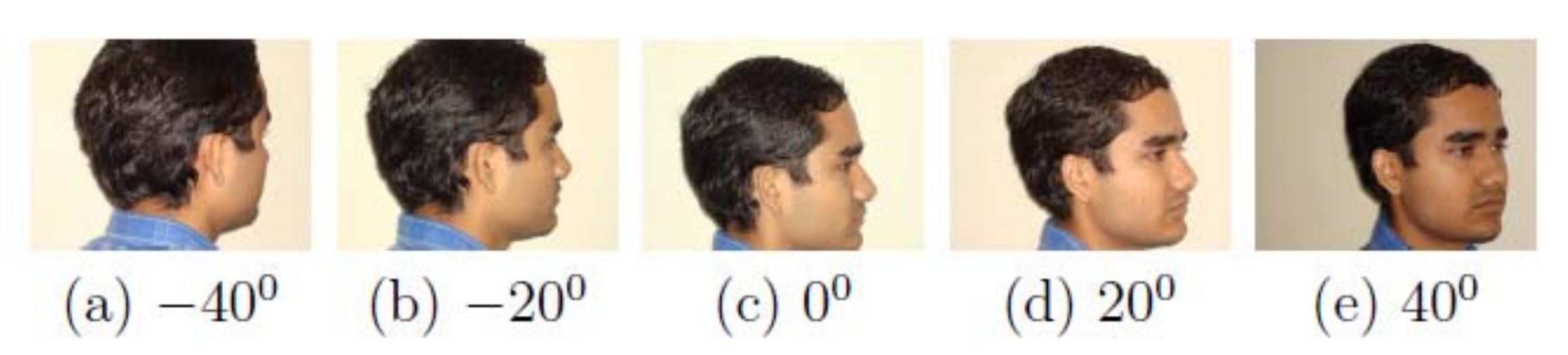}
		\caption{Sample images of IIT Kanpur dataset }
        \label{Kanpur}
	\end{figure}

\textbf{IITK:}
The Indian Institute of Kanpur dataset contributed by \cite{IITK}, contains side face images of 107 unique subjects which have captured at different occlusion (by hairs and earrings) and out of the plane rotations. During acquisition, the camera is moved on the circle and images were captured at different angles, and for each angle position two images are obtained. The dataset has 1070 images, few of them are shown in Fig. \ref{Kanpur}.

\textbf{USTB:} The University of Science and Technology Beijing dataset \cite{USTB}: The dataset has three subsets, USTB-DB1, USTB-DB2, USTB-DB3. In this paper, we have used USTB-DB3 subset as the other two subsets have cropped ears.\\

\begin{figure}[h]
		\centering	
		\includegraphics[scale=0.50]{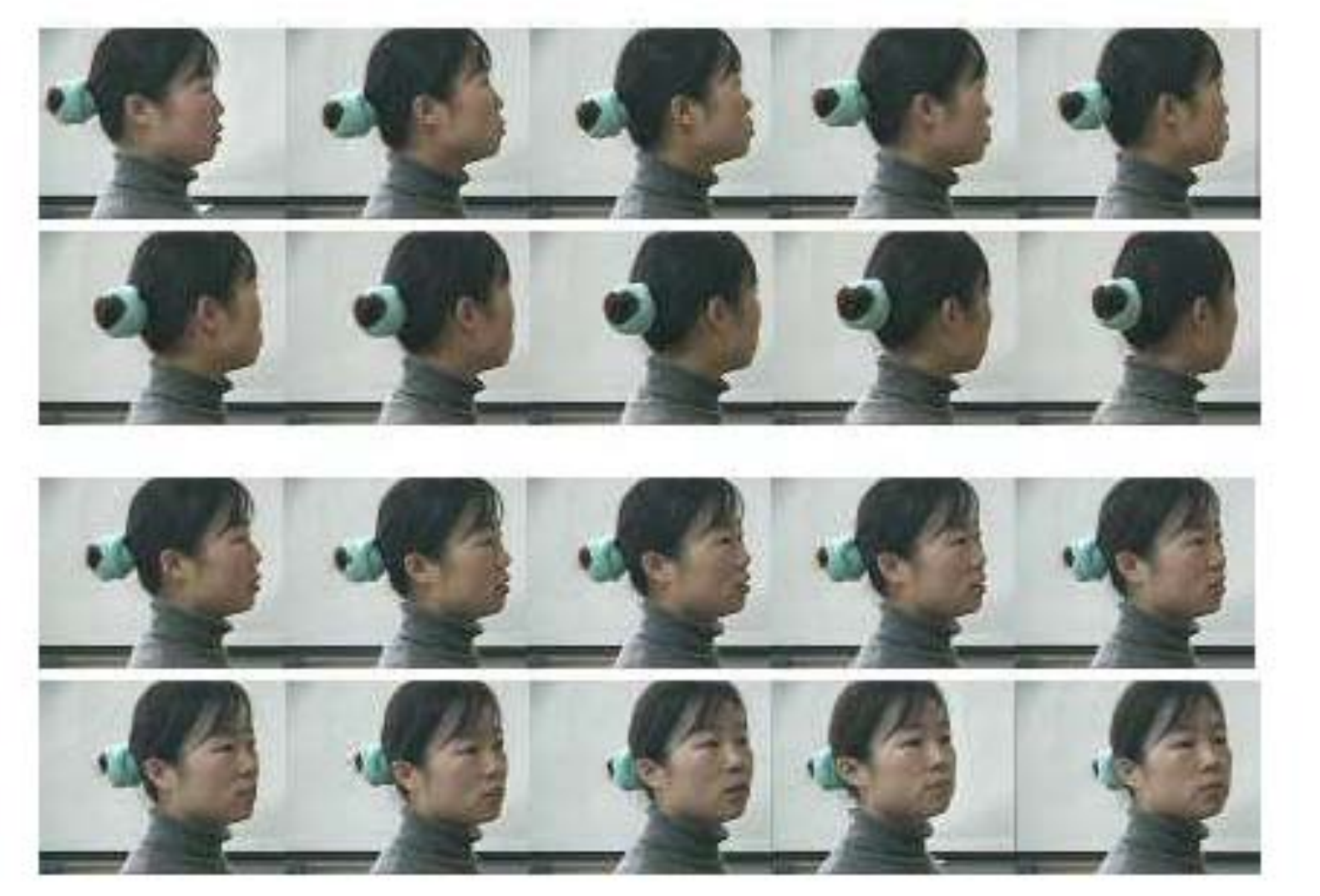}
		\caption{Sample images of USTB-DB3}
        \label{USTB-DB3}
	\end{figure}

  \begin{itemize}  
\item \textbf{USTB-DB3:} This dataset having images of the right side profile face with a resolution of 768*576 for 79 subjects, was captured during Nov 2004 to Dec 2004. These images were captured under various angle variations along with occlusion (by hairs) at a fixed distance of 1.5 meters. Sample images of the database are shown in Fig. \ref{USTB-DB3}.
\end{itemize}

\begin{figure}[h]
		\centering	
		\includegraphics[scale=0.70]{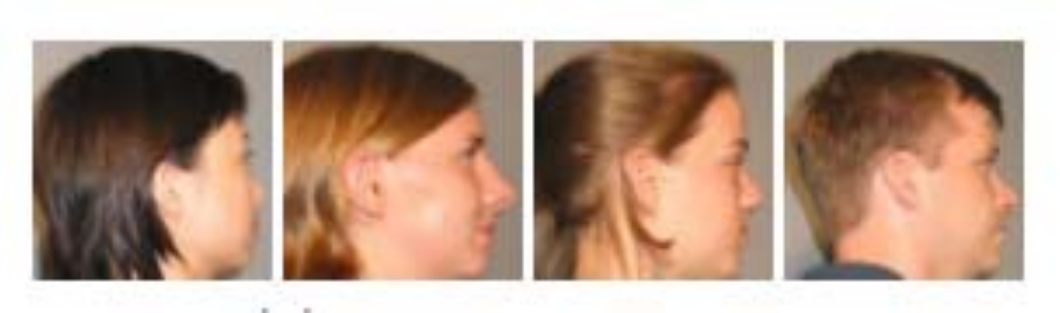}
		\caption{Sample images of UND-E Database}
        \label{UND-E}
	\end{figure}

\begin{figure}[h]
		\centering	
		\includegraphics[scale=0.80]{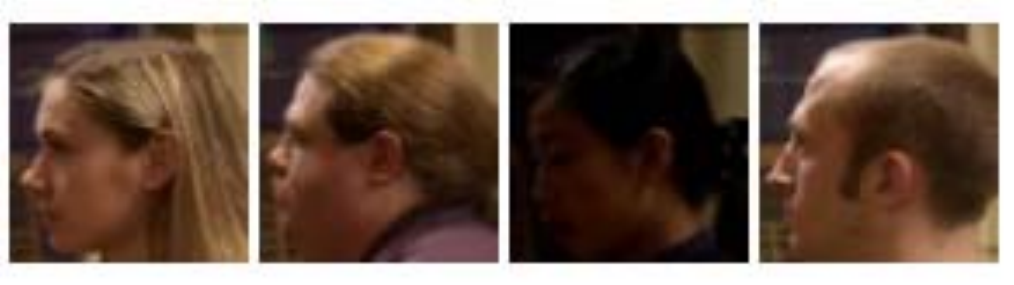}
		\caption{Sample images of UND-J2 Database}       
		\label{UND-J2}
	\end{figure}

\textbf{UND Dataset} The University of Notre Dame (UND) dataset  created for academic research in ear recognition \cite{UND}, acquired from 2002 to 2005 and has many collections. These side face images are captured under varying illumination conditions, partially occluded by ears, and at diverse angle variations. This dataset has four different collections: UND-E, UND-F, UND-G, UND-J2. In this work, we have used the following two collections:

\begin{itemize}
    \item \textbf{UND-E:} This collection has 464 ear images of 114 subjects. Fig. \ref{UND-E} shows some sample images. 
    
\item \textbf{UND-J2:} This collection has 942 side face images of 302 subjects. Fig. \ref{UND-J2} shows some sample images.
\end{itemize}

\textbf{UBEAR:} University of Beira Interior Ear Dataset having 8000 images, captured from the video sequence of 128 subjects under unconstrained environment \cite{UBEAR}. Further, from each video sequence 
17 frames were selected with a different pose and angle variations, occlusion, illumination variations, and partially captured ears. The images are gray-scale with 1280x960 pixels. A distance of 7m is fixed between subject and camera. Sample images of the database are shown in Fig. \ref{UBEAR}. The database has following two collections:

\begin{figure}[h]
		\centering	
		\includegraphics[scale=0.25]{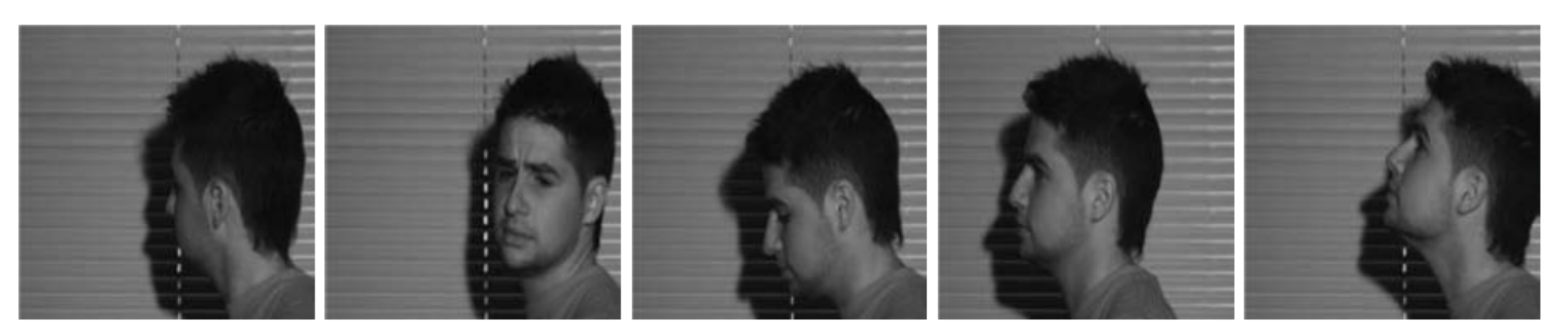}
		\caption{Sample images of UBEAR Database}
        \label{UBEAR}
	\end{figure}
	
\begin{itemize}
    \item \textbf{UBEAR-1:} is a collection of 4412 images captured in an unconstrained environment and ear segmentation mask are also provided.

   \item \textbf{UBEAR-2:} is a collection of 4606 images captured in an unconstrained environment and ear segmentation mask are not provided.
\end{itemize}
The datasets mentioned in this paper contains images captured under varying illumination conditions, occluded by earrings and hairs, images of different size and varying scales, images having side face of a person at different angle variations, poor quality images, a person from different age groups, different ethnicity and different nationality.

    \begin{table}[ht]
		\caption{Benchmarked databases for ear recognition} 
		\centering
	\begin{tabular}{|p{0.6cm}|p{1.7cm}|p{0.8cm}|p{1cm}|p{2.4cm}|}

    		\hline 	
	 Sr.No. & Database & Total Images & Subjects & Environment Condition\\ \hline 
	
1.& IIT Delhi &	471	& 121 & Cropped ear images captured under indoor environment \\ \hline

2.  & IIT Kanpur &	1070 &	107 &	Profile face images at various scales and angle \\ \hline

3. & USTB-DB3 &	651 & 79	& Face images captured under different angles and occlusion \\ \hline

4.  & UND Collection-E  &	464  &	114	  & Side Face Images at varying pose and illumination condition \\ \hline
5.  & UND Collection-J2  & 2414  &	415	  & Side Face Images at various angle rotation and illumination variations, partial occlusion \\ \hline

6. & UBEAR-1 & 4412 &	126 &  Side Face Images at diverse angles and occlusion by hairs and earrings. Ear segmentation mask are provided \\ \hline

7. & UBEAR-2 & 4606 &	126 &  Side Face Images at diverse angles and occlusion by hairs and earrings. Ear segmentation mask are not provided \\ \hline

	 \end{tabular}
 	
    \end{table} 
    
 \section{Testing Protocol and Performance Evaluation Parameters}

As our models are based on deep learning, and they require data in  huge amount for full model learning. In this work, we have collected data from six different benchmarked ear databases mentioned in Table I with an approximate of 14396 images. From each database, 50\% (Approximately: 7100) of the images are used for training models and remaining 50\% of the images of individual database are used to test the performance of different models. As we have only less number of images for training, hence we have performed horizontal flipping, rotation and blurring to increase the training data. Even after the data augmentation on images, one cannot train deep convolution neural network like VGG-16 from the scratch as the network requires million of images to train, hence we have used pre-trained weights\cite{VGG} of VGG, which has been trained on 1.8 million images of different categories in ILSVRC-2014 competition.

To measure the performance of the ear localization model there are standard parameters: (Intersection Over Union, Accuracy, Precision, Recall and F1-Score), which are discussed in detail as below:

\textbf{1. Intersection Over Union (IOU):} is a very crucial parameter to evaluate the accuracy of any object detection model and is calculated using equation (5).

\begin{equation}
\\IOU=\frac{G\cap P}{G\cup P}
\end{equation}

a) Ground truth bounding boxes (G): These boxes are manually drawn on test images to specify where the object is located in the image. 

b) Predicted bounding boxes (P): These are the boxes predicted by the model on test images.

Here $G\cap P$ is the intersection area between ground truth and predicted bounding box.
$G\cup P$ is the area of union between ground truth and predicted bounding box. The value of IOU ranges from 0 to 1; 0 indicates no overlapping whereas the value 1 indicates complete overlapping between predicted bounding boxes and ground truth boxes. An accurate biometric recognition system needs IOU to score more than 0.8 for perfect matching.

\textbf{2. Accuracy:} It measures the proportion of true results, which is calculated as the ratio between the number of test images with IOU $>i$
(i is a threshold value between 0 to 1) to the total number of test images as per the equation (6).

\textbf{3. Precision:} It is the ratio of true positive bounding boxes predicted by the model to the sum of true positive and false positive bounding boxes based on the ground truth and is calculated as per the equation (7).

\textbf{4. Recall:} It is the ratio of true positive bounding boxes predicted by the model to the sum of true positive and false negative bounding boxes based on the ground truth and is calculated as per the equation (8).

\textbf{5. F1 Score:} It measures the overall evaluation of the system and is calculated as per the equation (9). 

\begin{equation}
\\Accuracy=\frac{TP+TN}{TP+TN+FN+FP}
\end{equation}

\begin{equation}
\\Precision=\frac{TP}{TP+FP}
\end{equation}

\begin{equation}
\\Recall = \frac{TP}{TP+FN}
\end{equation}

\begin{equation}
\\F1-Score=2 * \frac{Precision * Recall}{Precision + Recall} 
\end{equation}

Here, 

TP (True Positive) = These are the images in which ear is correctly detected.

FP (False Positive) = These are the images in which ear is detected mistakenly.

FN (False Negative) = These are the images in which background (non-ear region) is detected as a ear. 

TN (True Negative) = 0, as we have to detect only one object (i.e. ear in an image).

\section{Results and Discussion}

\begin{figure*}[h]
    \centering
    
    \begin{subfigure}[c]{0.32\textwidth}
        \includegraphics[width=1\textwidth]{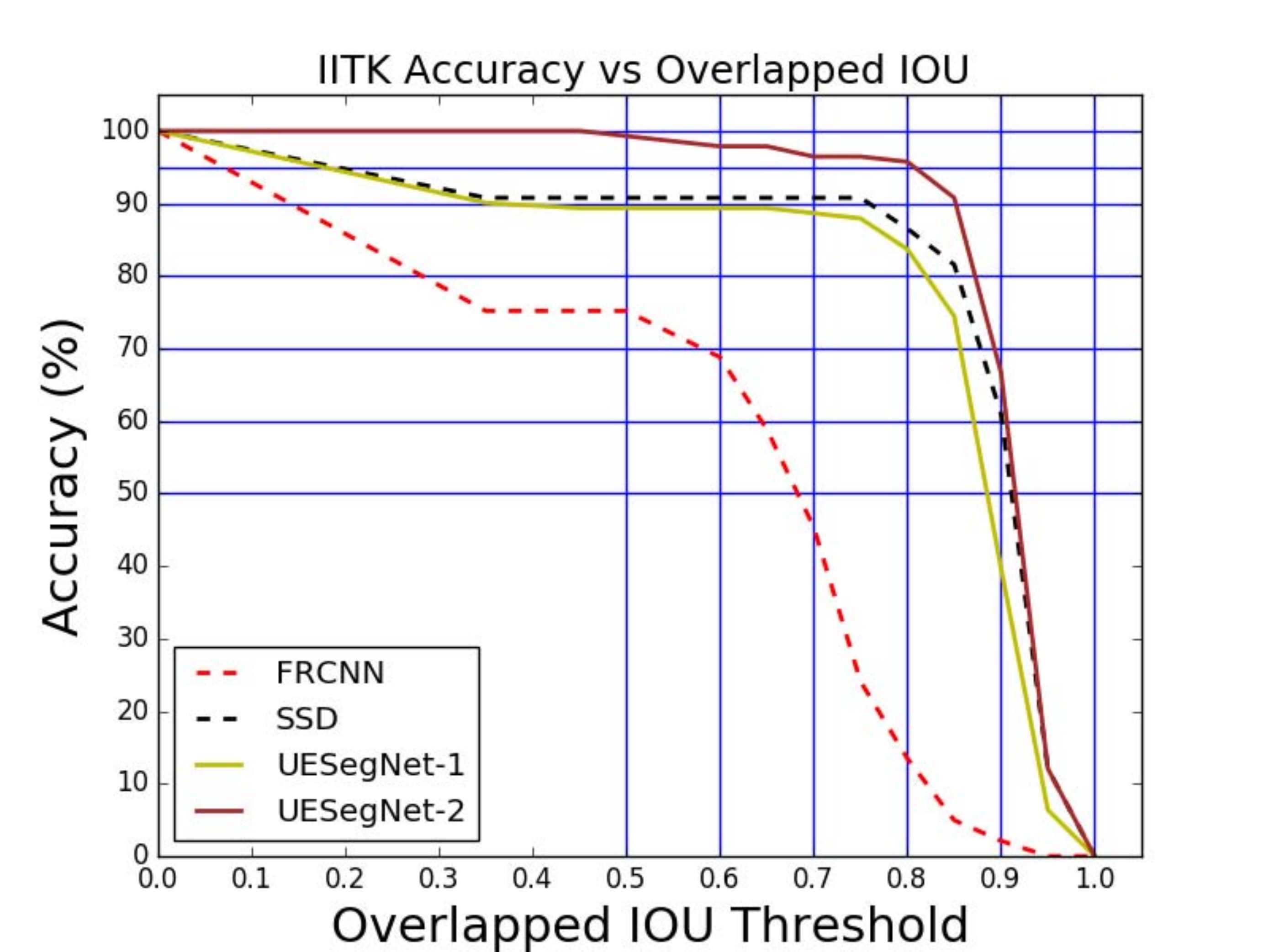}
        \caption{Accuracy on IITK Database }
        \label{IITK-Accuracy}
    \end{subfigure}
    \begin{subfigure}[c]{0.32\textwidth}
        \includegraphics[width=1\textwidth]{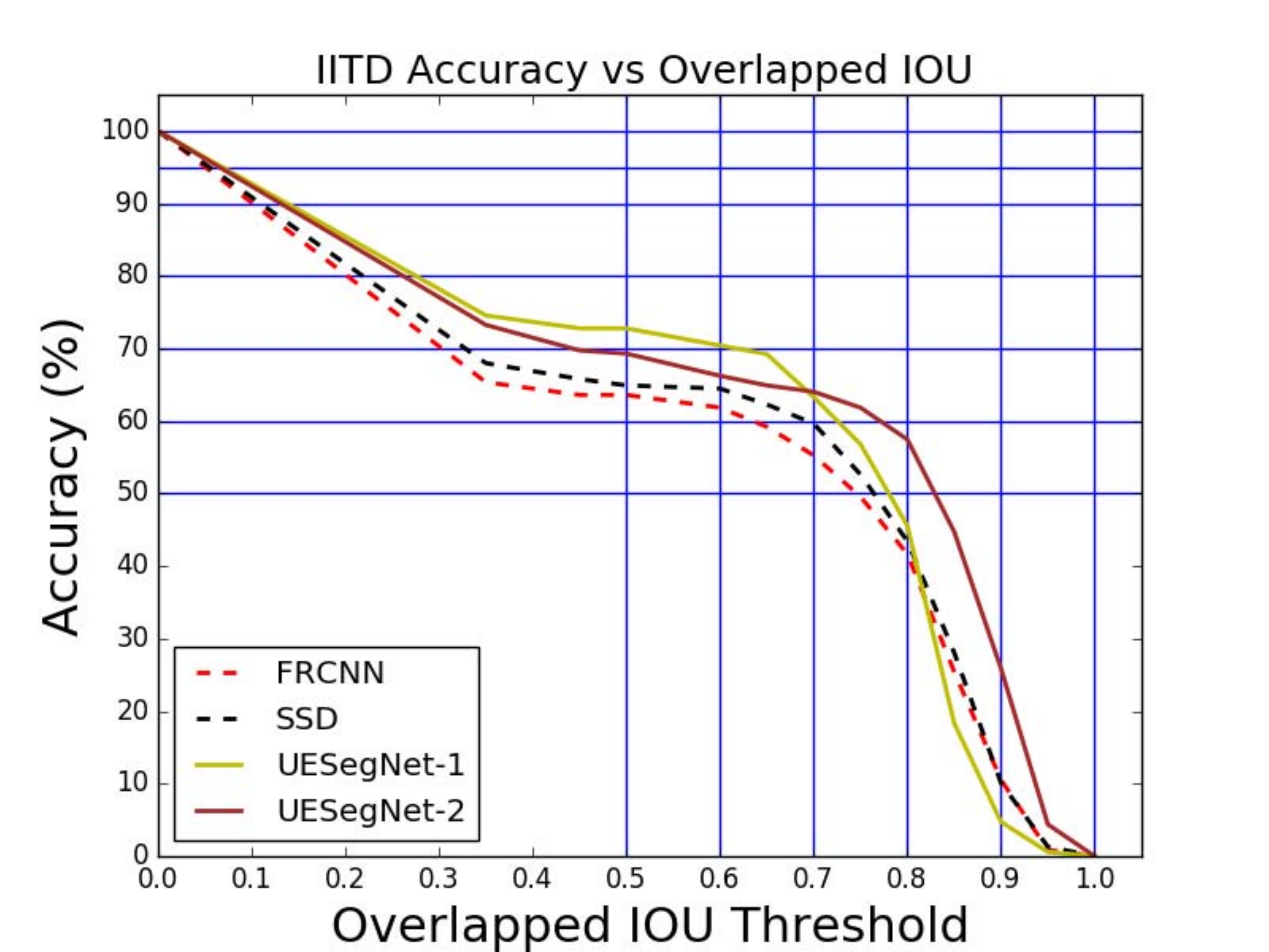}
        \caption{Accuracy on IITD Database}
        \label{IITD-Accuracy}
    \end{subfigure}
    \begin{subfigure}[c]{0.32\textwidth}
        \includegraphics[width=1\textwidth]{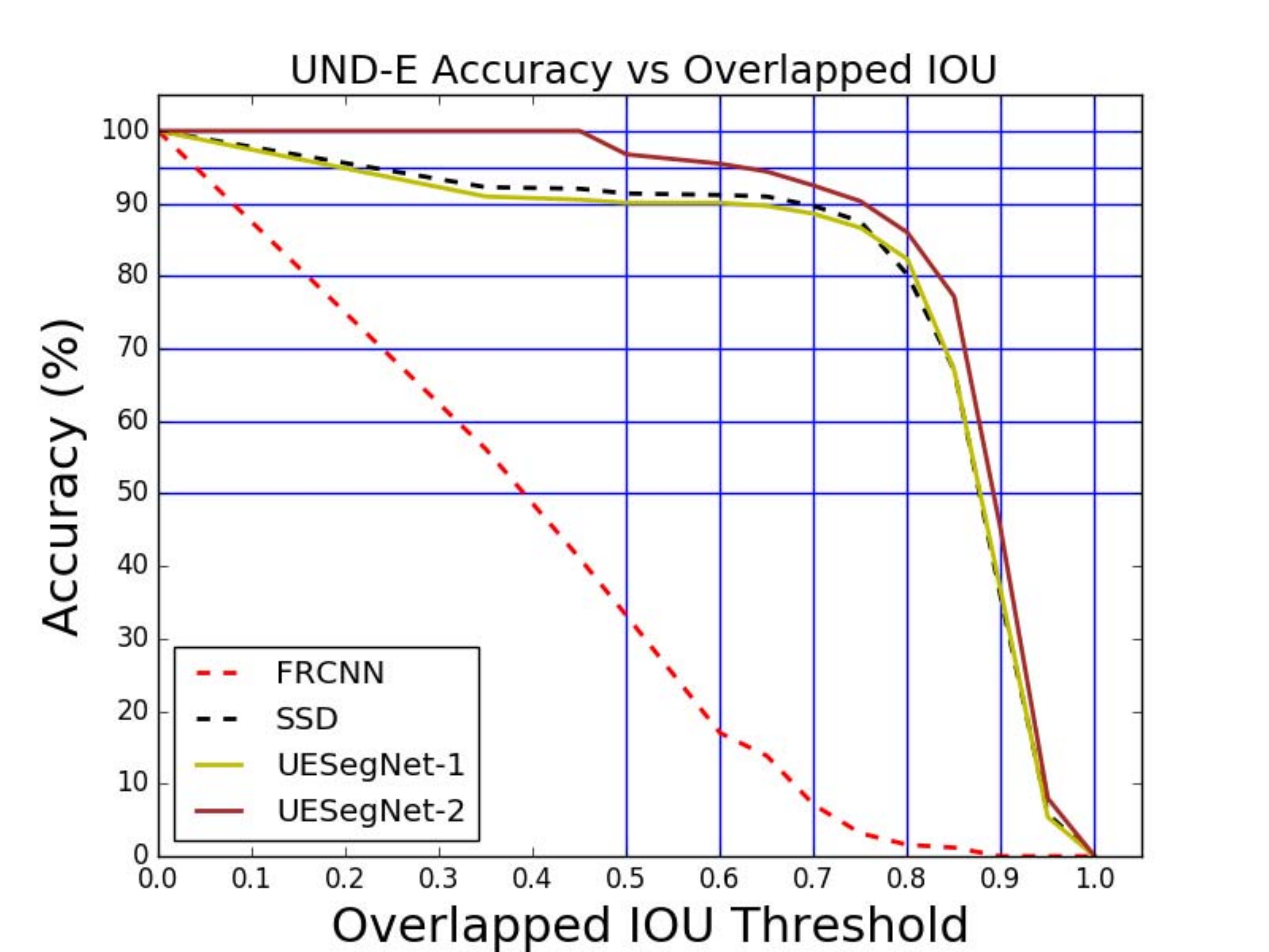}
        \caption{Accuracy  on UND-E Database}
        \label{UND-E-Accuracy}
    \end{subfigure}
    \begin{subfigure}[c]{0.32\textwidth}
        \includegraphics[width=1\textwidth]{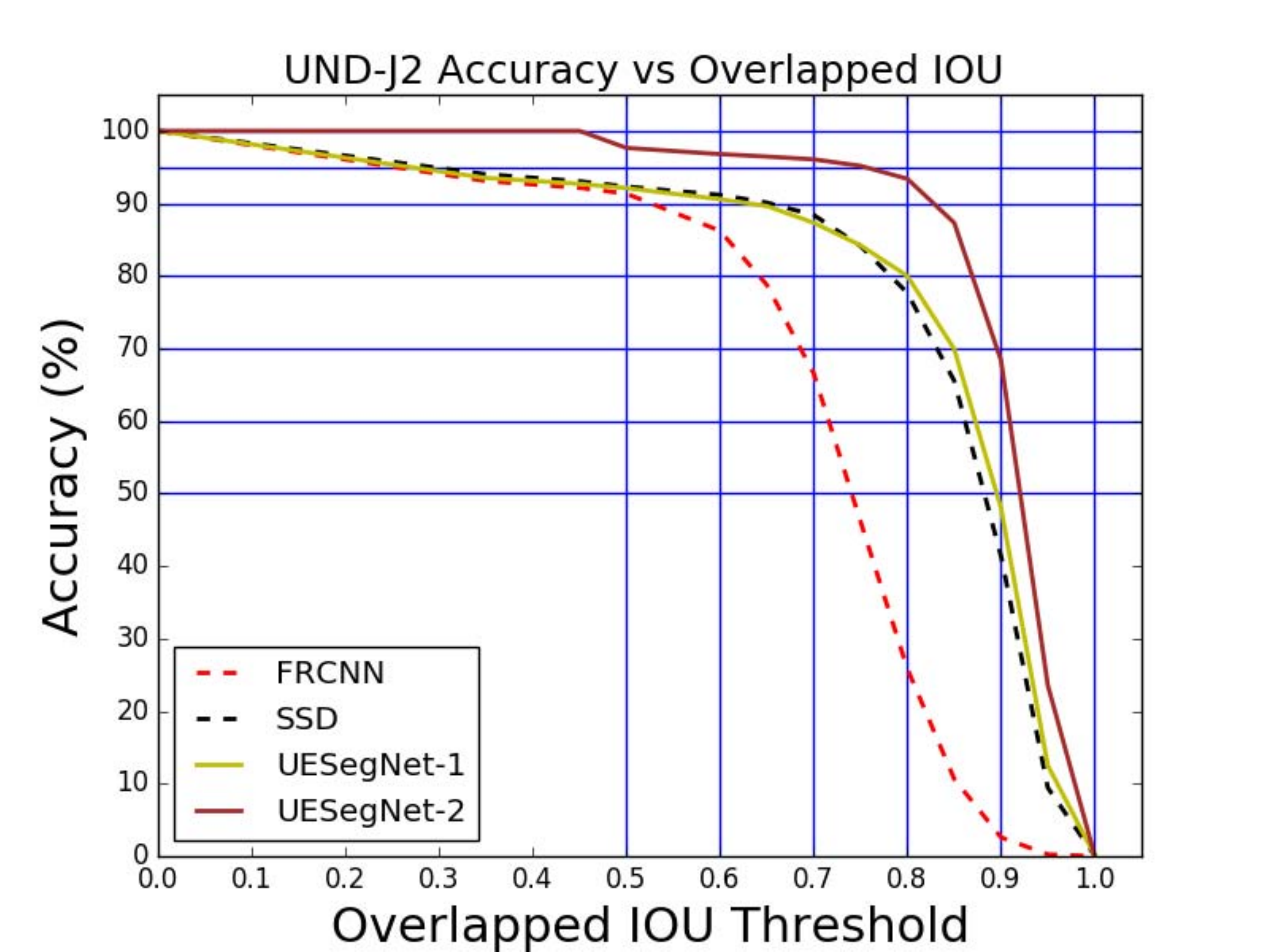}
        \caption{Accuracy on UND-J2 Database}
        \label{UND-J2-Accuracy}
    \end{subfigure}
     \begin{subfigure}[c]{0.32\textwidth}
        \includegraphics[width=1\textwidth]{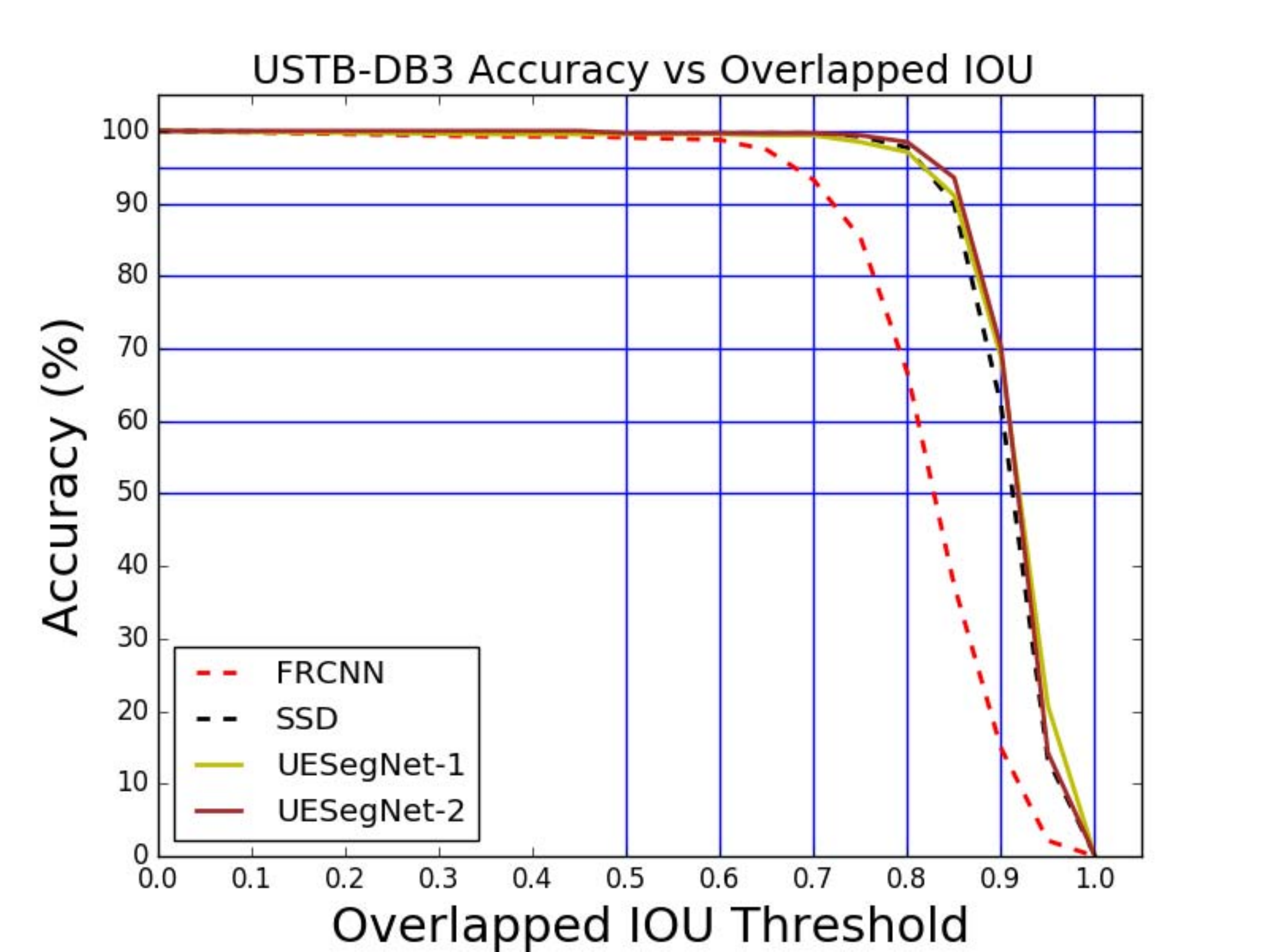}
        \caption{Accuracy on USTB-DB3 Database}
        \label{USTB-DB3-Accuracy}
    \end{subfigure}
     \begin{subfigure}[c]{0.32\textwidth}
        \includegraphics[width=1\textwidth]{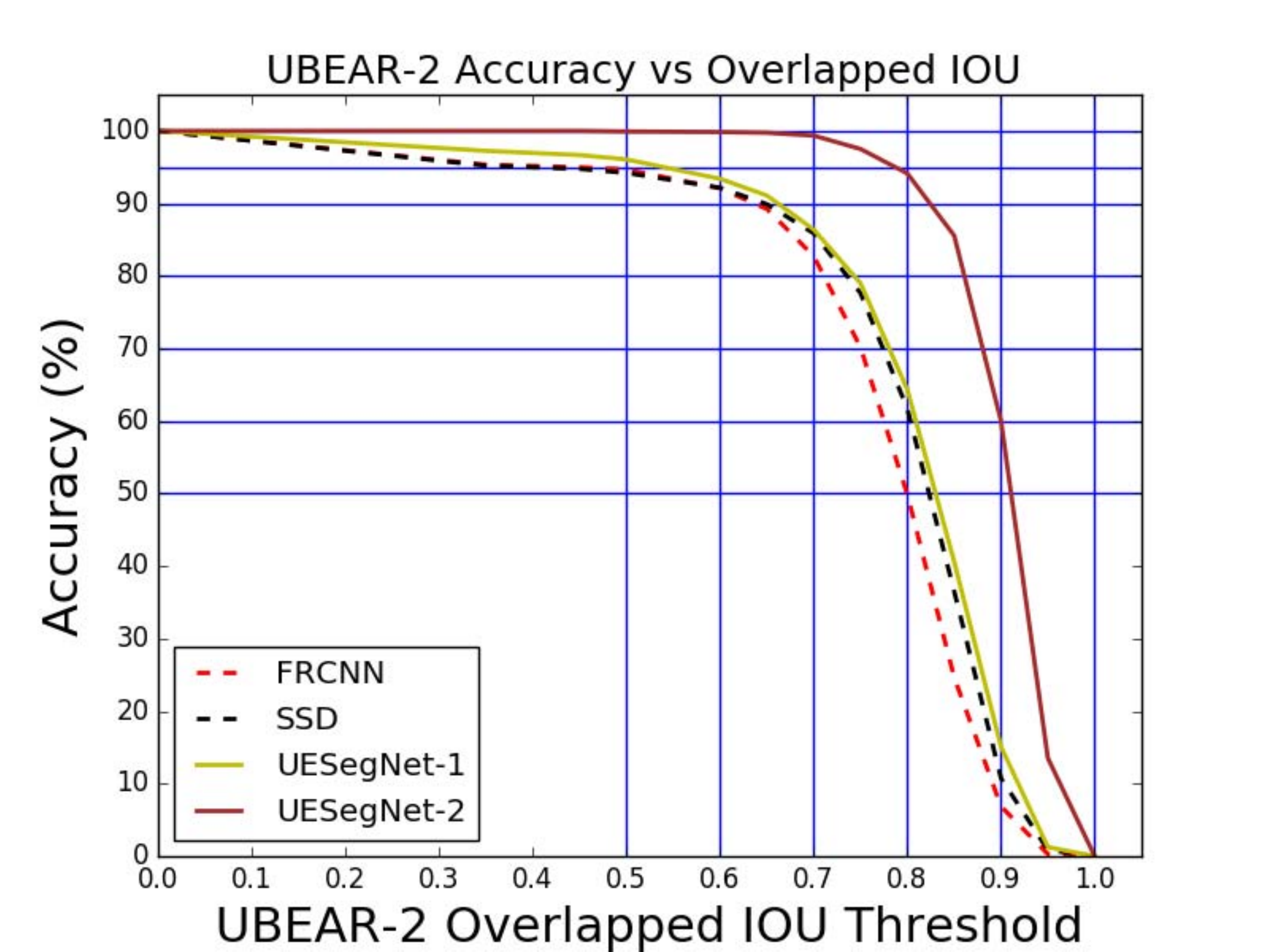}
        \caption{Accuracy on UBEAR-2 Database}
        \label{UBEAR-Accuracy}
    \end{subfigure}
    
    \caption[]{Proposed models performance on individual databases }
    \label{Models-On-Same_Data}
\end{figure*}

\begin{figure*}[h]
    \centering
    
    \begin{subfigure}[c]{0.32\textwidth}
        \includegraphics[width=1\textwidth]{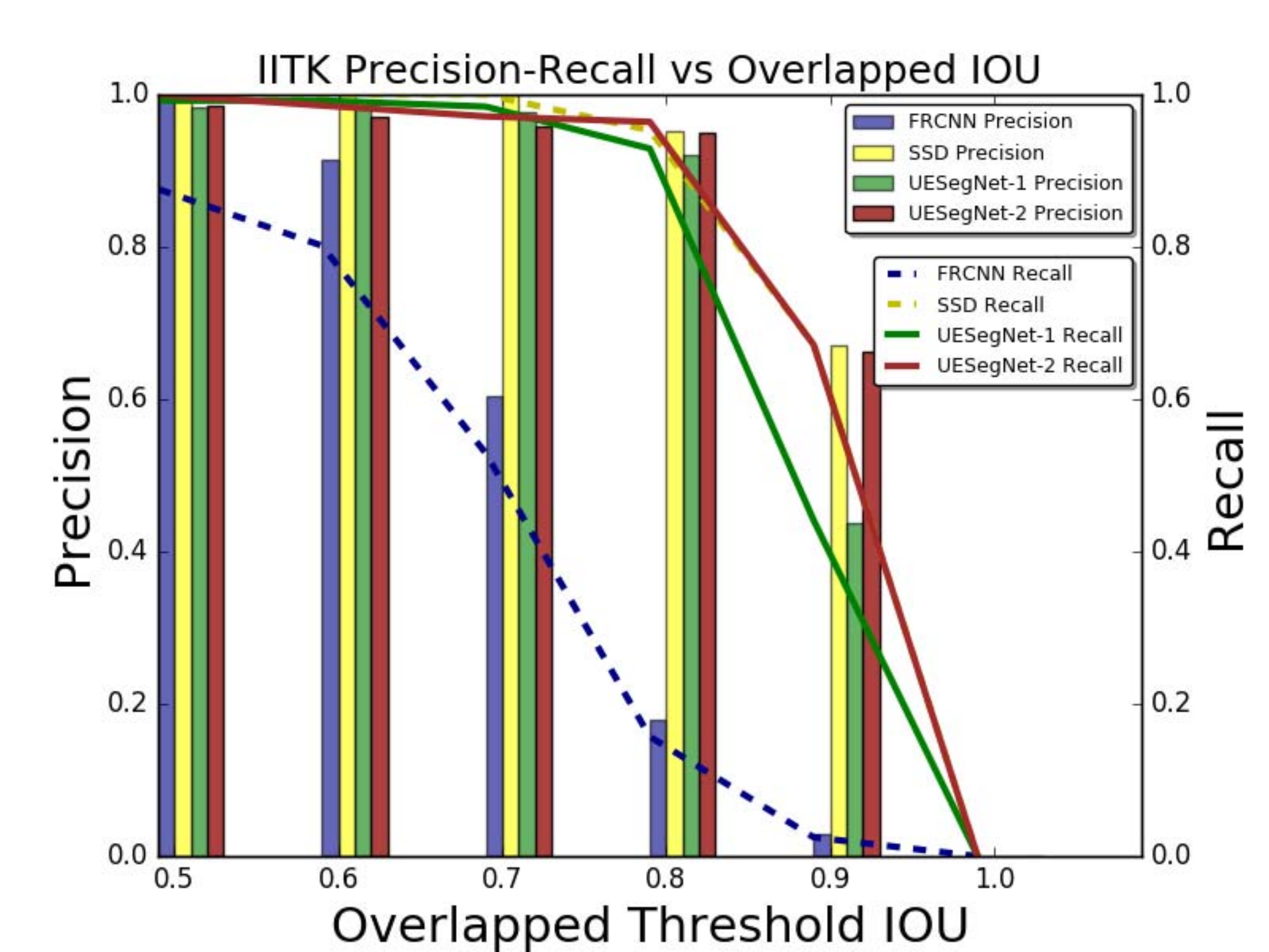}
        \caption{IITK Database }
        \label{IITK-Precision}
    \end{subfigure}
    \begin{subfigure}[c]{0.32\textwidth}
        \includegraphics[width=1\textwidth]{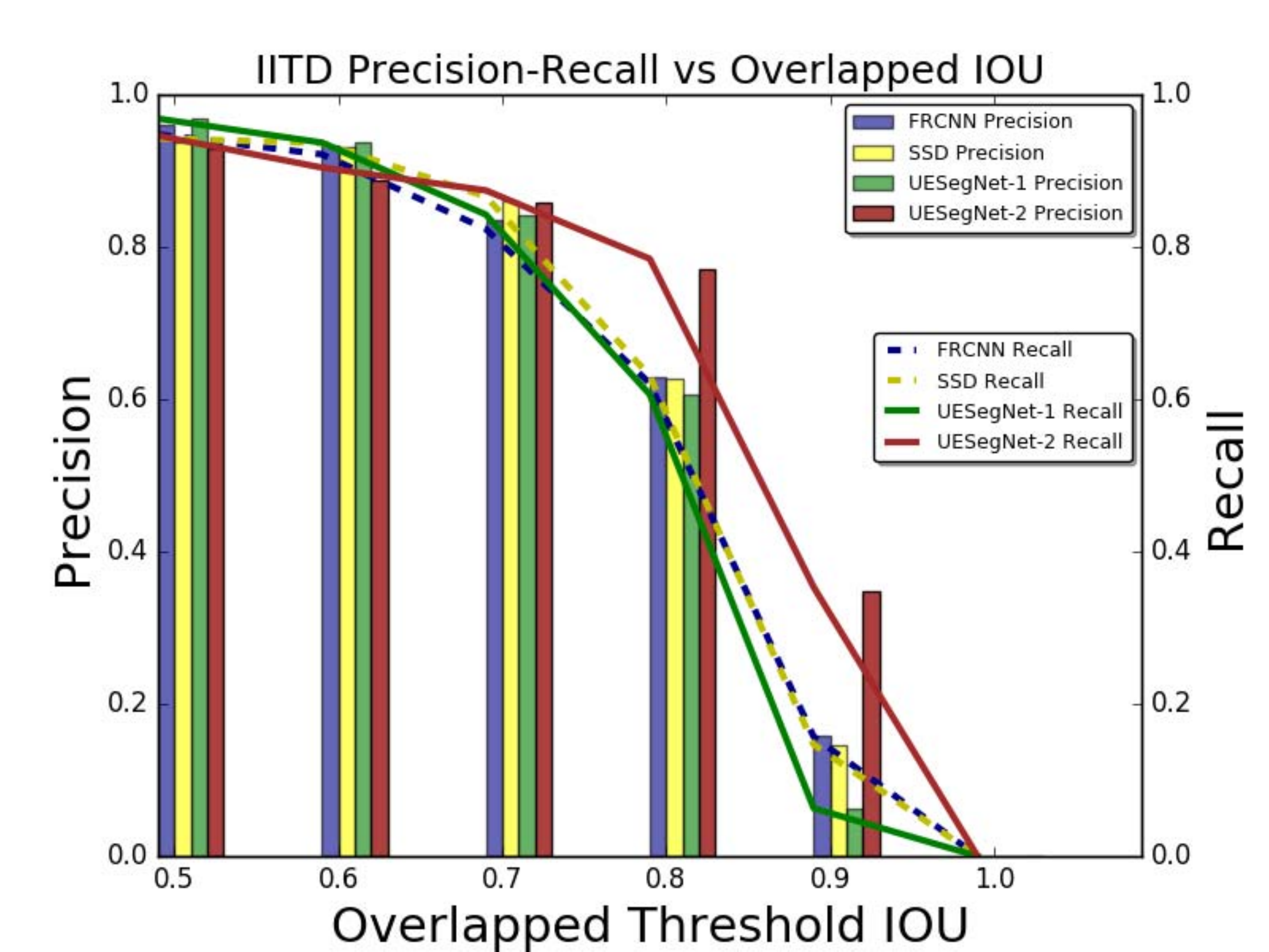}
        \caption{IITD Database}
        \label{IITD-Precision}
    \end{subfigure}
    \begin{subfigure}[c]{0.32\textwidth}
        \includegraphics[width=1\textwidth]{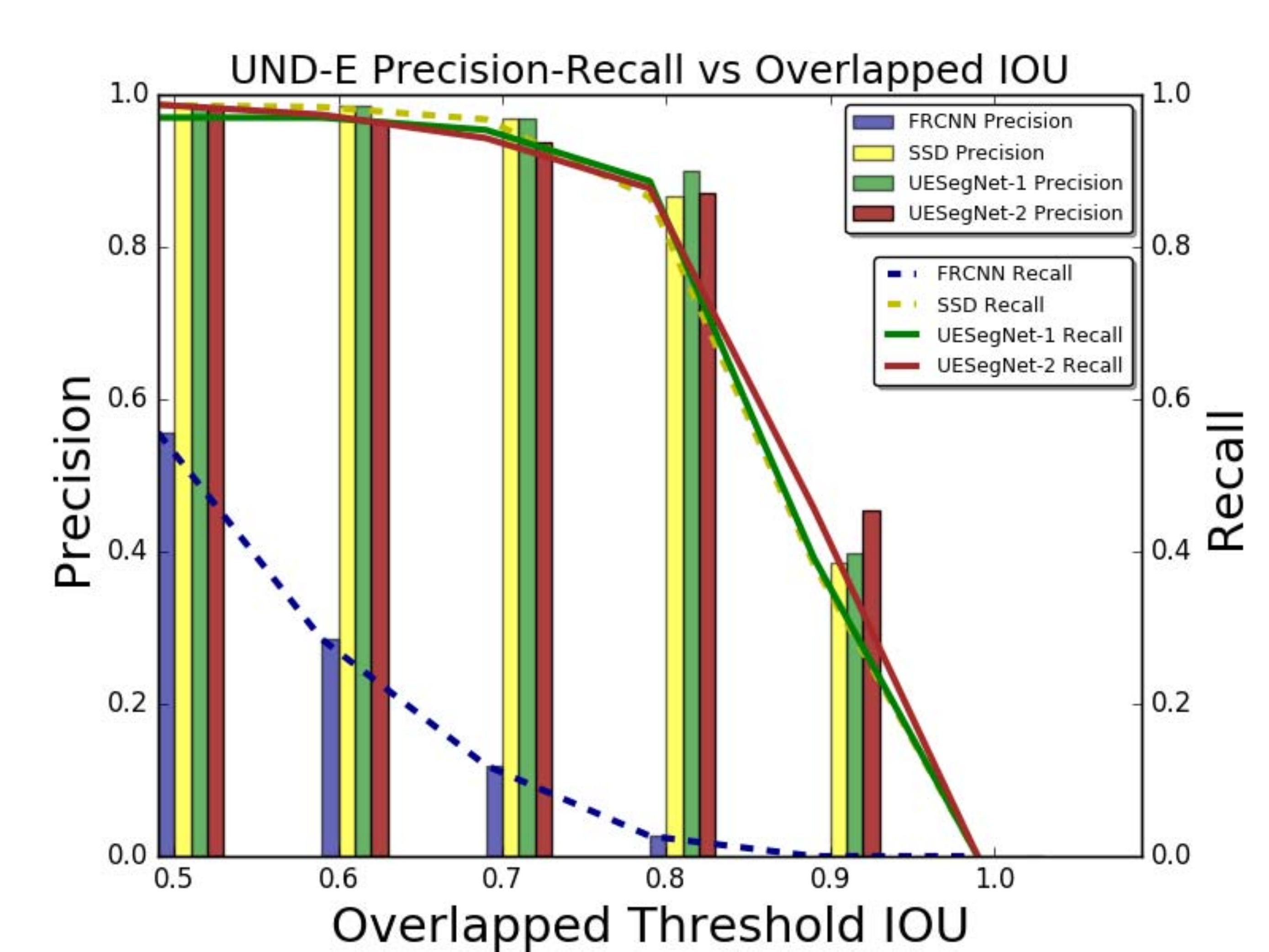}
        \caption{UND-E Database}
        \label{UND-E-Precision}
    \end{subfigure}
    \begin{subfigure}[c]{0.32\textwidth}
        \includegraphics[width=1\textwidth]{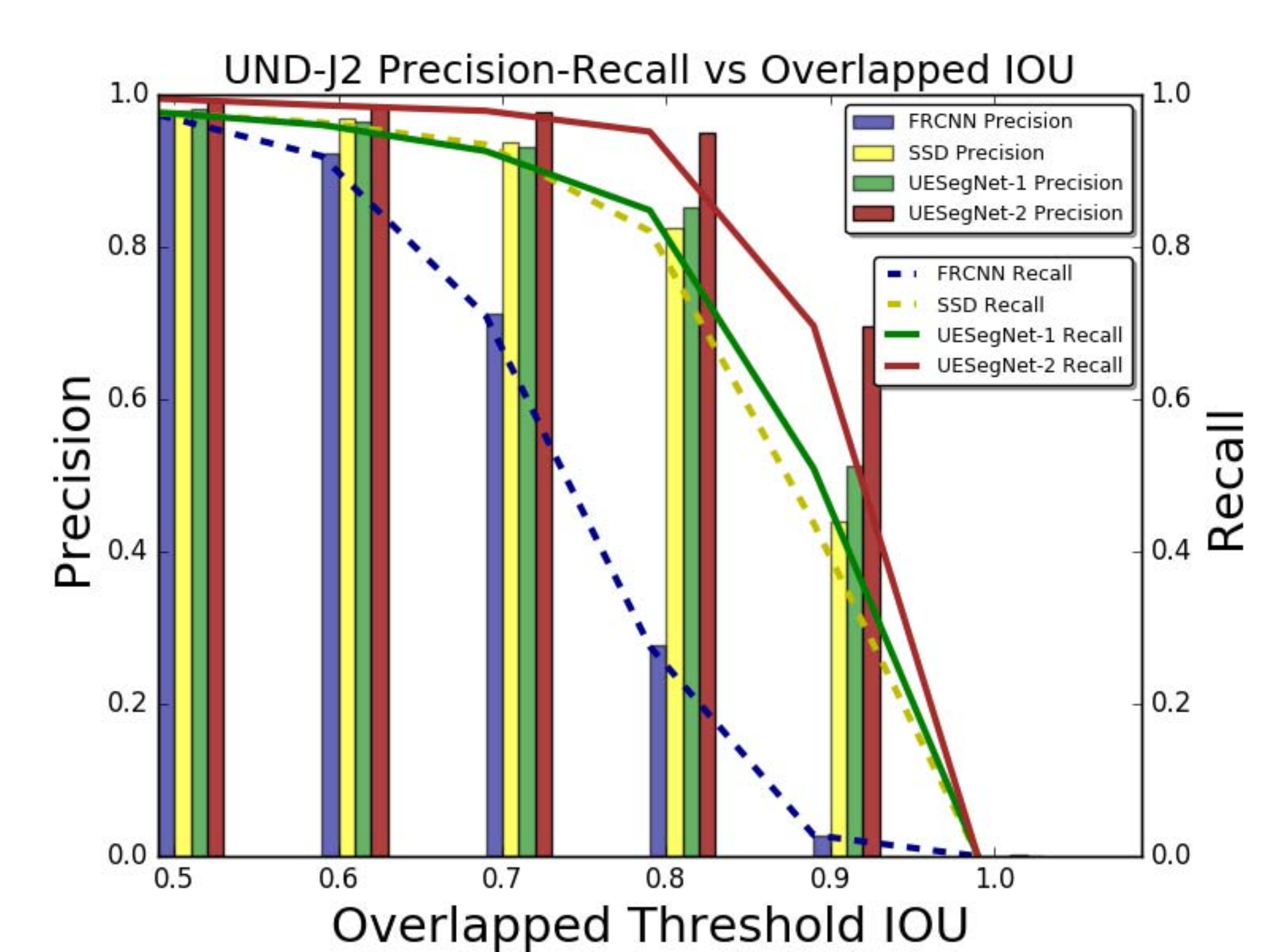}
        \caption{UND-J2 Database}
        \label{UND-J2-Precision}
    \end{subfigure}
     \begin{subfigure}[c]{0.32\textwidth}
        \includegraphics[width=1\textwidth]{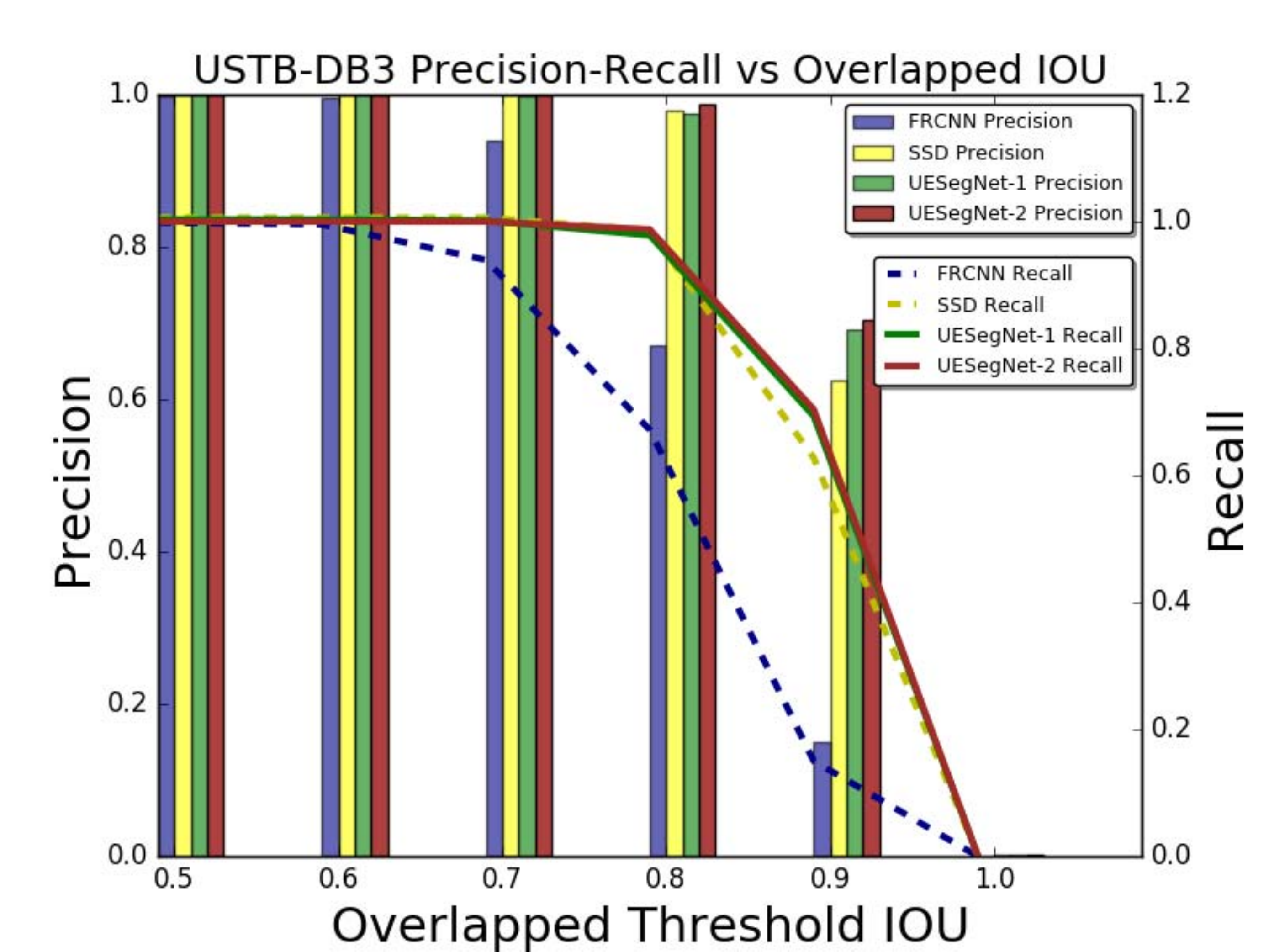}
        \caption{USTB-DB3 Database}
        \label{USTB-DB3-Precision}
    \end{subfigure}
     \begin{subfigure}[c]{0.32\textwidth}
        \includegraphics[width=1\textwidth]{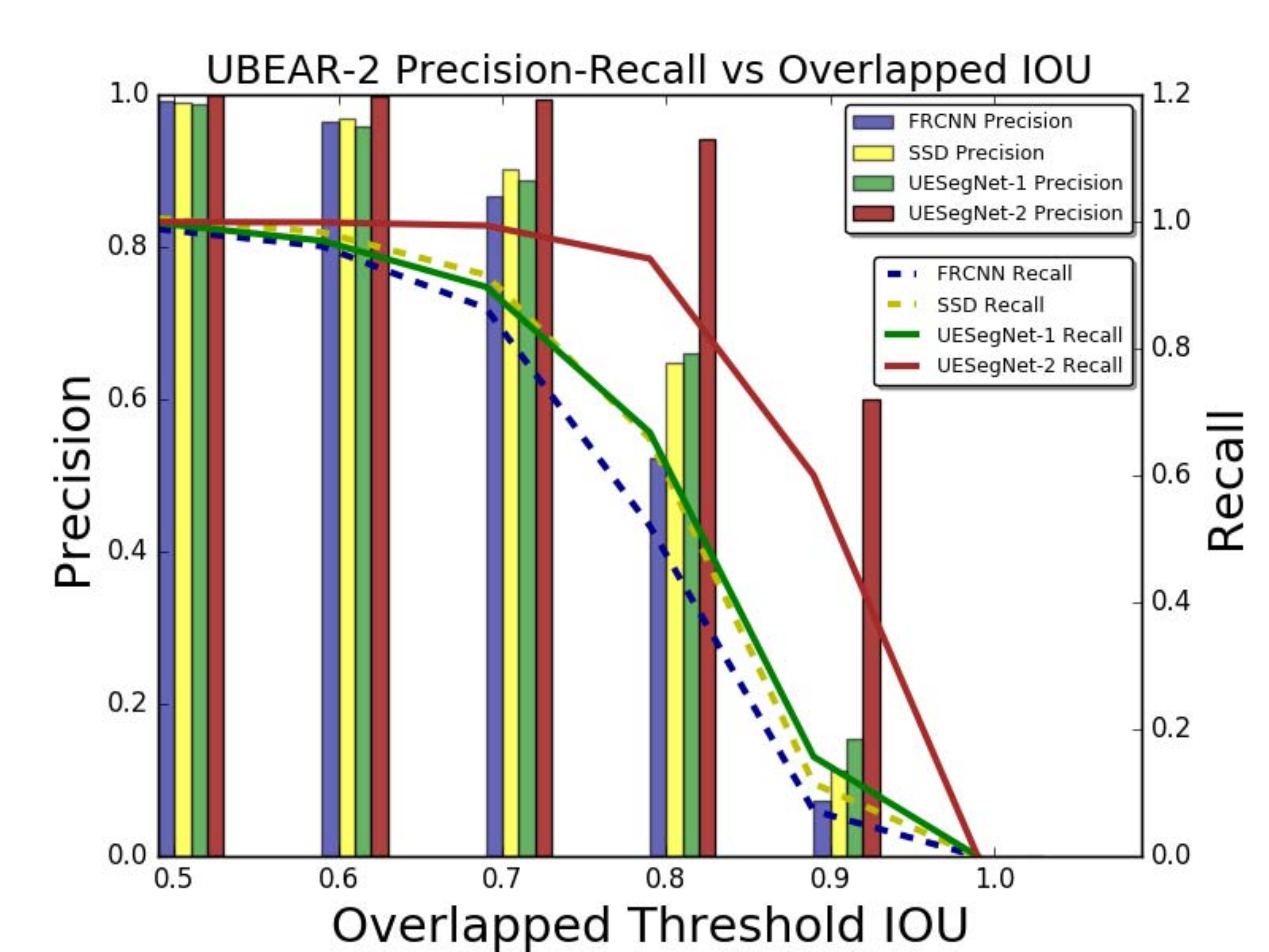}
        \caption{UBEAR-2 Database}
        \label{UBEAR-Precision}
    \end{subfigure}
    
    \caption[]{Proposed models Precision and Recall on individual databases }
    \label{Precision-On-Same_Data}
\end{figure*}

In this section, the performance of models is tested on different databases and various graphs for performance parameters are plotted and shown in Fig. \ref{Models-On-Same_Data} and Fig. \ref{Precision-On-Same_Data} respectively. Moreover, the results of the models are shown in Table II at different values of IOUs.

\subsection{Performance Comparison of Models on Individual Database }

\textbf{Performance on IITK database:} As shown in Fig. \ref{IITK-Accuracy}, it has been observed that at IOU=0.5, the accuracy of all models stays above 90\% except FRCNN and the maximum accuracy is obtained by UESegNet-2, which is 99\%. From IOU=0.6 to 0.7, the performance of FRCNN drops significantly from 70\% to 50\% but the accuracy of UESegNet-1, SSD, UESegNet-2 stays above 89\%. At an IOU=0.8 the UESegNet-2 has obtained maximum accuracy of 95.74\% while the accuracy of FRCNN, SSD, UESegNet-1, drops to 13.48\%, 86.52\%, 83.69\% respectively. The precision and recall values on this database are shown in Fig. \ref{IITK-Precision}, and the model UESegNet-2 have better results at higher values of IOU.

\textbf{Performance on IITD database:} As displayed in Fig. \ref{IITD-Accuracy}, it has been observed that the accuracy of all models is less among all the databases. This may be due to the size of images in the database, as it has cropped ear images having size 272*204. Since the image size is very small, it becomes very difficult to localized ear at this scale. The maximum accuracy is obtained by UESegNet-1, at IOU=0.5 it has achieved an accuracy of 72\%. However, the performance of all the models decreases significantly for higher values of IOUs. The Fig. \ref{IITD-Precision} shows the precision and recall values and for our proposed model UESegNet-2 it stays higher than other models.

\textbf{Performance on UND-E database:} It has been observed that the accuracy for all models stays more than 90\% till an IOU=0.6, except the FRCNN as it performs very poorly due to the less images of this database as shown in Fig.\ref{UND-E-Accuracy}. The UESegNet-2 has obtained maximum accuracy of 95.47\% for IOU=0.6. At an IOU=0.8, the accuracy for UESegNet-2 and UESegNet-1 stays above to 83\%, but for SSD it drops to 80\%. The precision and recall values are shown in Fig. \ref{UND-E-Precision} and our proposed models UESegNet-1 and UESegNet-2 get better results than existing models.

\textbf{Performance on UND-J2 database:} On this database the accuracy of all models remains above 90\% till an IOU=0.5. However, the UESegNet-2 has obtained maximum accuracy of 98\% at IOU=0.5 as shown in Fig. \ref{UND-J2-Accuracy}. However, at IOU=0.6 the performance of FRCNN slightly decreases to 86.23\%, while for other models it stays above 90\%. At an IOU=0.8, the UESegNet-2 has obtained maximum accuracy of 93.39\%, whereas the accuracy for SSD, UESegNet-1, and FRCNN drops to 77.65\%, 80\%, 25.84\% respectively. The Fig. \ref{UND-J2-Precision} shows the precision and recall values and they are higher for our proposed models.

\textbf{Performance on USTB-DB3 database:} As displayed in Fig.  \ref{USTB-DB3-Accuracy}, accuracy of individual model stays close to  99\% till an IOU=0.6. At an IOU=0.7, still, the performance is close to 99\%, except FRCNN whose performance decreases to 93.24\%. However, at IOU=0.8 the accuracy of FRCNN drops to 66.67\% while UESegNet-1, SSD and UESegNet-2 have achieved accuracy of 97.08\%, 97.7\%, 93.55\% respectively. The values of precision and recall are shown in Fig. \ref{USTB-DB3-Precision}  and our proposed models get better results.

\textbf{Performance on UBEAR database:} As shown in Fig. \ref{UBEAR-Accuracy}, it has been observed that the accuracy of all the models stays above 92\% till an IOU=0.5, and UESegNet-2 has achieved maximum accuracy of 100\%. However, at IOU=0.6 the performance of all the models decreases below  95\%, except UESegNet-2 which stays at 100\%. At IOU=0.8 the accuracy of FRCNN, SSD, UESegNet-1, and UESegNet-2  drop to 50\% ,61.67\%, 64\%, 94.13\% respectively. The Fig. \ref{UBEAR-Precision} shows precision and recall values of both our proposed model gets better results than existing models.

 After analyzing the performance of each model on different databases, it has been observed that FRCNN performs well till an IOU=0.5, with the increase in IOU its performance decreases drastically. The UESegNet-1 and  SSD have performed very close to each other until an IOU=0.7 on the majority of the databases, and their performance is much better than FRCNN but not as good as UESegNet-2. However, for higher values of IOU, the UESegNet-1 performs better than SSD on the majority of the databases. The UESegNet-2 outperformed all the proposed models on the majority of the databases mentioned in this paper and obtained excellent results for higher values of IOUs. At an IOU=0.5 this model has achieved an accuracy close to 100\% on the majority of the databases and it stays above 90\% till an IOU=0.8.

 \begin{table*}[h]
 \label{Results}
 \caption{The Accuracy - Precision - Recall and  F1-Score Values at different Overlap (IOU)  using FRCNN and SSD and $UESEGNET-1$ and $UESEGNET-2$}
\centering

\begin{tabular}{ |p{2.0cm}  ||p{.8cm}|p{.8cm}|p{.8cm}||p{.8cm}|p{.8cm}|p{.8cm}||p{.8cm}|p{.8cm}|p{.8cm}||p{.8cm}|p{.8cm}|p{.8cm}|p{.8cm}|p{.8cm}|p{.8cm}| }
 \hline
 \textbf{Model} &\multicolumn{3}{|c|}{\textbf{Accuracy}}&\multicolumn{3}{|c|}{\textbf{Precision}}&\multicolumn{3}{|c|}{\textbf{Recall} } &\multicolumn{3}{|c|}{\textbf{F1-Score} } \\
 \hline
 \hline
 \textbf{Database} &\multicolumn{12}{|c|}{ \textbf{IIT Kanpur}}\\
 \hline
   \textbf{IOU Threshold} & \textbf{0.6} &\textbf{0.7}&\textbf{0.8} & \textbf{0.6} &\textbf{0.7}&\textbf{0.8}& \textbf{0.6} &\textbf{0.7}&\textbf{0.8} & \textbf{0.6} &\textbf{0.7}&\textbf{0.8}\\
 \hline

 \textbf{FRCNN}  & 68.79 & 45.39 & \cellcolor{red!75} 13.48  & 91.51 & 60.38 & 17.92  & 80.17 & 52.89 & 15.7  & 85.46 & 56.39 & 16.74\\
 \hline
 \textbf{SSD}  & 90.78 & 90.78 & 86.52  & 100.0 & 100.0 & 95.31  & 100.0 & 100.0 & 95.31  & 100.0 & 100.0 & 95.31 \\
 \hline
 \textbf{UESegNet-1}  & 89.36 & 88.65 &  83.69  & 98.44 & 97.66 & 92.19  & 99.21 & 98.43 & 92.91  & 98.82 & 98.04 & 92.55 \\
 \hline
 \textbf{UESegNet-2}  & \cellcolor{green!75}  97.87 &  \cellcolor{green!75}96.45 & \cellcolor{green!65}95.74  & 97.18 & 95.77 & \cellcolor{green!75}95.07  & 98.57 & 97.14 & \cellcolor{green!75} 96.43  & 97.87 & 96.45 & \cellcolor{green!75} 95.74 \\
 \hline
 \hline
 \textbf{Database}&\multicolumn{12}{|c|}{ \textbf{IIT Delhi}}\\
 \hline
   \textbf{IOU Threshold} & \textbf{0.6} &\textbf{0.7}&\textbf{0.8} & \textbf{0.6} &\textbf{0.7}&\textbf{0.8}& \textbf{0.6} &\textbf{0.7}&\textbf{0.8} & \textbf{0.6} &\textbf{0.7}&\textbf{0.8}\\
 \hline

 \textbf{FRCNN}  & 61.84 & 55.26 & \cellcolor{red!75} 41.67  & 93.38 & 83.44 & 62.91  & 92.16 & 82.35 & 62.09  & 92.76 & 82.89 & 62.5\\
 \hline

 \textbf{SSD} & 64.47 & 59.65 & 43.42  & 93.04 & 86.08 & 62.66  & 93.63 & 86.62 & 63.06  & 93.33 & 86.35 & 62.86
 \\
 \hline
  \textbf{UESegNet-1}& \cellcolor{green!75} 70.65 & 63.35 &  42.77  & 93.7 & 84.25 & 60.63  & 63.98 & 57.53 & 41.4  & 76.04 & 68.37 & 49.2\\
 \hline
 \textbf{UESegNet-2}  &  66.23 & \cellcolor{green!75} 64.04 & \cellcolor{green!65}57.46  &  88.82 & 85.88 & \cellcolor{green!75} 77.06  & 90.42 & 87.43 & \cellcolor{green!75} 78.44  & 89.61 & 86.65 & \cellcolor{green!75} 77.74\\
 \hline
\hline

 \textbf{Database}&\multicolumn{12}{|c|}{ \textbf{UND-E}}\\
 \hline
   \textbf{IOU Threshold} & \textbf{0.6} &\textbf{0.7}&\textbf{0.8} & \textbf{0.6} &\textbf{0.7}&\textbf{0.8}& \textbf{0.6} &\textbf{0.7}&\textbf{0.8} & \textbf{0.6} &\textbf{0.7}&\textbf{0.8}\\
 \hline
 \textbf{FRCNN}  & 17.0 & 7.11 & \cellcolor{red!75} 1.58  & 28.48 & 11.92 & 2.65  & 28.48 & 11.92 & 2.65  & 28.48 & 11.92 & 2.65\\
 \hline
 \textbf{SSD} & 91.16 & 89.66 &  80.17  & 98.6 & 96.97 & 86.71  & 98.37 & 96.74 & 86.51  & 98.49 & 96.86 & 86.61 \\
 \hline
 \textbf{UESegNet-1} & 90.09 & 88.58 & 82.33  & 98.58 & 96.93 & \cellcolor{green!75} 90.09  & 96.98 & 95.36 & \cellcolor{green!75} 88.63  & 97.78 & 96.14 & \cellcolor{green!75}89.36 \\
 \hline
 \textbf{UESegNet-2} & \cellcolor{green!75}  95.47 & \cellcolor{green!75} 92.46 & \cellcolor{green!65}85.99  & 96.72 & 93.67 & 87.12  & 97.36 & 94.29 & 87.69  & 97.04 & 93.98 &  87.4
 \\
 \hline
 \hline
\textbf{Database} &\multicolumn{12}{|c|}{ \textbf{UND-J2}}\\
 \hline
   \textbf{IOU Threshold} & \textbf{0.6} &\textbf{0.7}&\textbf{0.8} & \textbf{0.6} &\textbf{0.7}&\textbf{0.8}& \textbf{0.6} &\textbf{0.7}&\textbf{0.8} & \textbf{0.6} &\textbf{0.7}&\textbf{0.8}\\
 \hline
 \textbf{FRCNN} & 86.23 & 66.57 & \cellcolor{red!75} 25.84  & 92.37 & 71.31 & 27.68  & 91.91 & 70.95 & 27.55  & 92.14 & 71.13 & 27.61\\
 \hline
 \textbf{SSD} & 91.17 & 88.4 &  77.65  & 96.79 & 93.84 & 82.44  & 96.35 & 93.42 & 82.06  & 96.57 & 93.63 & 82.25 \\
 \hline
  \textbf{UESegNet-1}  & 90.58 & 87.33 & 80.0  & 96.5 & 93.05 & 85.23  & 96.02 & 92.59 & 84.81  & 96.26 & 92.82 & 85.02 \\
 \hline
 \textbf{UESegNet-2}  & \cellcolor{green!75} 96.8 & \cellcolor{green!75} 96.08 & \cellcolor{green!65}93.39  &98.44 & 97.7 &\cellcolor{green!75} 94.97  & 98.61 & 97.87 & \cellcolor{green!75} 95.13  & 98.52 & 97.79 & \cellcolor{green!75} 95.05  \\
 \hline
\hline

 \textbf{Database}&\multicolumn{12}{|c|}{ \textbf{USTB-DB3}}\\
 \hline
   \textbf{IOU Threshold} & \textbf{0.6} &\textbf{0.7}&\textbf{0.8} & \textbf{0.6} &\textbf{0.7}&\textbf{0.8}& \textbf{0.6} &\textbf{0.7}&\textbf{0.8} & \textbf{0.6} &\textbf{0.7}&\textbf{0.8}\\
 \hline
 \textbf{FRCNN} & 98.77 & 93.24 & \cellcolor{red!65} 66.67  & 99.54 & 93.96 & 67.18  & 99.54 & 93.96 & 67.18  & 99.54 & 93.96 & 67.18\\
 \hline
 
 \textbf{SSD}  & 99.69 & \cellcolor{green!75}99.69 & \cellcolor{green!75} 97.7  & 100.0 & 100.0 & \cellcolor{green!75}98.0  & 100.0 & 100.0 & \cellcolor{green!75} 98.6  & 100.0 & 100.0 & 98.3 \\
 \hline
 \textbf{UESegNet-1}  & 99.54 & 99.39 & 97.08  & 100.0 & 99.85 &  97.53  & 100.0 & 100.0 & 97.83  & 100.0 & 100.0 & 97.68\\
 \hline
 \textbf{UESegNet-2} & \cellcolor{green!75}  99.69 & 99.39 & 93.55  & 100.0 & 99.69 & 93.84  & 100.0 & 99.69 & 93.84  & 100.0 & 100.0 & \cellcolor{green!75} 98.77 \\
 \hline
\hline

\textbf{Database} &\multicolumn{12}{|c|}{ \textbf{UBEAR-2}}\\
 \hline
  \textbf{IOU Threshold} & \textbf{0.6} &\textbf{0.7}&\textbf{0.8} & \textbf{0.6} &\textbf{0.7}&\textbf{0.8}& \textbf{0.6} &\textbf{0.7}&\textbf{0.8} & \textbf{0.6} &\textbf{0.7}&\textbf{0.8}\\
 \hline
 \textbf{FRCNN}   & 92.09 & 82.74 & \cellcolor{red!75} 49.86  & 96.51 & 86.71 & 52.25  & 96.12 & 86.35 & 52.04  & 96.31 & 86.53 & 52.14 \\
 \hline

 \textbf{SSD}  & 92.17 & 85.88 & 61.67  & 96.78 & 90.18 & 64.76  & 98.32 & 91.62 & 65.79  & 97.55 & 90.89 & 65.27 \\
 \hline
  \textbf{UESegNet-1} & 93.4 & 86.4 & 64.32  & 95.94 & 88.75 & 66.07  & 96.98 & 89.72 & 66.78  & 96.46 & 89.23 & 66.42\\
 \hline
 \textbf{UESegNet-2}   & \cellcolor{green!75} 99.84 & \cellcolor{green!75}99.35 & \cellcolor{green!65}94.13  & 99.84 & 99.35 & \cellcolor{green!75} 94.13  & 99.87 & 99.38 & \cellcolor{green!75}94.16  & 99.86 & 99.36 & \cellcolor{green!75} 94.14 \\
 \hline

\end{tabular}

\end{table*}  

\subsection{Performance evaluation based on IOU and Objectness Score }

 \begin{figure*}[h]
 \centering
 \begin{subfigure}[c]{0.24\textwidth}
        \includegraphics[width=1\textwidth]{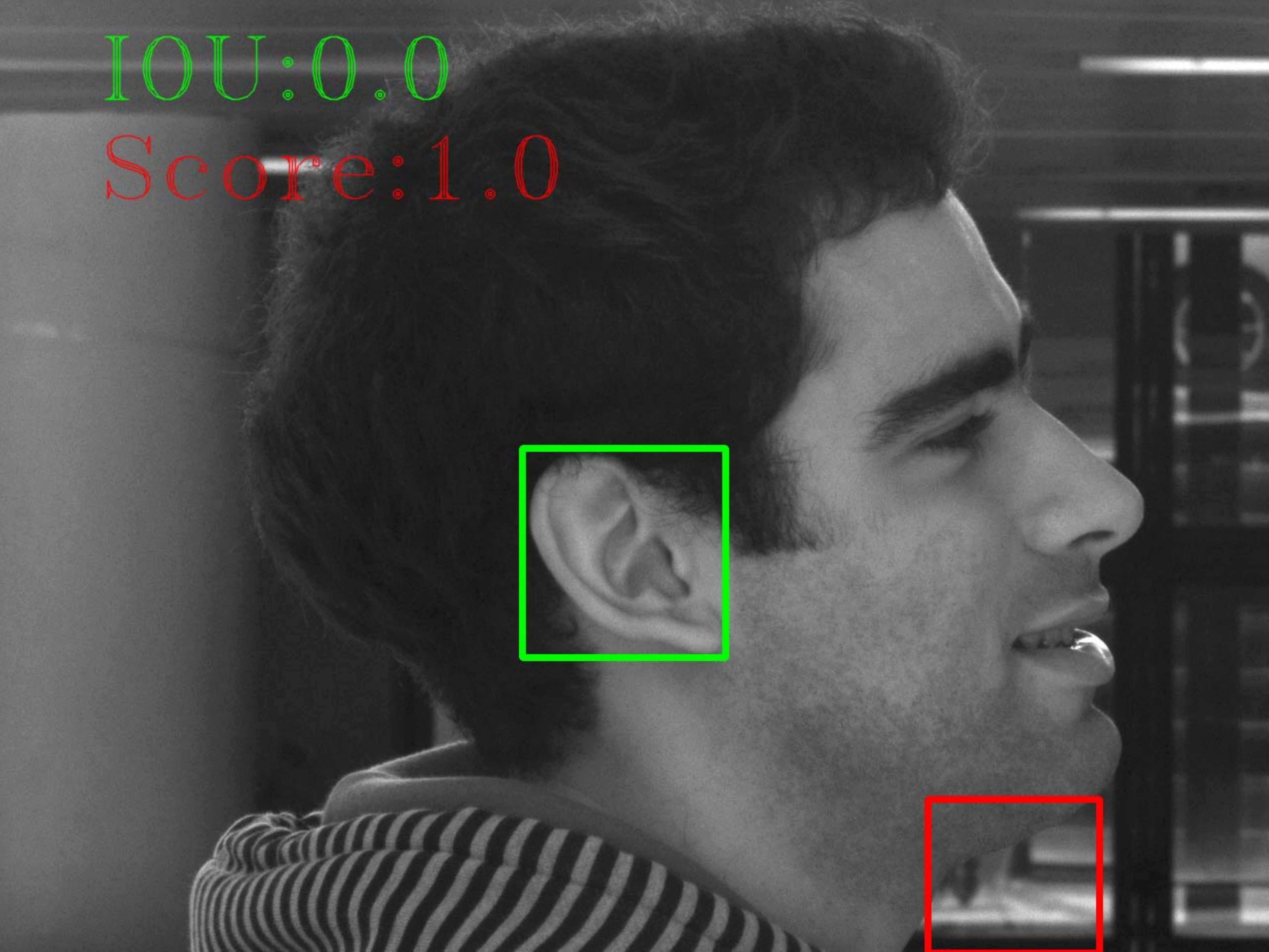}
        \caption{Sample-1}
        \label{Sample-1}
    \end{subfigure}
     \begin{subfigure}[c]{0.24\textwidth}
        \includegraphics[width=1\textwidth]{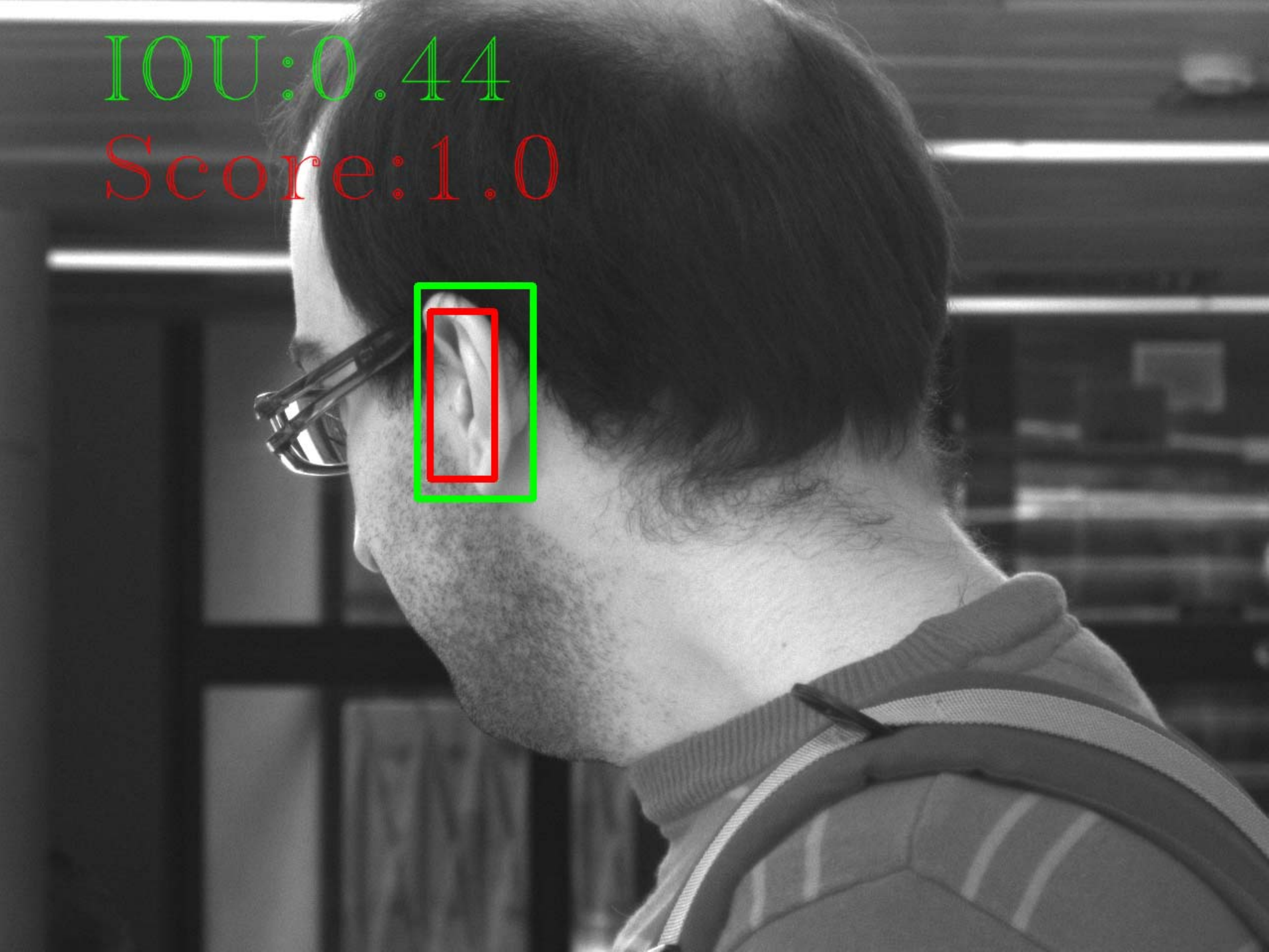}
        \caption{Sample-2}
        \label{Sample-2}
    \end{subfigure}
     \begin{subfigure}[c]{0.24\textwidth}
        \includegraphics[width=1\textwidth]{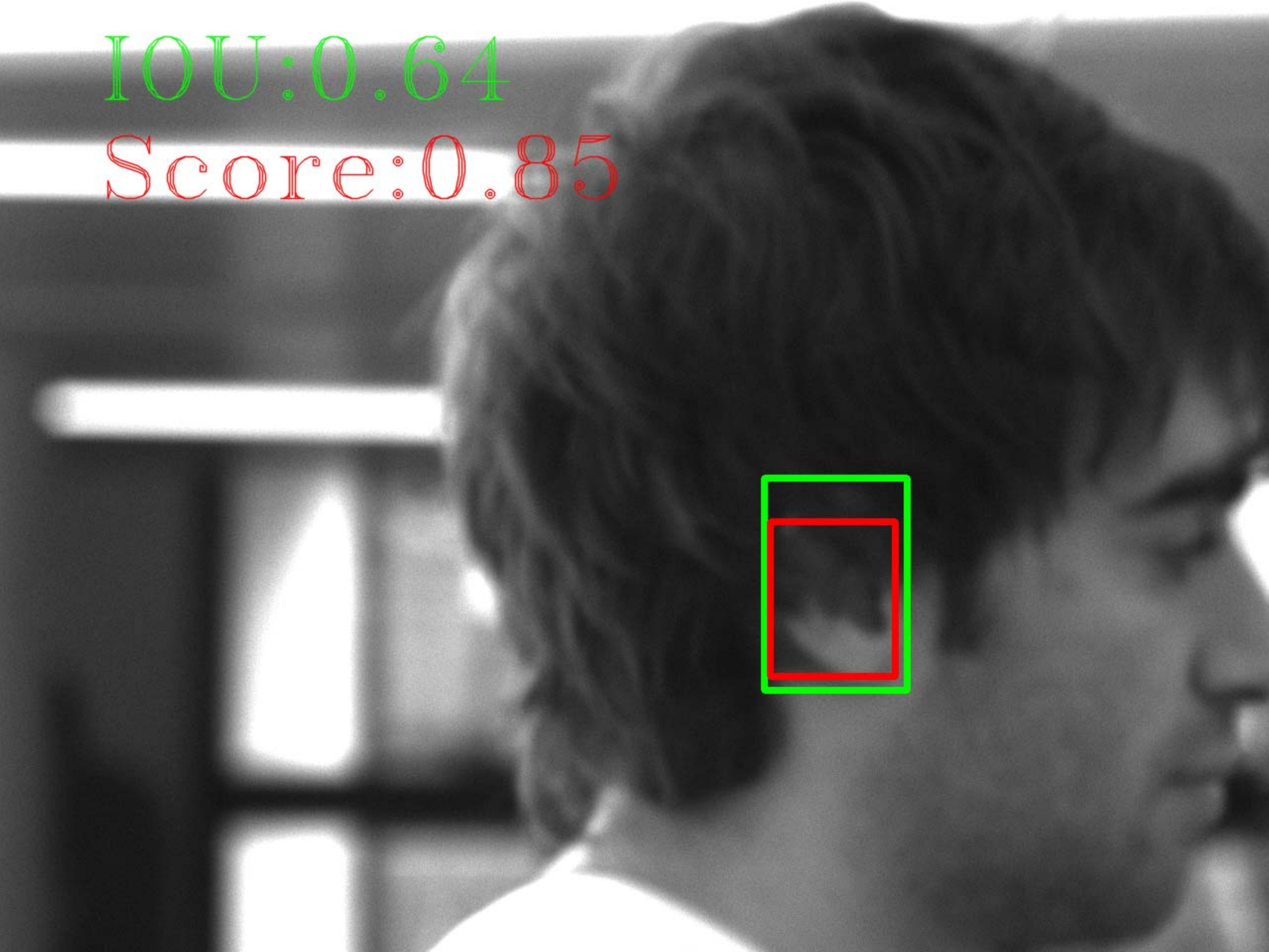}
        \caption{Sample-3}
        \label{Sample-3}
    \end{subfigure}
    \begin{subfigure}[c]{0.24\textwidth}
        \includegraphics[width=1\textwidth]{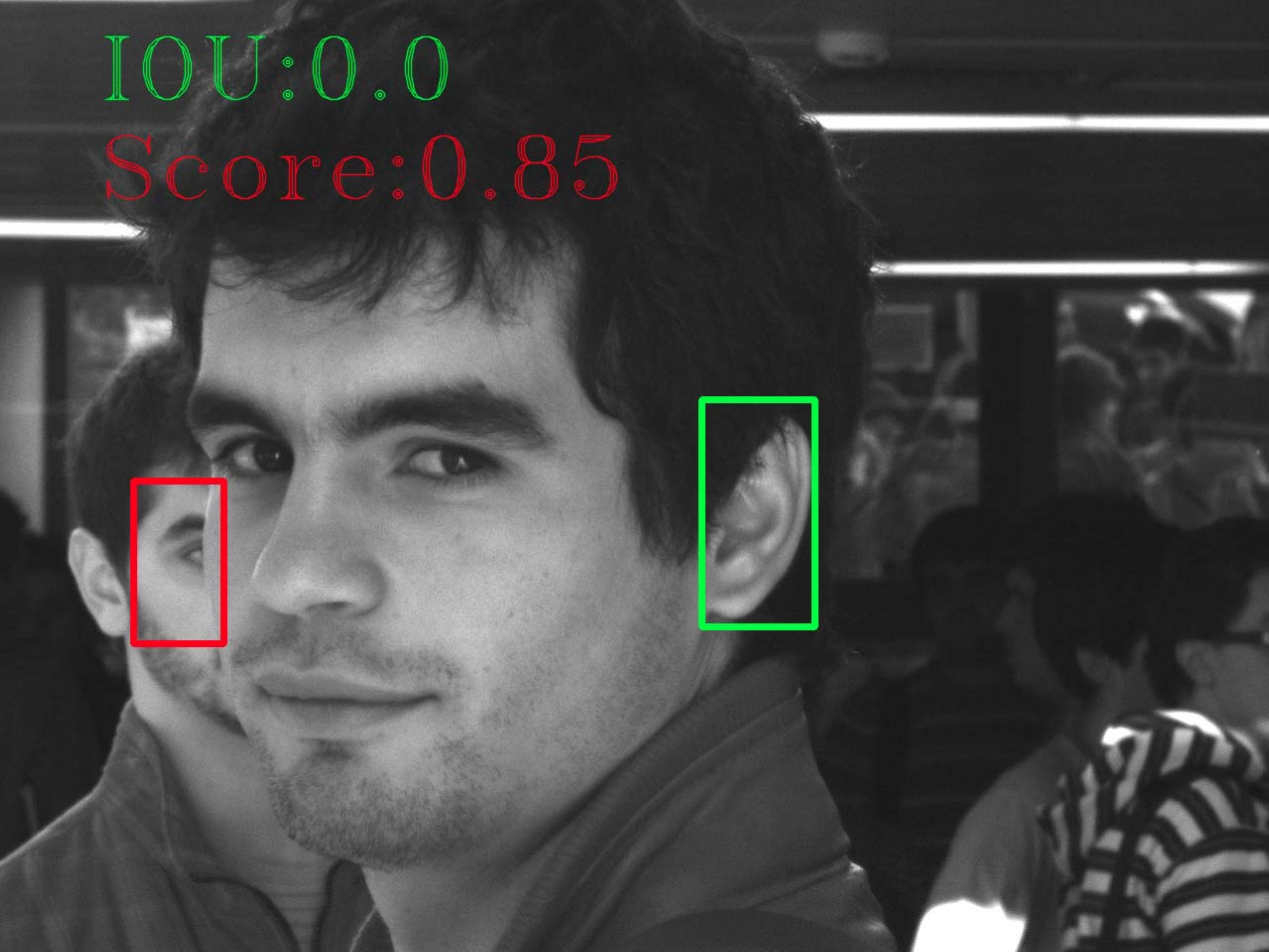}
        \caption{Sample-4}
        \label{Sample-4}
    \end{subfigure}  
 
    \caption[]{ IOU and Objectness Score predicted by UESegNet-2 on UBEAR sample images}
    \label{Objectneass Score}
\end{figure*}

\begin{figure}[h]
		\centering	
		\includegraphics[scale=0.33]{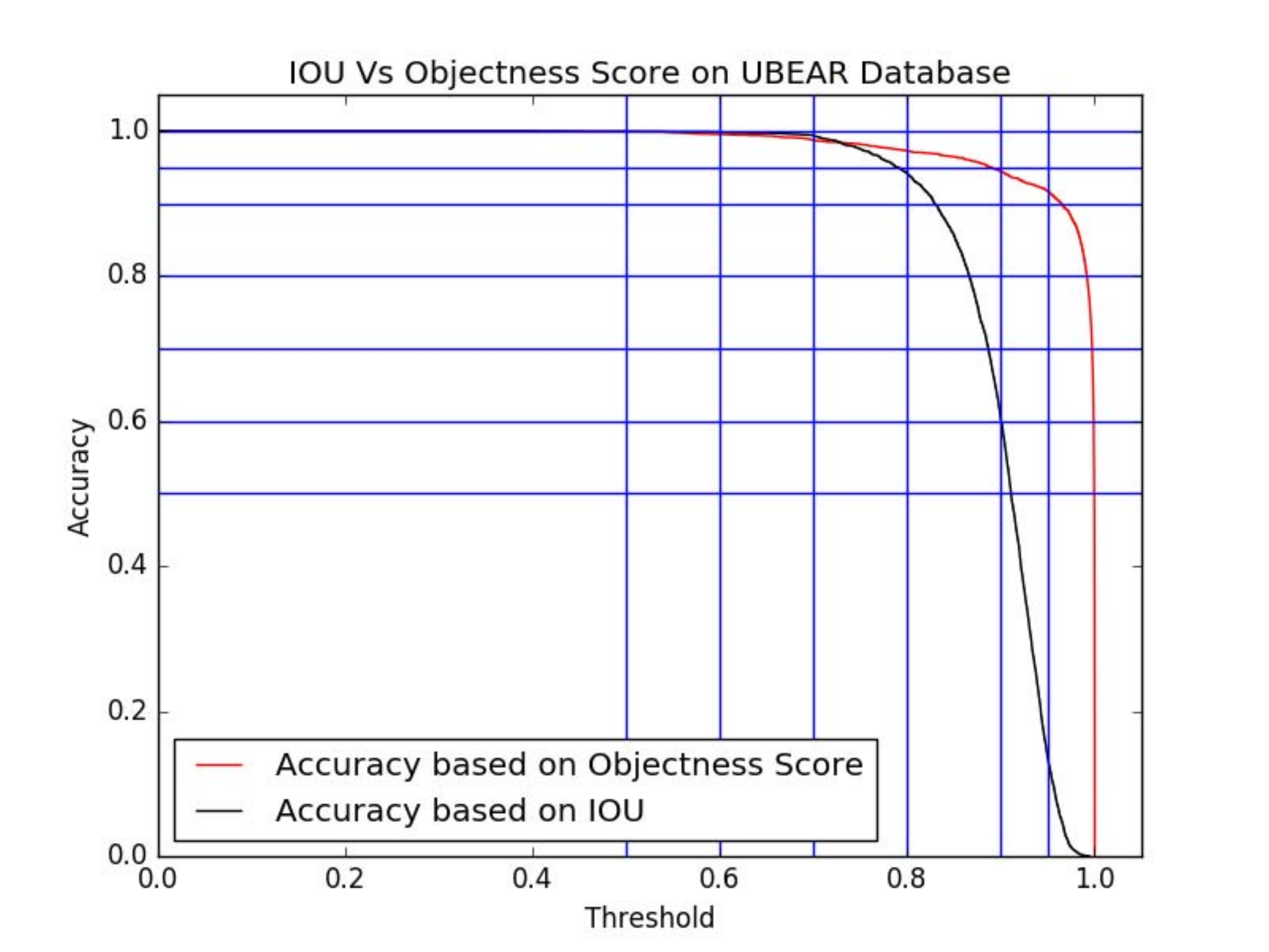}
		\caption{Accuracy based on IOU and Ojectness Score}
        \label{Objecscore}
\end{figure}

\begin{table}[h]
    \centering
     \caption{Objectness Score and IOU}
    \begin{tabular}{|c|c|c|}
        \hline
         \textbf{Sample Images} & \textbf{Objectness Score} & \textbf{IOU}  \\ \hline
       Fig.\ref{Sample-1}. & 1.0  & 0.0 \\ \hline
       Fig.\ref{Sample-2}. & 1.0  & 0.44 \\\hline
       Fig.\ref{Sample-3}. &  0.85  & 0.64   \\\hline
       Fig.\ref{Sample-4}. & 0.85  & 0.0 \\ \hline
    \end{tabular}
    \label{tab:objectscore}
\end{table}

 In \cite{1}, the authors have evaluated the performance of their ear localization model based on the objectness score. A deep learning model calculates the objectness score for the predicted proposals, which indicate how likely the predicted proposal contains an object of any class. However, this is not the exact metric to indicates the accuracy of any object detection model. Hence, the accuracy of any object detection model needs to be measured based on Intersection Over Union (IOU) parameter.\cite{ObjectProposal}, \cite{UBSegNEt}, \cite{UNITBOX} presented a method to measure the accuracy of the predicted proposal by model, and signifies the importance of IOU. To signify the importance of IOU parameter, We have taken some sample images from UBEAR database and evaluated accuracy based on objectness score and IOU for predicted bounding boxes. The results are shown in Fig. \ref{Objectneass Score}. The bounding box in green is the actual ground truth and in red is predicted by our proposed model UESegNet-2. The table \ref{tab:objectscore} depicts the values predicted by model on sample images, which clearly indicates that higher value of objectness score does not signify the exact location of the object in the image, whereas the IOU indicates how tightly the predicted bounding box fit on the ground truth bounding box. Due to the aforementioned reason, we have evaluated the performance of our models based on IOU rather than objectness score. In addition, we have evaluated the accuracy of our model UESegNet-2 based on objectness score and IOU on UBEAR database as shown in Fig. \ref{Objecscore}. It has been observed from the graph that the most of the time accuracy based on objectness score remains above 95\%, whereas the accuracy based on IOU drops significantly for the higher IOU overlapped threshold. Moreover, the accuracy of our proposed model UESegNet-2 based on objectness score on UBEAR database is 95\% at threshold 0.9, whereas the accuracy of the model proposed by \cite{1} at a threshold 0.9 is 90\%.

\subsection{ Qualitative Results }

\begin{figure*}[h]
 \centering
 \begin{subfigure}[c]{0.24\textwidth}
        \includegraphics[width=1\textwidth]{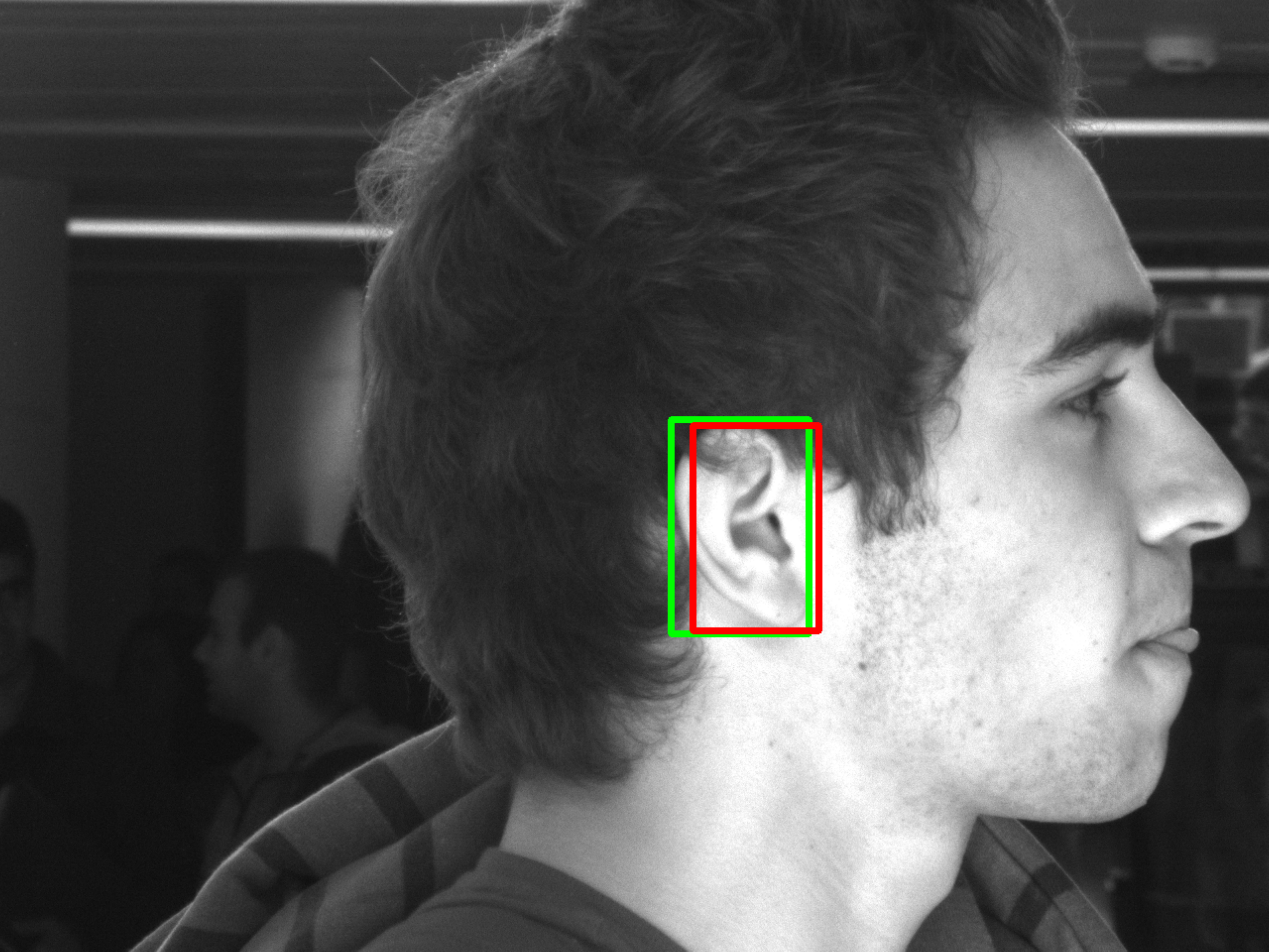}
        \caption{FRCNN}
        \label{FRCNN}
    \end{subfigure}
     \begin{subfigure}[c]{0.24\textwidth}
        \includegraphics[width=1\textwidth]{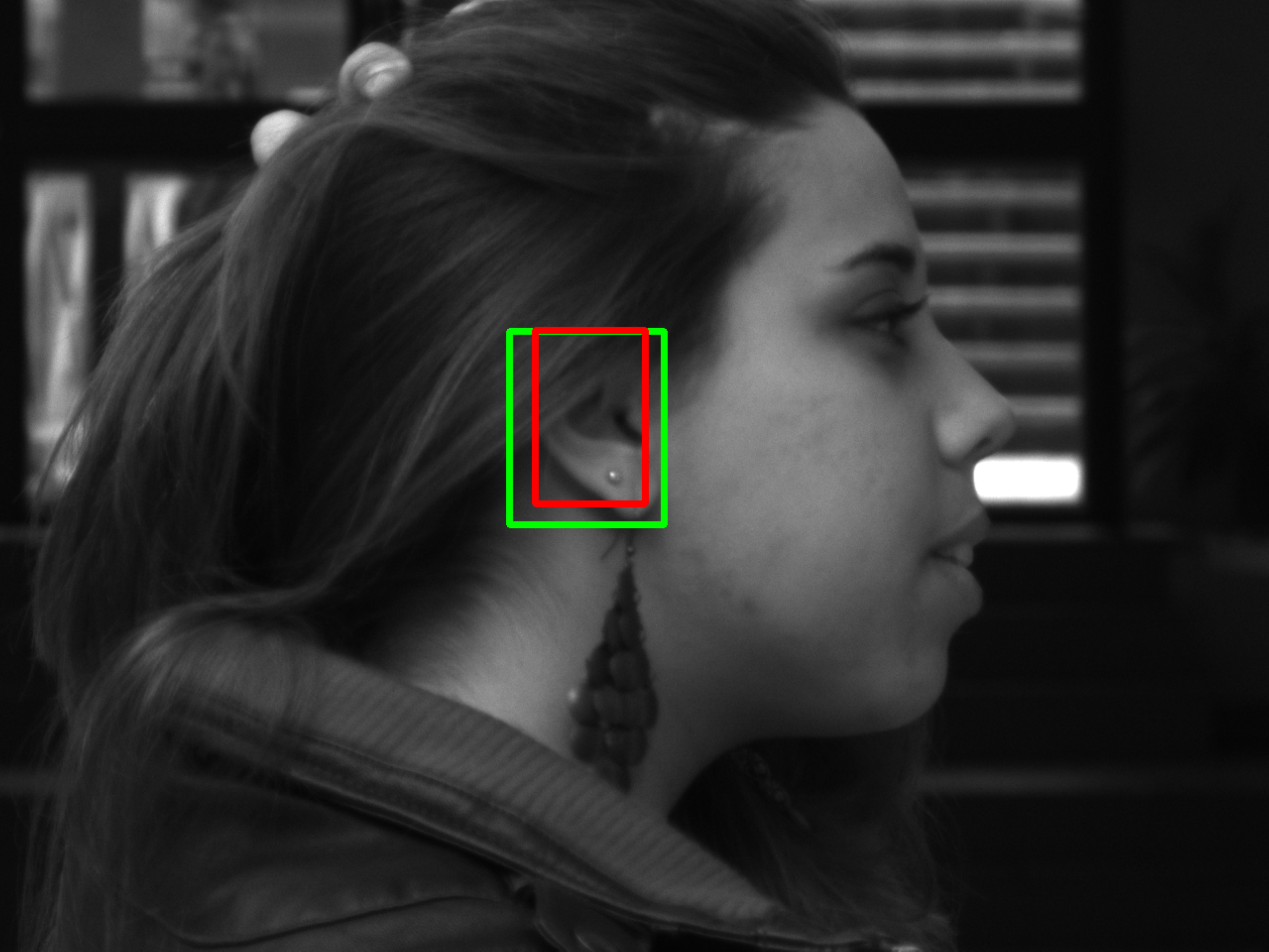}
        \caption{FRCNN}
        \label{C-FRCNN2}
    \end{subfigure}
  \begin{subfigure}[c]{0.24\textwidth}
        \includegraphics[width=1\textwidth]{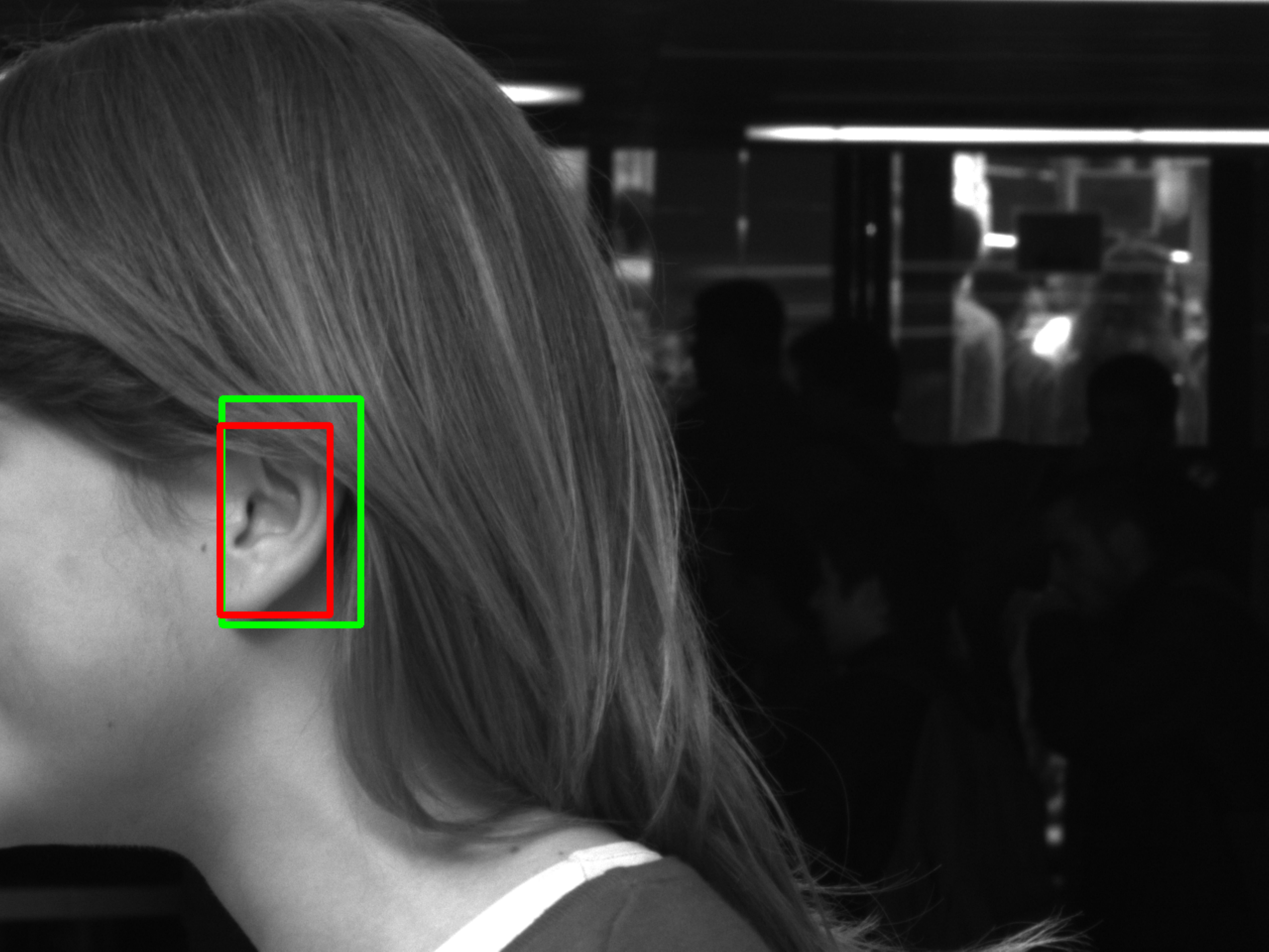}
        \caption{SSD}
        \label{C-SSD}
    \end{subfigure}
     \begin{subfigure}[c]{0.24\textwidth}
        \includegraphics[width=1\textwidth]{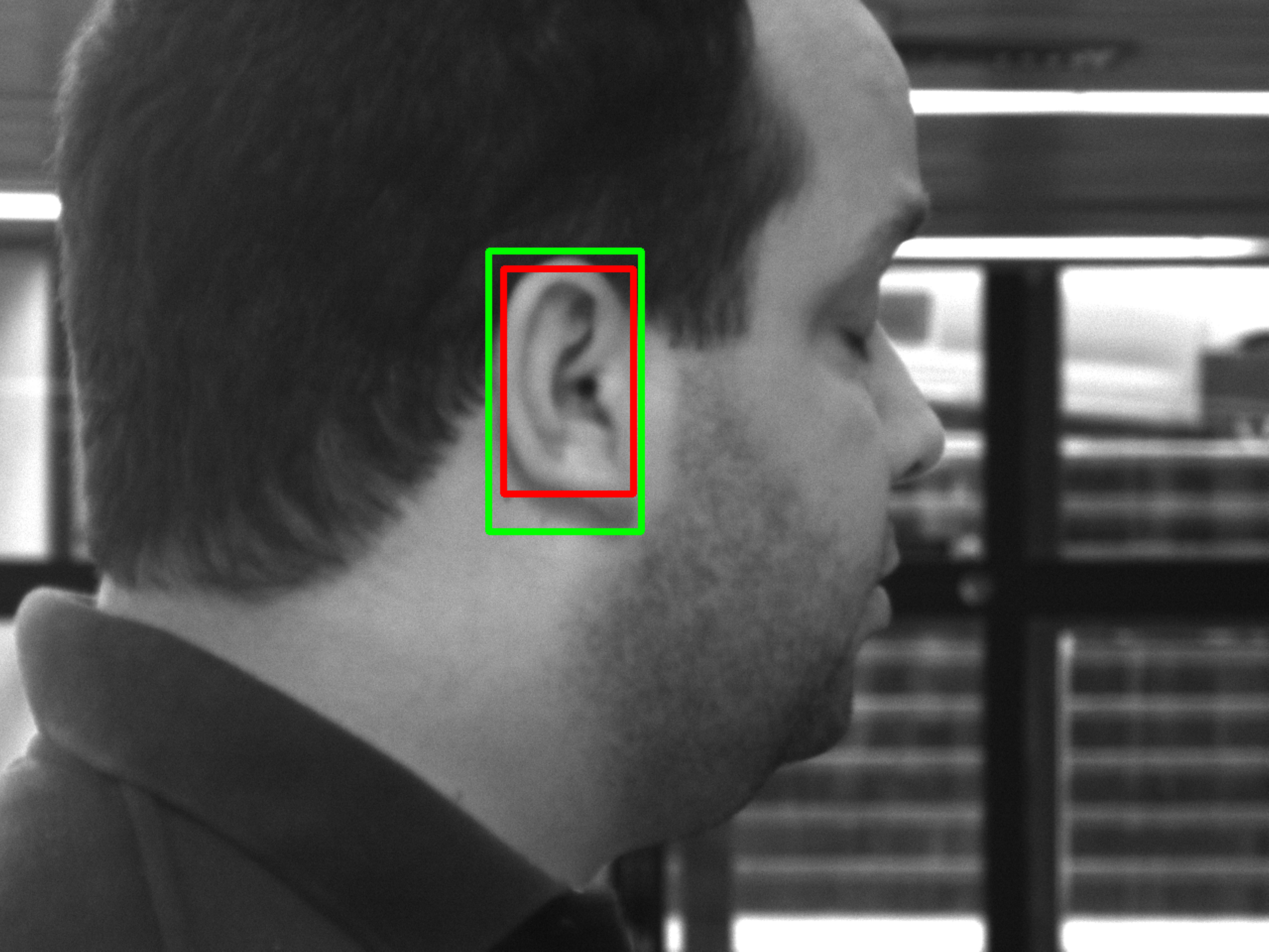}
        \caption{SSD}
        \label{CSSD-2}
    \end{subfigure}
       \begin{subfigure}[c]{0.24\textwidth}
        \includegraphics[width=1\textwidth]{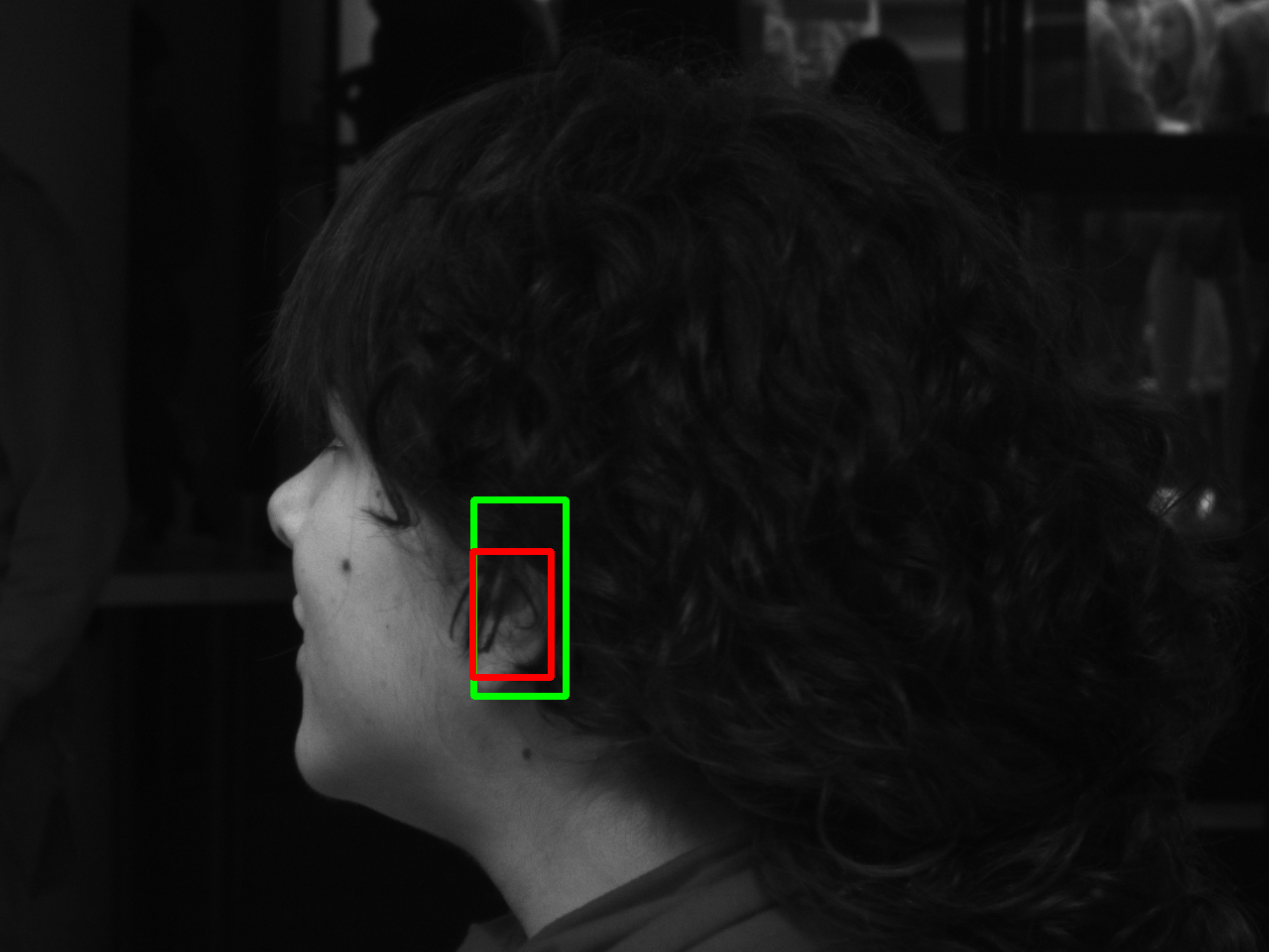}
        \caption{UESegNet-1}
        \label{C-SSH}
    \end{subfigure}
    \begin{subfigure}[c]{0.24\textwidth}
        \includegraphics[width=1\textwidth]{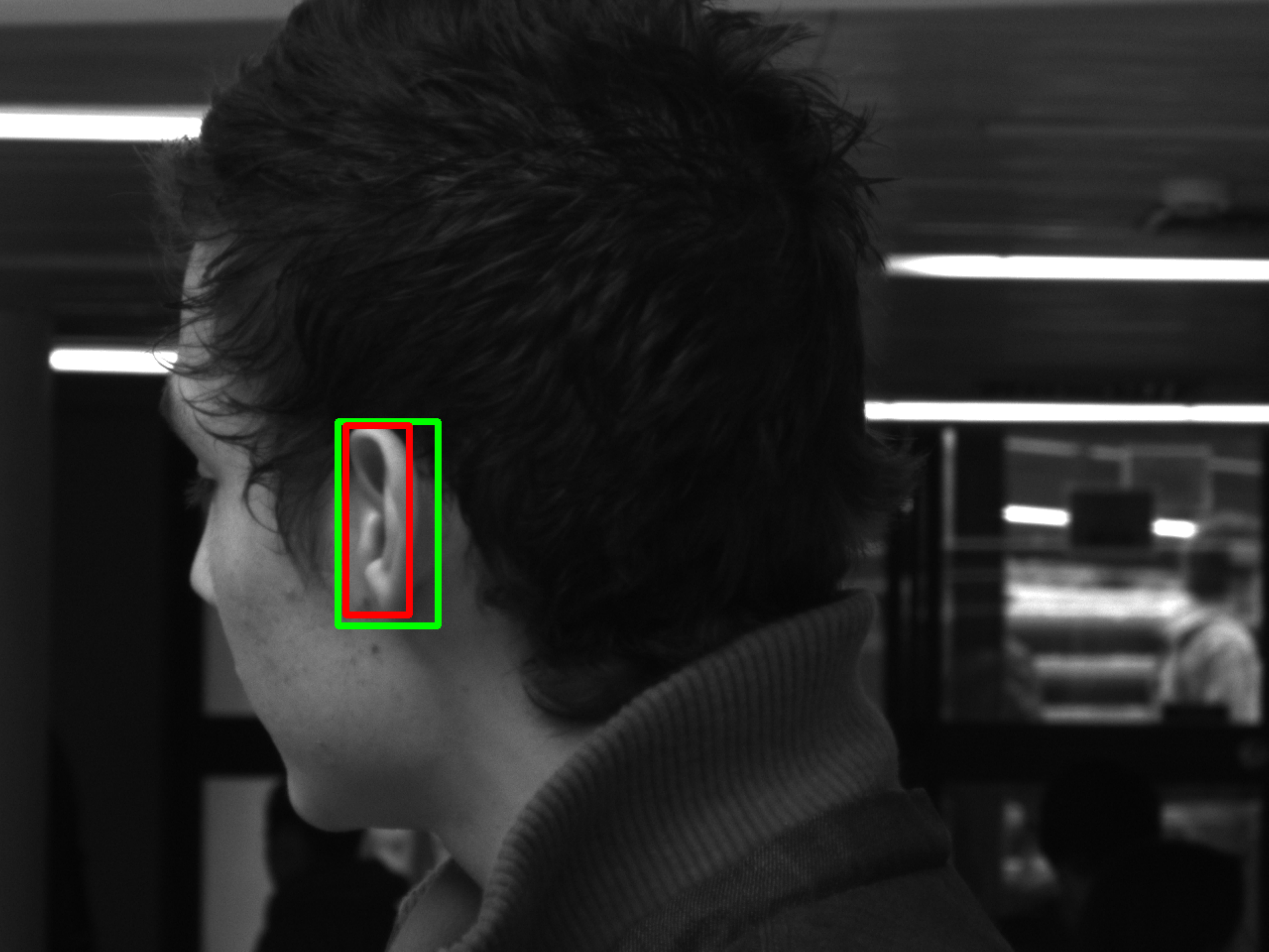}
        \caption{UESegNet-1}
        \label{C-SSH2}
    \end{subfigure}  
     \begin{subfigure}[c]{0.24\textwidth}
        \includegraphics[width=1\textwidth]{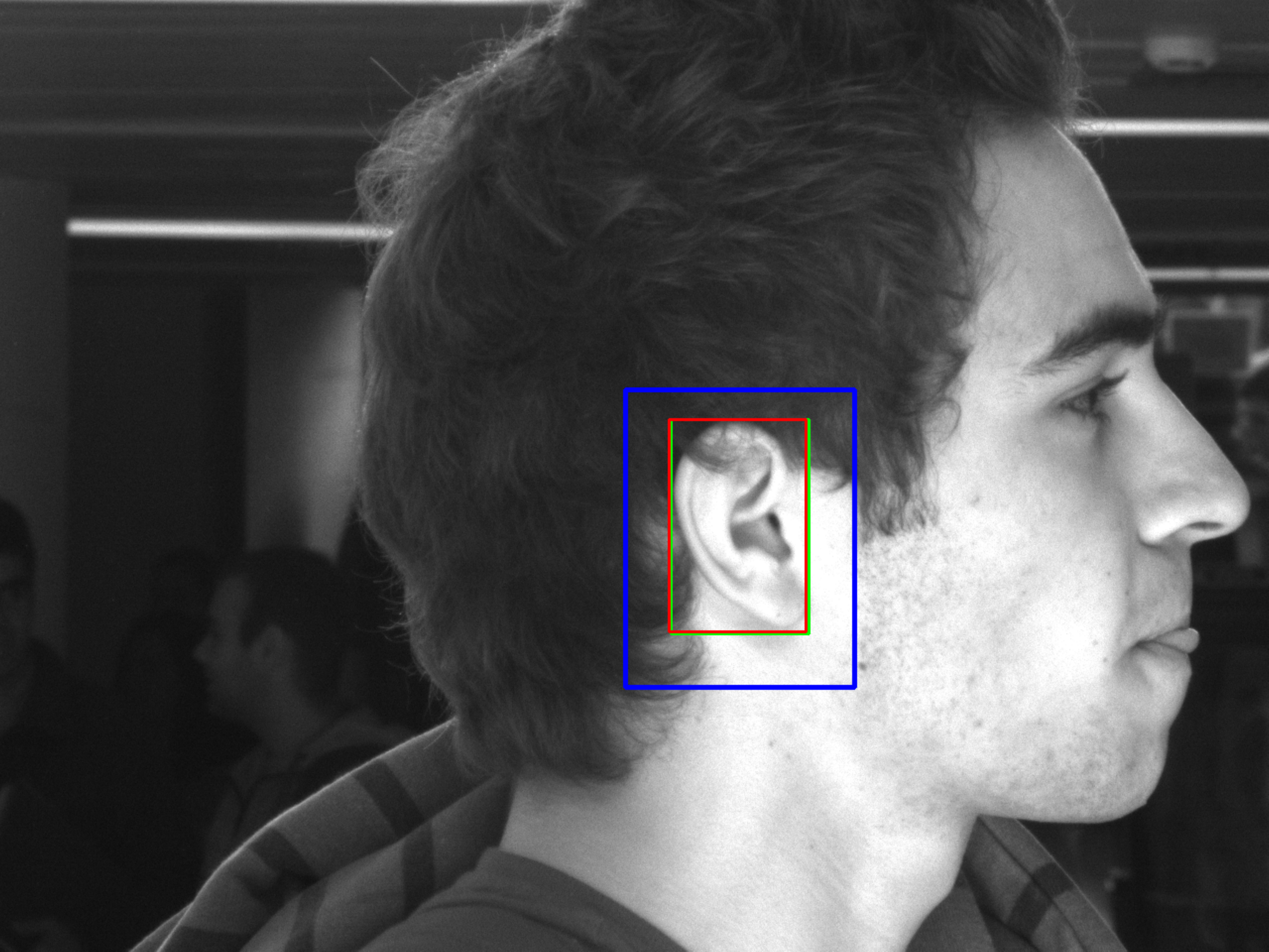}
        \caption{UESegNet-2} 
        \label{C-ssd2}
    \end{subfigure}
    \begin{subfigure}[c]{0.24\textwidth}
        \includegraphics[width=1\textwidth]{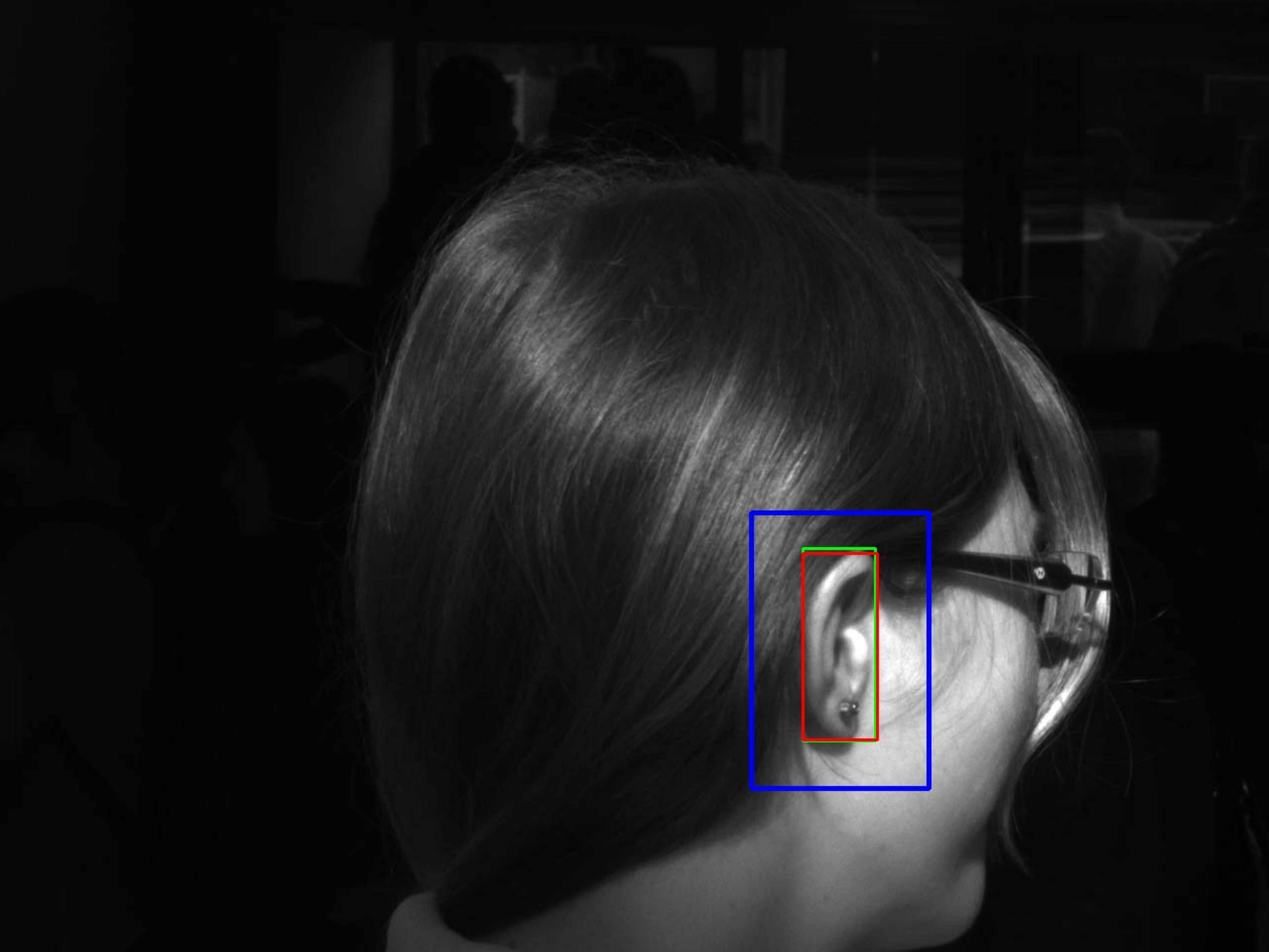}
        \caption{UESegNet-2}
        \label{C-ssd2-2}
    \end{subfigure}     
    \caption[]{Results on Challenging Images (Images contains angle variations, occlusion, illuminations, and scale variations) The bounding box in green is the actual ground truth and in bounding box, in red is predicted by the model and blue is the pan area. }
    \label{Results on Challenge images}
\end{figure*}

The Fig. \ref{Results on Challenge images} shows the qualitative results of models on challenging images selected from UBEAR database. The models are able to localize the ear very accurately in the side face images captured in wild.

 \begin{figure*}[h!]
 \centering
 \begin{subfigure}[c]{0.24\textwidth}
        \includegraphics[width=1\textwidth]{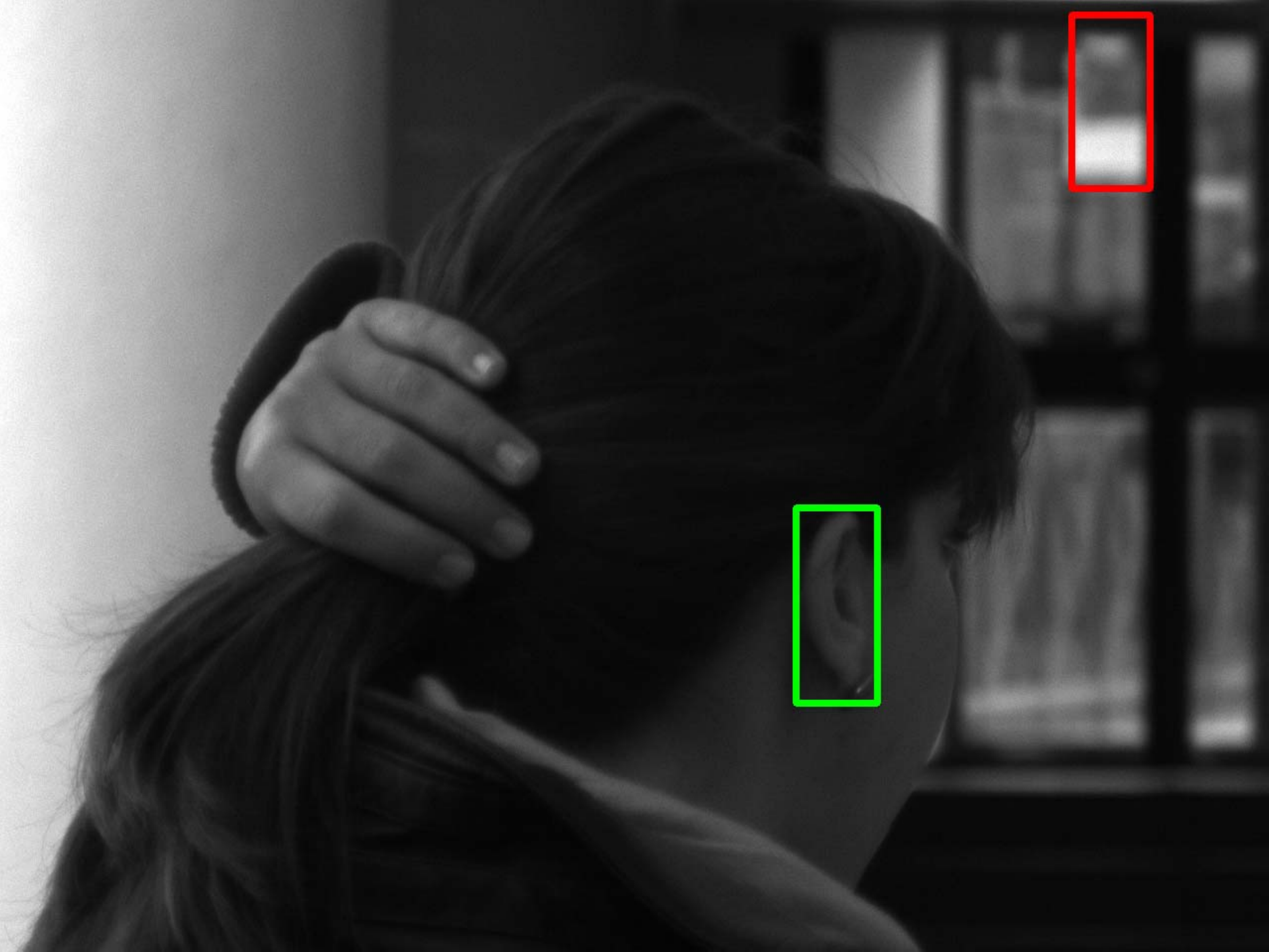}
        \caption{FRCNN}
        \label{M_FRCNN1}
    \end{subfigure}
     \begin{subfigure}[c]{0.24\textwidth}
        \includegraphics[width=1\textwidth]{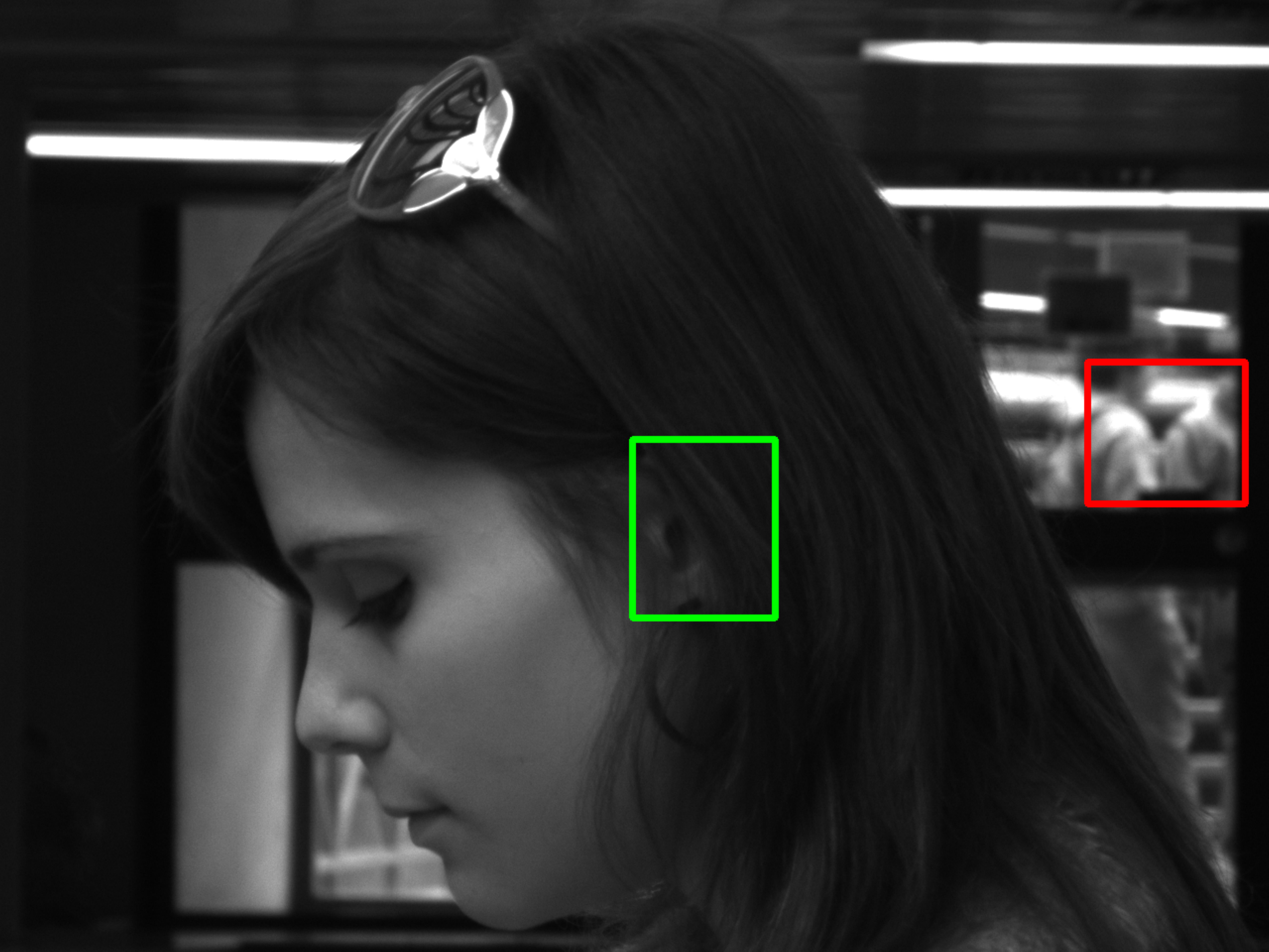}
        \caption{FRCNN}
        \label{M_FRCNN2}
    \end{subfigure}
  \begin{subfigure}[c]{0.24\textwidth}
        \includegraphics[width=1\textwidth]{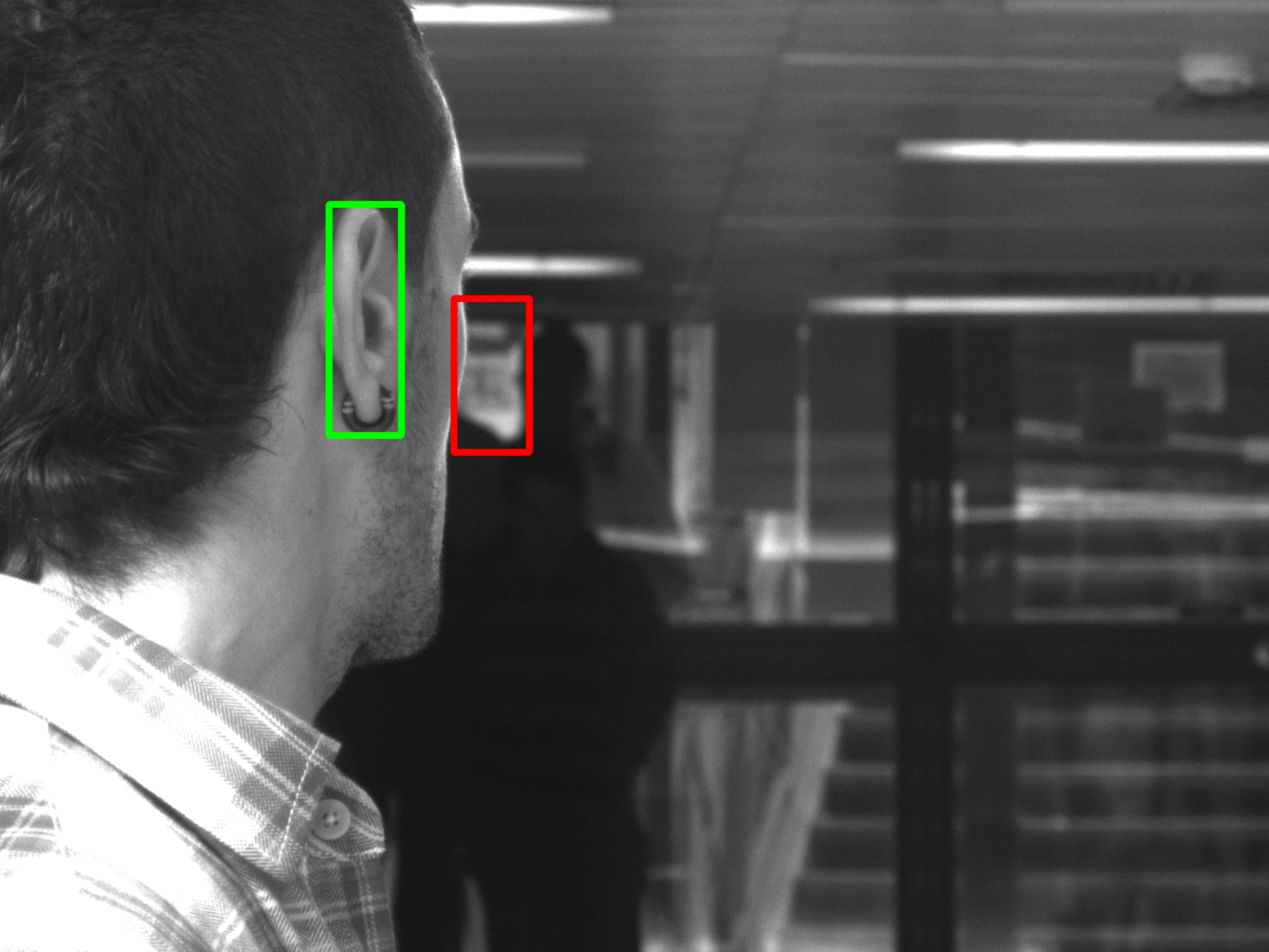}
        \caption{SSD}
        \label{M_SSD_1}
    \end{subfigure}
     \begin{subfigure}[c]{0.24\textwidth}
        \includegraphics[width=1\textwidth]{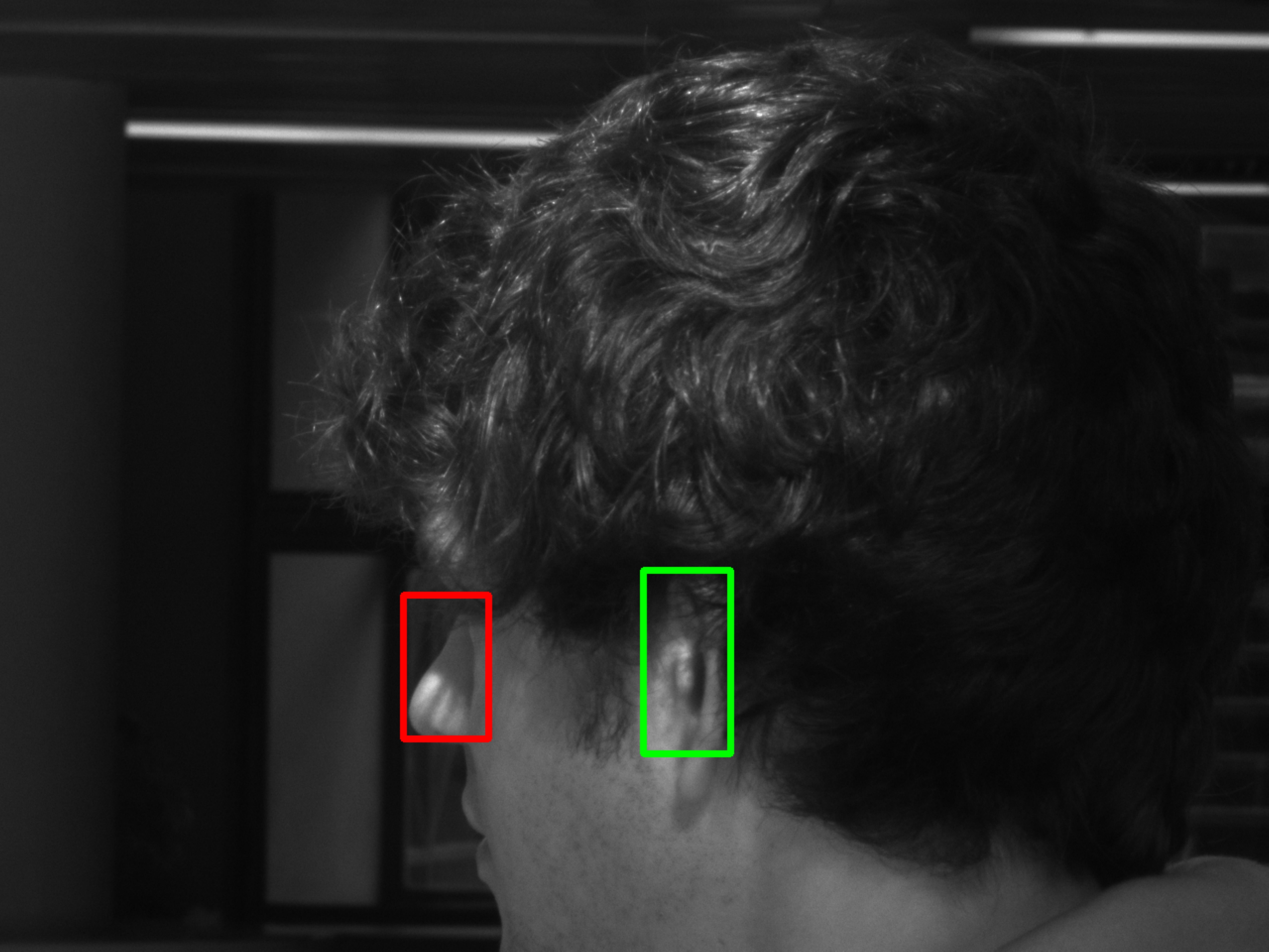}
        \caption{SSD}
        \label{M_SSD_2}
    \end{subfigure}
         \begin{subfigure}[c]{0.24\textwidth}
        \includegraphics[width=1\textwidth]{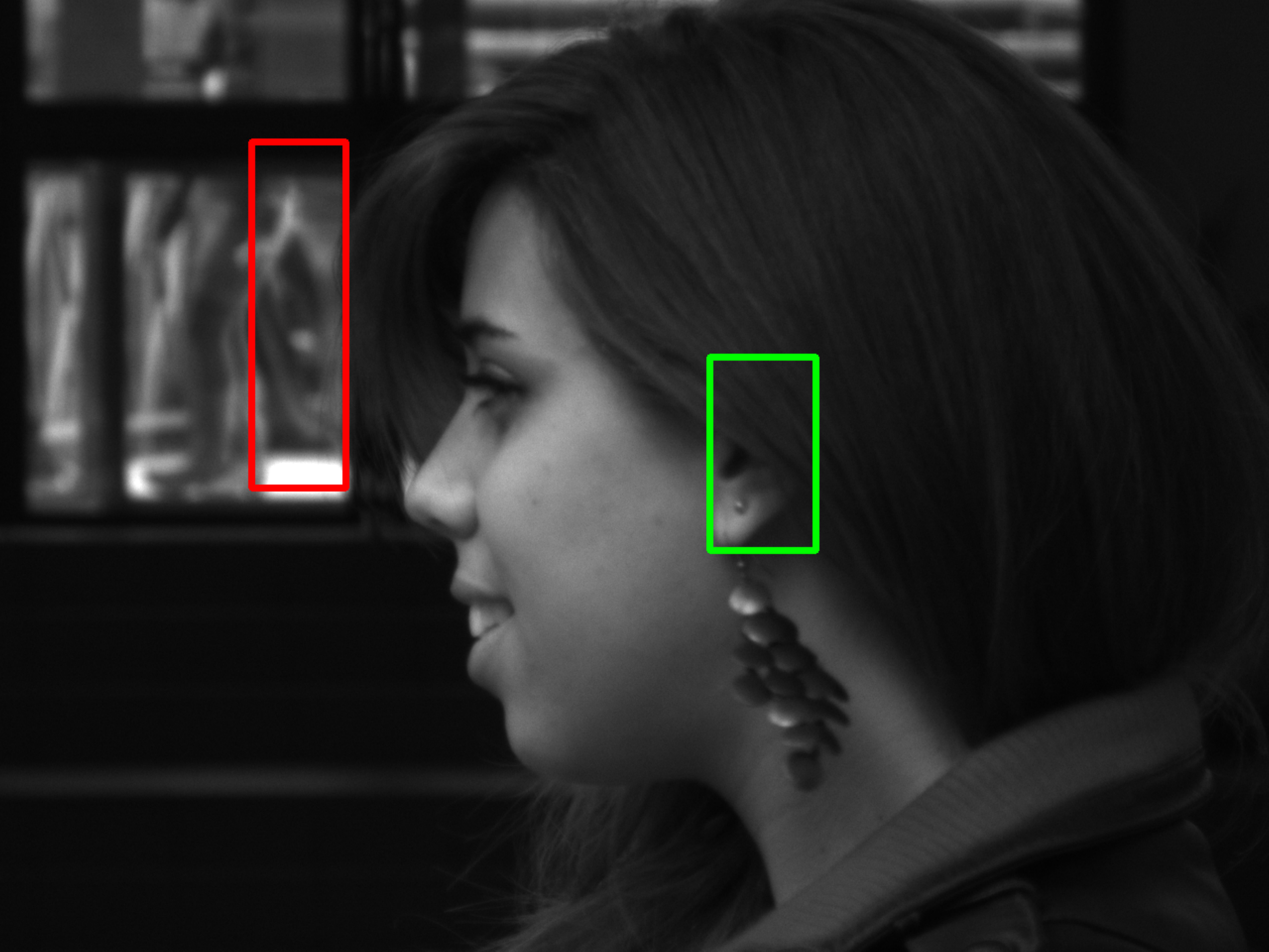}
        \caption{UESegNet-1}
        \label{M_SSH1}
    \end{subfigure}
    \begin{subfigure}[c]{0.24\textwidth}
        \includegraphics[width=1\textwidth]{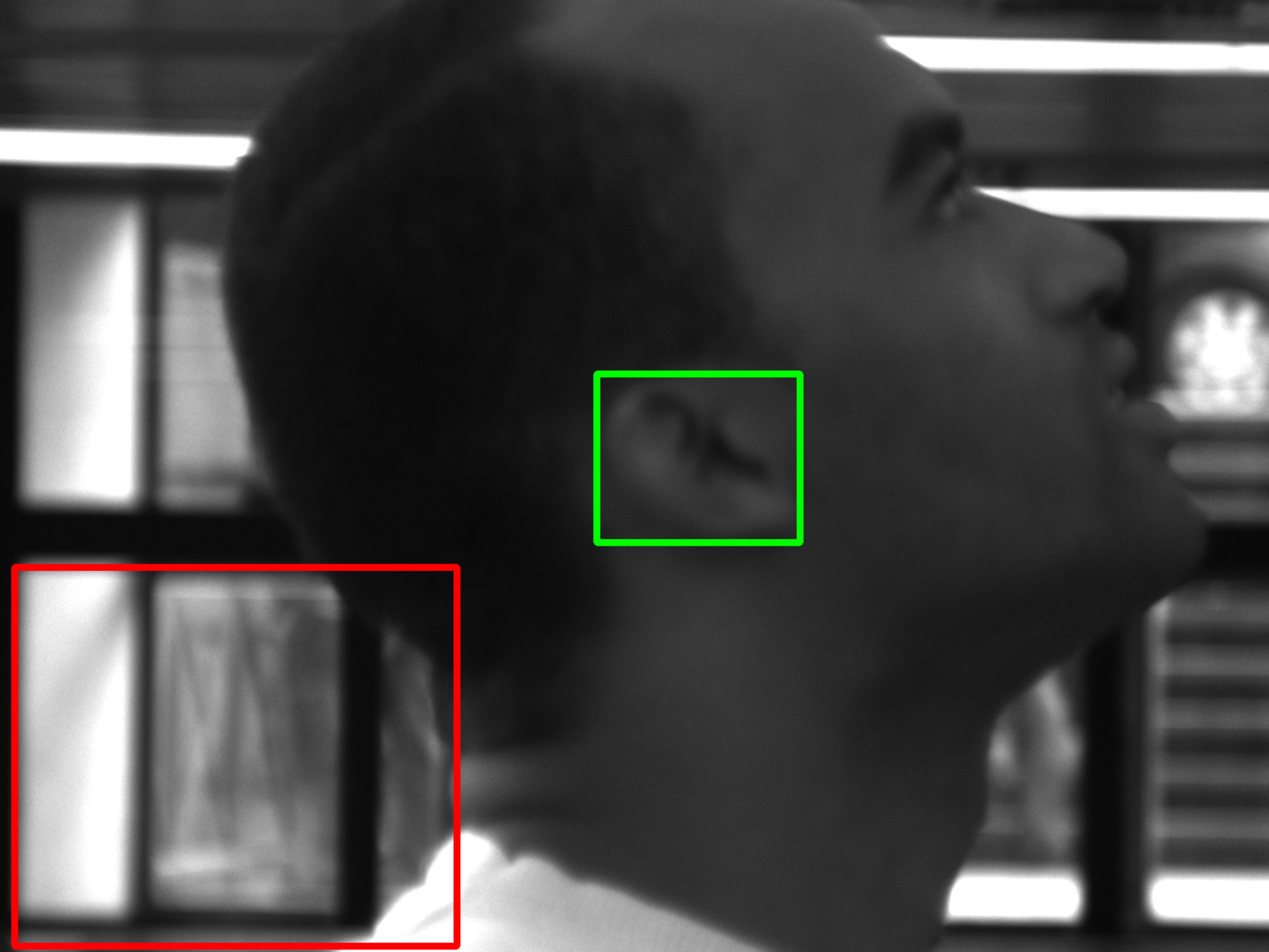}
        \caption{UESegNet-1}
        \label{M_SSH2}
    \end{subfigure} 
     \begin{subfigure}[c]{0.24\textwidth}
        \includegraphics[width=1\textwidth]{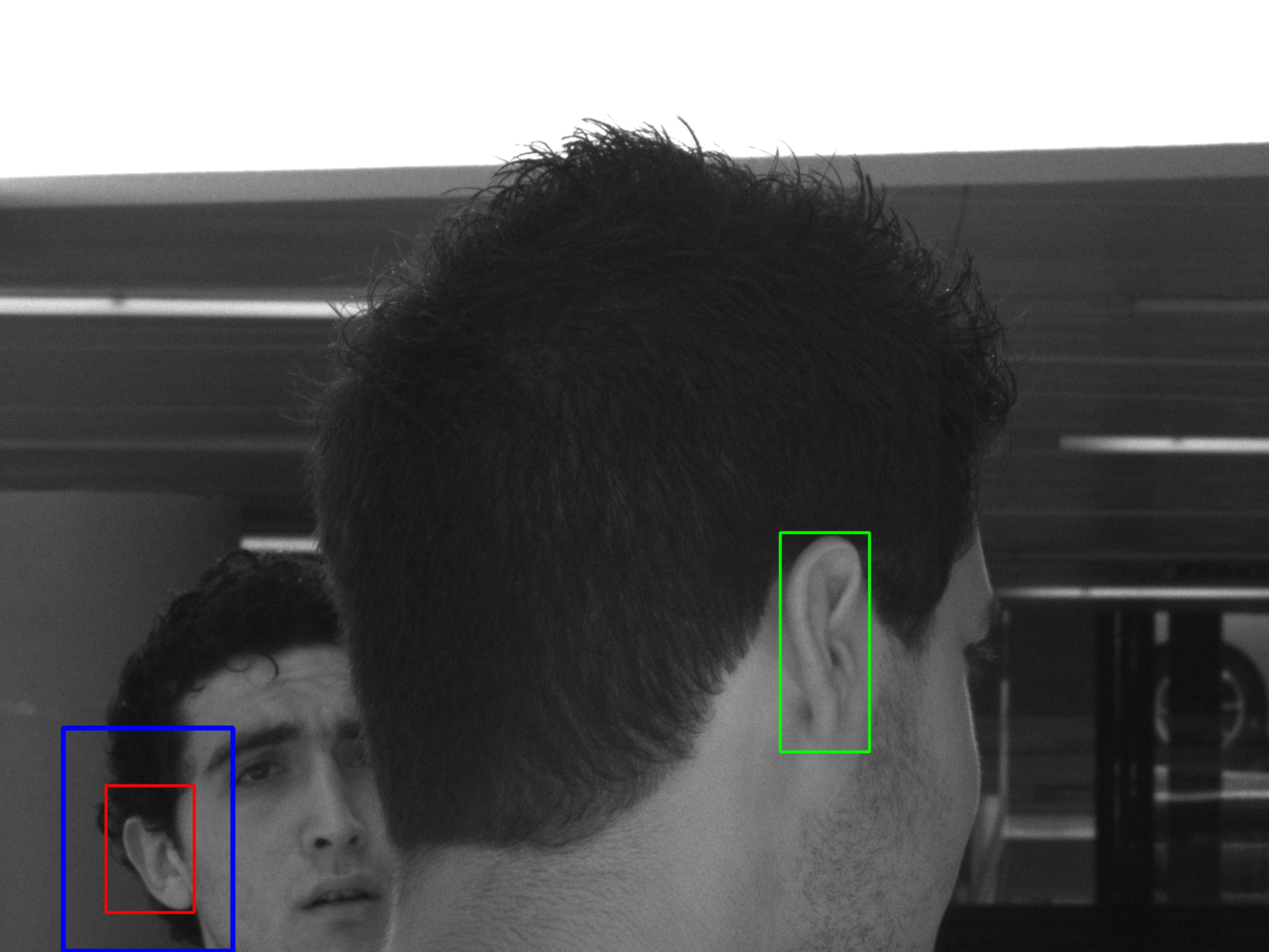}
        \caption{UESegNet-2}
        \label{M_SSD2_1}
    \end{subfigure}
    \begin{subfigure}[c]{0.24\textwidth}
        \includegraphics[width=1\textwidth]{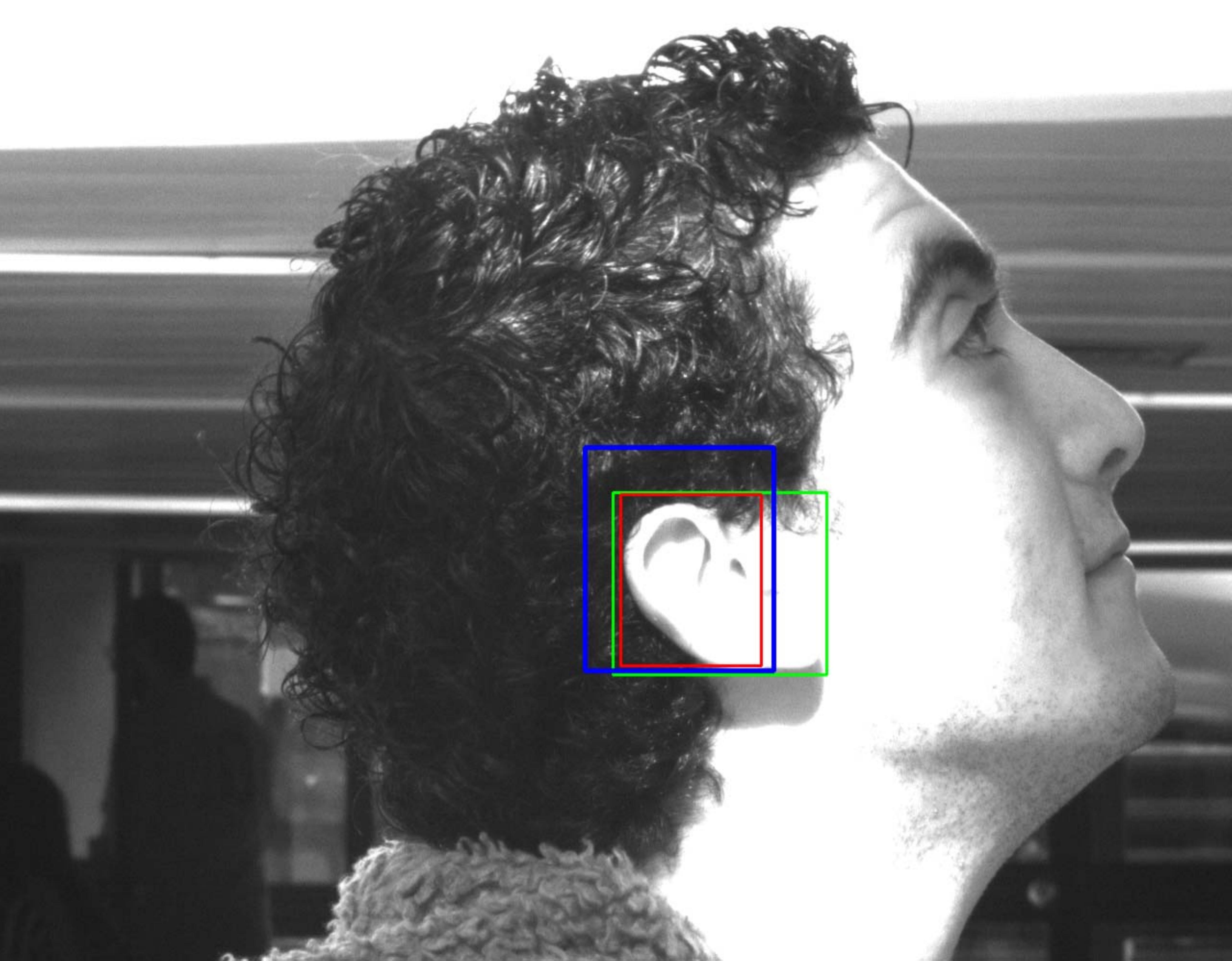}
        \caption{UESegNet-2}
        \label{M_SSD2_2}
    \end{subfigure}     
    
    \caption[]{Misclassified images by models. The bounding box in green is the actual ground truth and in bounding box, in red is predicted by the model and blue is the pan area. }
    \label{Misclassified Cases}
\end{figure*}
 
 \subsection{ Miss-Classified Images}
 
 The Fig. \ref{Misclassified Cases} shows some miss-classified images by models. The FRCNN is failed for images, as shown in Fig.\ref{M_FRCNN1} and  Fig.\ref{M_FRCNN2}, is due to huge angle variation and occlusion (by hairs) respectively.
 The model SSD miss-classified the images, as shown in Fig. \ref{M_SSD_1} and in Fig. \ref{M_SSD_2} is because of extreme angle position and similar features like ear shape. Fig. \ref{M_SSH1} and Fig. \ref{M_SSH2} shows the images in which the UESegNet-1 is unable to localize ear, is due to occlusion (by hairs) and low resolution. As shown in Fig. \ref{M_SSD2_1} the UESegNet-2 is not able to detect the right ear, as the image has two ears. The Fig. \ref{M_SSD2_2} as ear region is under huge illumination.

 \subsection{Comparative analysis with state-of-the-art}

 \begin{table*}[h]
 \caption{Comparative Performance Analysis [NM:Not Mentioned]}
 \label{Comparisons}
 \centering
 \begin{tabular}{{ |p{1.7cm}||p{2cm}|p{2cm}|p{5.5cm}|p{1.8cm}|p{1.5cm}|}}
 
\hline
	\textbf{Database} & \textbf{Reference}  & \textbf{ Test Images} & \textbf{Technique} & \textbf{ Threshold } &  \textbf{Accuracy } \\ \hline 

	\multirow{4}{*}{\textbf{IIT Kanpur}} & \cite{29}   & 500 &  Color based skin segmentation & NM   & 94.6\%  \\ \cline{2-6}
    
	 & \cite{20}  &  2672  & Connected components, Graph based approach & NM & 95.61\% \\ \cline{2-6}  
  
   &  Proposed Approach  &  530 & UESegNet-2 &IOU=0.5  & \textbf{99.29\%}  \\ \cline{2-6}
   &  Proposed Approach  &  530 & UESegNet-2 &IOU=0.6  & \textbf{97.89\%}  \\ \cline{2-6}	\hline
    
  \multirow{7}{*}{\textbf{UND-J2}} & \cite{35}   & 940 
	& Haar features with cascaded Adaboost classifier & NM & 95\% \\ \cline{2-6}
	 & \cite{20}  & 2244 & Connected components, Graph based approach 
	 & NM & 96.63\%
	\\
	\cline{2-6}
	 & \cite{36}  &  200  & Ear Template based approach & NM &  98\%
	\\
\cline{2-6}

  & \cite{40}  & 1776 & Edges, shapes and context information  & NM & 99\%  \\ \cline{2-6}
    
 & \cite{1} & 
1800  & Multiscale Faster Region Based CNN & Objectness Score & 100\% \\
\cline{2-6}
 &  Proposed Approach  &  1207 & UESegNet-2 & IOU=0.5 & \textbf{97.65\%} \\ \cline{2-6}
  &  Proposed Approach  &  1207 & UESegNet-2 & IOU=0.6 & \textbf{96.80\%} \\ \cline{2-6} \hline
  
  \multirow{3}{*}{\textbf{UBEAR}}
	  & \cite{1} & 9121   & Multi-Scale Faster Region Based CNN & NM & 98.66\% \\ \cline{2-6}

 &  Proposed Approach  & 4606 & UESegNet-2 & IOU=0.5 & \textbf{99.92\%}  \\ \cline{2-6}
 &  Proposed Approach  & 4606 & UESegNet-2 & IOU=0.6 &  \textbf{99.84\%}  \\ \hline

\end{tabular}
\end{table*}
 
 In this work, we have discussed four different models and tested their performance on six different databases at various values of IOUs. To compare the performance of the proposed model with existing approaches, we consider an IOU=0.5 and 0.6. As for good object detection proposal, an IOU should be more than 0.5. However, among the proposed models, the UESegNet-2 has obtained promising results, so we compared the performance of this model with existing state-of-the-art methods. In the literature, it has been found that most of the researchers have used IITK, UND-J2, and UBEAR databases, hence we compared the performance of UESegNet-2 with existing methods for these databases and results are shown in Table \ref{Comparisons}. On IIT Kanpur database the UESegNet-2 have achieved an accuracy of 99.29\% at IOU=0.5 and 97.89\% for IOU=0.6, which is better than the existing methods as in the literature a maximum of 95.61\% accuracy is reported by \cite{20}. On UND-J2 database, The UESegNet-2 has achieved an accuracy of 97.65\% at IOU=0.5 and 96.80\% at IOU=0.6  which is lesser than the accuracy achieved by \cite{1} on this database, as the authors have shown 100\% ear localization accuracy. However, they have not evaluated their model based on IOU. On UBEAR database, the UESegNet-2 has achieved an maximum accuracy of 99.92\% at IOU=0.5 and 99.84\% at IOU=0.6 and to the best of our knowledge, there is only one method proposed by \cite{1} used this database, in which authors have achieved an accuracy of 98.66\%. However, they did not evaluated their model based on IOU, rather they have calculated the accuracy based on the objectness score which is not the right parameter to measure accuracy as explained in section V. The results clearly indicate that our proposed models have achieved significantly better results than state-of-the-art methods.

\section{Conclusion and Future Direction}

Ear localization in 2D side face images captured in unconstrained environment  has great significance in the real world applications. Researchers have reported different approaches for ear localization and achieved significant accuracy. However, most of these approaches are on the constrained environment, this is due to the lack of availability of databases which satisfy all the conditions of the unconstrained environment. To accurately measure the accuracy of any object detection model an IOU parameter is used. However, the majority of the work discussed in the literature have ignored the IOU parameter to measure accuracy. In this paper, we have discussed four different models, and their performance is evaluated on six different benchmarked databases at different values of IOUs. Our proposed models UESegNet-1 and UESegNet-2 outperformed the existing state-of-the-art models FRCNN and SSD. Furthermore, the proposed models can be generalized for an object detection task in various areas. In future work, we will extend this problem for ear based personal authentication system in the wild.    

\textbf{Conflicts of Interest:} The authors declare no conflict of interest.

\begin{acknowledgements}
This is a pre-print of an article published in Pattern Analysis and Applications. The final authenticated version is available online at: https://doi.org/10.1007/s10044-020-00914-4”. 
\end{acknowledgements}

%bibliographystyle{plain}

%\section*{References}
\bibliographystyle{spmpsci}
%\bibliography{ear}

%\begin{thebibliography}{}
% etc
%\end{thebibliography}

\end{document}